\documentclass[10pt]{article}
\usepackage[utf8]{inputenc}
\usepackage{natbib}
\usepackage{graphicx,amsmath,amsfonts,amssymb,fullpage,url}
\usepackage{algorithm}
\usepackage{algorithmic}

\usepackage{parskip}
\usepackage{cmbright}
\usepackage{wrapfig}
\usepackage{booktabs}
\usepackage{bookmark}
\usepackage{subcaption}
\usepackage[most]{tcolorbox}

\usepackage{xcolor}
\usepackage{hyperref}

\usepackage[OT1]{fontenc}
\newcommand{\argmax}{\operatornamewithlimits{argmax}}
\newcommand{\figref}[1]{Figure~\ref{#1}}
\newcommand{\extfigref}[1]{Extended Fig.~\ref{#1}}

\renewcommand{\eqref}[1]{\hyperref[#1]{equation~\ref*{#1}}}

\definecolor{black}{RGB}{0, 0, 0}
\definecolor{orange}{RGB}{230, 159, 0}
\definecolor{skyblue}{RGB}{86, 180, 233}
\definecolor{bluishgreen}{RGB}{0, 158, 115}
\definecolor{yellow}{RGB}{240, 228, 66}
\definecolor{blue}{RGB}{0, 114, 178}
\definecolor{vermillion}{RGB}{213, 94, 0}
\definecolor{reddishpurple}{RGB}{204, 121, 167}
\definecolor{lightsteel}{HTML}{8f8f8c}
\definecolor{myred}{HTML}{e33222}

\newcommand{\goal}{{\color{skyblue}{g}}}
\newcommand{\offtask}{{\color{myred}{g'}}}
\newcommand{\newtask}{{g_{\tt new}}}

\newcommand{\deltart}{\Delta \text{Log RT}}

\newcounter{extendedfigure}

\newenvironment{extfigure}[1][]{%
    \begin{figure}[#1]%
    \refstepcounter{extendedfigure}%
}{%
    \end{figure}%
}

\newcommand{\envstate}{s}
\newcommand{\Envstate}{\mathcal{S}}
\newcommand{\buffer}{\mathcal{B}}
\newcommand{\traintasks}{\mathbb{M}_{\tt train}}
\newcommand{\testtasks}{\mathbb{M}_{\tt test}}

\title{\textbf{Preemptive Solving of Future Problems:\\Multitask Preplay in Humans and Machines}}
\author{\textbf{Wilka Carvalho}$^1$\footnote{Contact author: wcarvalho@g.harvard.edu}, \textbf{Sam Hall-McMaster}$^2$, \textbf{Honglak Lee}$^{3,4}$, \textbf{Samuel J. Gershman}$^{1,2}$ \\
$^1$Kempner Institute for the Study of Natural and Artificial Intelligence, Harvard University\\
$^2$Department of Psychology and Center for Brain Science, Harvard University \\
$^3$Department of Computer Science \& Engineering, University of Michigan \\
$^4$LG AI Research
}
\date{}

\begin{document}
\maketitle
\begin{abstract}
Humans can pursue a near-infinite variety of tasks, but typically can only pursue a small number at the same time. We hypothesize that humans leverage experience on one task to preemptively learn solutions to other tasks that were accessible but not pursued. We formalize this idea as \emph{Multitask Preplay}, a novel algorithm that replays experience on one task as the starting point for ``preplay''---counterfactual simulation of an accessible but unpursued task. Preplay is used to learn a predictive representation that can support fast, adaptive task performance later on. When implemented within deep reinforcement learning, it predicts human generalization to novel tasks more accurately than traditional planning and predictive representation methods across grid-worlds and a 2D Minecraft domain. Additionally, Multitask Preplay improves an artificial agent's ability to generalize to new environments sharing task co-occurrence structure. These findings demonstrate that Multitask Preplay is a scalable theory of human counterfactual learning and generalization; endowing artificial agents with this capacity can significantly improve their performance in challenging multitask environments.
\end{abstract}

While searching for coffee in a new neighborhood, you may come across gyms, grocery stores, and parks, enabling you to quickly find these locations later when you need them. One theory of how humans achieve this form of generalization posits that they learn an internal model of the environment, representing what is where~\cite{kool2016does}. ``Model-based'' algorithms can use this model to chart a path from the current location to any goal. Model-based approaches provide maximum flexibility (generalization to arbitrary goals and starting points), but come at the cost of requiring a possibly prohibitive amount of computation at decision-time.

At the other extreme, ``model-free'' approaches directly learn predictions of future states or rewards based on interacting with the environment~\cite{carvalho2024predictive}. For example, model-free deep reinforcement learning (RL) systems modify the weights of a neural network that outputs predictions of future reward based on sensory input~\cite{mnih2015human}. Predictions are effectively ``cached'' in the sense that they can be retrieved with relatively little additional computation. By choosing the actions that maximize future reward, a near-optimal policy can be cheaply computed. However, this comes at the cost of flexibility: a change in the task can require learning new weights from on-task experience.

In between these extremes are approaches that use offline computation (``background planning'') to compile model-based knowledge into cached predictions~\cite{sutton1990integrated,gershman2014retrospective}. This combines the flexibility of model-based approaches with the computational efficiency of model-free approaches. Classically, such approaches only cache predictions for a single task. This means that they are limited in their ability to flexibly generalize to new tasks without additional on-task experience, where each new task can require significant additional computation. Generalization is still bottlenecked by the need to rerun planning (either at decision-time or in the background) for each new task.

In fact, it is not necessary to run planning from scratch when different tasks share structure. For example, your search for coffee may lead you to a city center where many different shops are located; navigating to any of these potential goals will entail navigating to the city center and you can leverage this experience to learn about each of these goals. This shared structure can be leveraged by parameterizing a reward-predicting neural network to take the goal as input (what we refer to as \emph{goal conditioning};~\cite{schaul2015universal}).
However, learning to predict which actions will accomplish each goal still needs experience achieving that goal.
Inspired by an old idea in RL that one can use experience with one goal to learn about many goals~\cite{kaelbling1993learning}, we hypothesize that people \textit{preemptively} gain this experience by counterfactually simulating pursuit of locally accessible goals and providing this simulated experience as input to a learning algorithm.
Doing so then enables people to generalize to these goals later on with relatively little additional computation.
We refer to this as \emph{Multitask Preplay}. 

\begin{figure*}[tbhp]
    \centering
    \includegraphics[width=\linewidth]{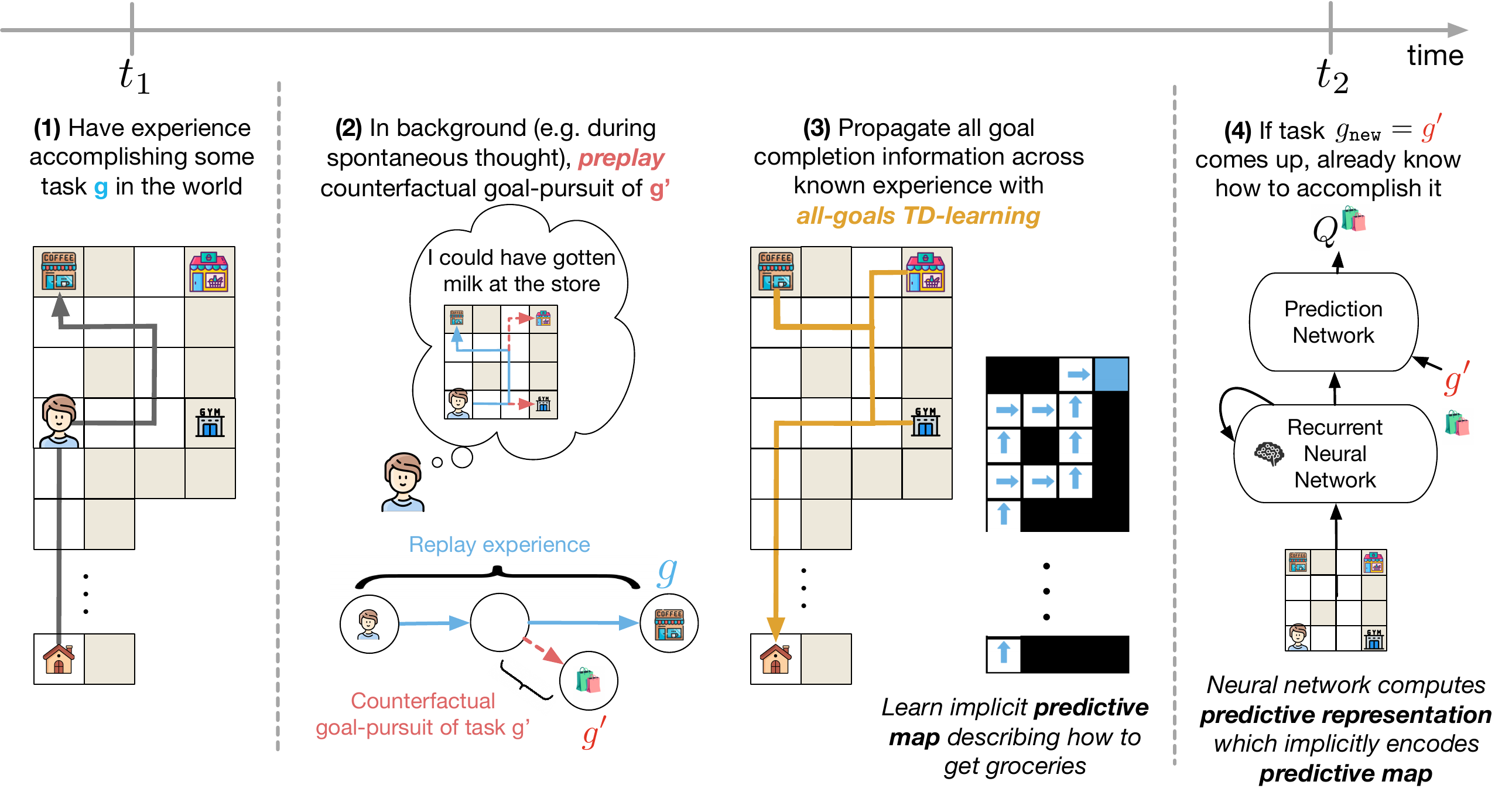}
    \caption{%
        Overview of Multitask Preplay. (1) At time $t_1$, an agent experiences a goal-directed task $\goal$~(e.g., going to a coffee shop). (2) During offline replay, the agent counterfactually simulates alternative accessible tasks $\offtask$~(e.g., going to a grocery store) based on the same experience trajectory. (3) Using temporal difference learning, the agent backs up information about both $g$ and $g'$ to learn a shared predictive representation. This enables it to form an implicit predictive map describing how to reach multiple goals. (4) At a future time $t_2$, if goal $\offtask$~becomes the actual task, the agent can perform it re-using the previously cached predictive representation---without requiring new experience or planning. This architecture allows generalization to unseen but structurally accessible tasks through preemptive counterfactual simulation.
    }\label{fig:preplay}
\end{figure*}

In this paper, we describe Multitask Preplay formally and present a series of human behavioral experiments testing its distinctive predictions. In particular, we present evidence that humans counterfactually simulate pursuit of alternative goals, using these simulations to update cached predictions.
We demonstrate the generalizability of Multitask Preplay by showing that our findings hold in both simple grid-worlds and in a complex, partially observable 2D Minecraft domain with numerous potential goals.
We conclude with AI simulation results that show that Multitask Preplay improves an artificial agent's ability to generalize to $10,000$ unique new environments with many hierarchical and inter-dependent tasks.
Taken together, these results suggest that preemptive counterfactual learning is a powerful algorithmic concept for both natural and artificial intelligence.

\section*{Preemptively solving future tasks with Multitask Preplay}

We formalize Multitask Preplay within the framework of sequential decision problems. At any given time $t$, the environment is described by a state $s_t$. The agent selects an action $a_t$ according to a probabilistic, state-dependent policy $\pi(a_t|s_t)$, and then receives reward $r_t$. The optimal policy $\pi^\ast$ maximizes expected discounted future return, or \emph{value}, defined as:
\begin{equation}\label{eq:q-value}
    Q^\pi(s,a) = \mathbb{E}\left[r_1 + \gamma r_2 + \gamma^2 r_3 + \ldots  | s_0 = s, a \sim \pi\right],
\end{equation}
where $\gamma \in [0,1)$ is a discount factor determining the relative weighting of long-term vs.~short-term rewards.
The agent can potentially have multiple goals, but is constrained to pursue them sequentially. We use $R_\goal(s)$ to denote the goal-dependent reward function. Accordingly, we use $Q(s,a,\goal)$ to express the fact that different tasks induce different value functions, which can be incorporated into a single ``universal'' value function~\cite{schaul2015universal}.

Value functions are interesting because they can be written recursively. This enables using a form of temporal difference (TD) learning known as \emph{Q-learning} to estimate it (see Methods for more details). The TD loss function for a single transition is:
\begin{equation}\label{eq:qlearning}
    L_{\tt Q\lambda} = \Big(
        \underbrace{r_t + \gamma \max_a \hat{Q}_{\theta'}^\pi(s_{t+1}, a, \goal)}_{\text{target}} - \underbrace{\hat{Q}_\theta^\pi(s_t,a_t,\goal)\vphantom{\max_a}}_{\text{prediction}}\Big)^2,
\end{equation}
where $\hat{Q}_\theta^\pi(s_t,a_t,\goal)$ is an approximation of the value function parametrized by $\theta$. We use deep neural networks to approximate the value function, updating the parameters by backpropagating the gradient of the TD loss through the network.

\begin{figure*}[!htb]
    \centering
    \includegraphics[width=.7\linewidth]{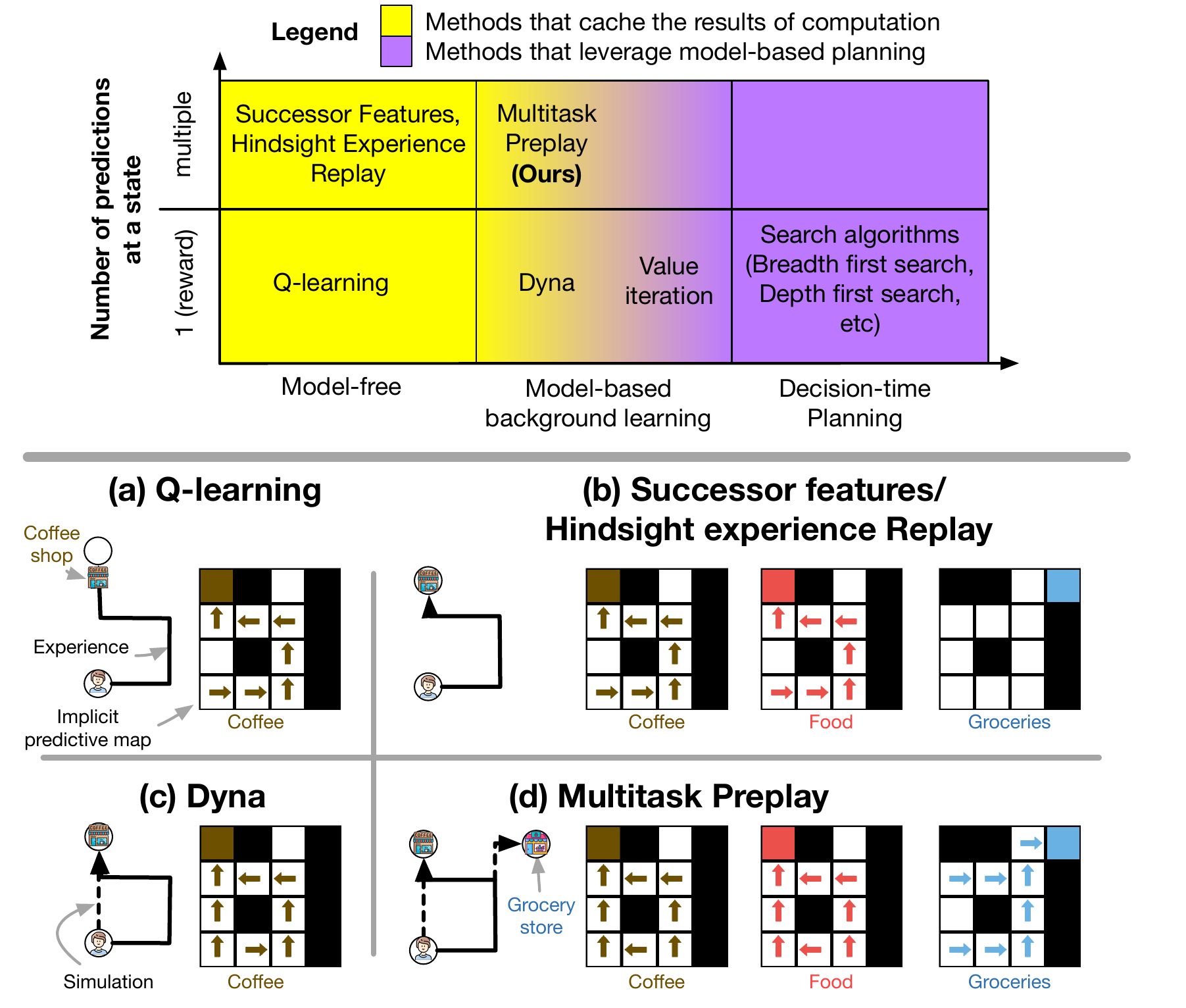}
    \caption{
        Overview of algorithms.
        (a) Q-learning is a model-free algorithm that updates cached reward predictions based on interactions with the environment.
        (b) Successor features are predictive representations that generalize reward-predictive representations to arbitrary features. Like Hindsight Experience Replay, they enable experience gathered while pursuing one goal to inform learning about other goals encountered along the way. Notably, if certain features (such as whether groceries are obtained) are not experienced, they are not encoded in the prediction.
        (c) Dyna uses model-based background learning to improve its cached reward predictions.
        (d) Multitask Preplay sits between Dyna and successor features. By leveraging counterfactual simulation of accessible tasks, it can learn to predict which actions lead not only to both the food and coffee, but also to the groceries.
    }\label{fig:algorithms}
\end{figure*}

Parameterizing the neural network as a function of $\goal$ allows its internal layers to learn shared structure across multiple tasks~\cite{schaul2015universal}, and thereby generalize to a new task with goal $\newtask$. For example, it might be many steps from your house to the coffee shop downtown, but once downtown, it might only be a few steps off your trajectory to the grocery store; much of the value function is the same for both goals. While there is some evidence that humans use this strategy~\cite{tomov2021multi}, non-trivial generalization in complex environments requires experience with millions of tasks~\cite{team2021open}. The bottleneck is direct experience: a neural network trained with TD learning can only generate good predictions in parts of the state space that are familiar to what the agent has explored before (\figref{fig:algorithms}a).

The same bottleneck applies to other methods. For example, the value function can be decomposed into a dot product between a  reward vector $w_\goal(s)$ and ``successor features'' (SFs):
\begin{align}
    \psi^\pi(s,a) = \mathbb{E}\left[\phi(s_1) + \gamma \phi(s_2) + \gamma^2 \phi(s_3)+ \ldots  | s_0 = s, a \sim \pi\right],
\end{align}
the expected future occurrence of features defined by $\phi(s)$~\cite{barreto2017successor}. The value function for a new goal $\newtask$ can then be approximated by taking the dot product between $\psi(s,a)$ and $w_{\newtask}(s)$~\cite{barreto2018transfer}. As for the value function, SFs can be estimated with a function approximator $\hat{\psi}_\theta^\pi$ such as a neural network. SFs have been successful at explaining some aspects of human generalization~\cite{tomov2021multi,momennejad2017successor}. 
However, it too can only generate predictions for familiar parts of the state space (\figref{fig:algorithms}b).

While agents might not have direct experience traversing all goal-relevant parts of the state space, they might be able to quickly synthesize a transition model. Think about what happens when you look at a map: even if you haven't traversed the area covered by the map, you can \emph{imagine} such traversals. This opens the door to forms of offline model-based computation for estimating values in the absence of direct experience. A classic version of this approach is the value iteration algorithm, which updates the values of all state-action pairs using dynamic programming. This is generally intractable for large state spaces, which is why most modern model-based algorithms rely on various forms of tree search, estimating values locally by simulating paths forward from the current state. Because these algorithms are applied at decision-time, they do not make use of offline computation. An alternative that mixes these ideas is Dyna~\cite{sutton1990integrated}, which uses TD learning to update value estimates locally around replayed past experiences, consistent with evidence from humans~\cite{gershman2014retrospective,momennejad2018offline}. 
However, this approach is still limited by the experience bottleneck (\figref{fig:algorithms}c).

We can uncork the bottleneck by using a model to simulate \emph{counterfactual} experience---in particular, the pursuit of accessible off-task goals. The intuition is that these are goals which are likely to be pursued in the future, and therefore the agent can benefit from preemptively ``preplaying'' pursuit of them, caching the results of this offline computation in the neural network parameters (\figref{fig:algorithms}d). This is the central idea underlying the Multitask Preplay algorithm (\figref{fig:preplay}), described more formally in the Methods. This algorithm combines and extends important aspects from the previously discussed approaches: it uses TD learning to efficiently train a deep neural network across accessible goals, with a goal-dependent parametrization to learn shared structure across goals, and uses offline simulation to estimate values for parts of the state space not directly experienced. While here we use Multitask Preplay to learn a value function, it could also be used to learn any predictive representation (e.g., successor features).
 
\section*{Evidence that humans use counterfactual simulation to preemptively solve new tasks}
\begin{figure*}[!tbh]
\centering
\includegraphics[width=\textwidth]{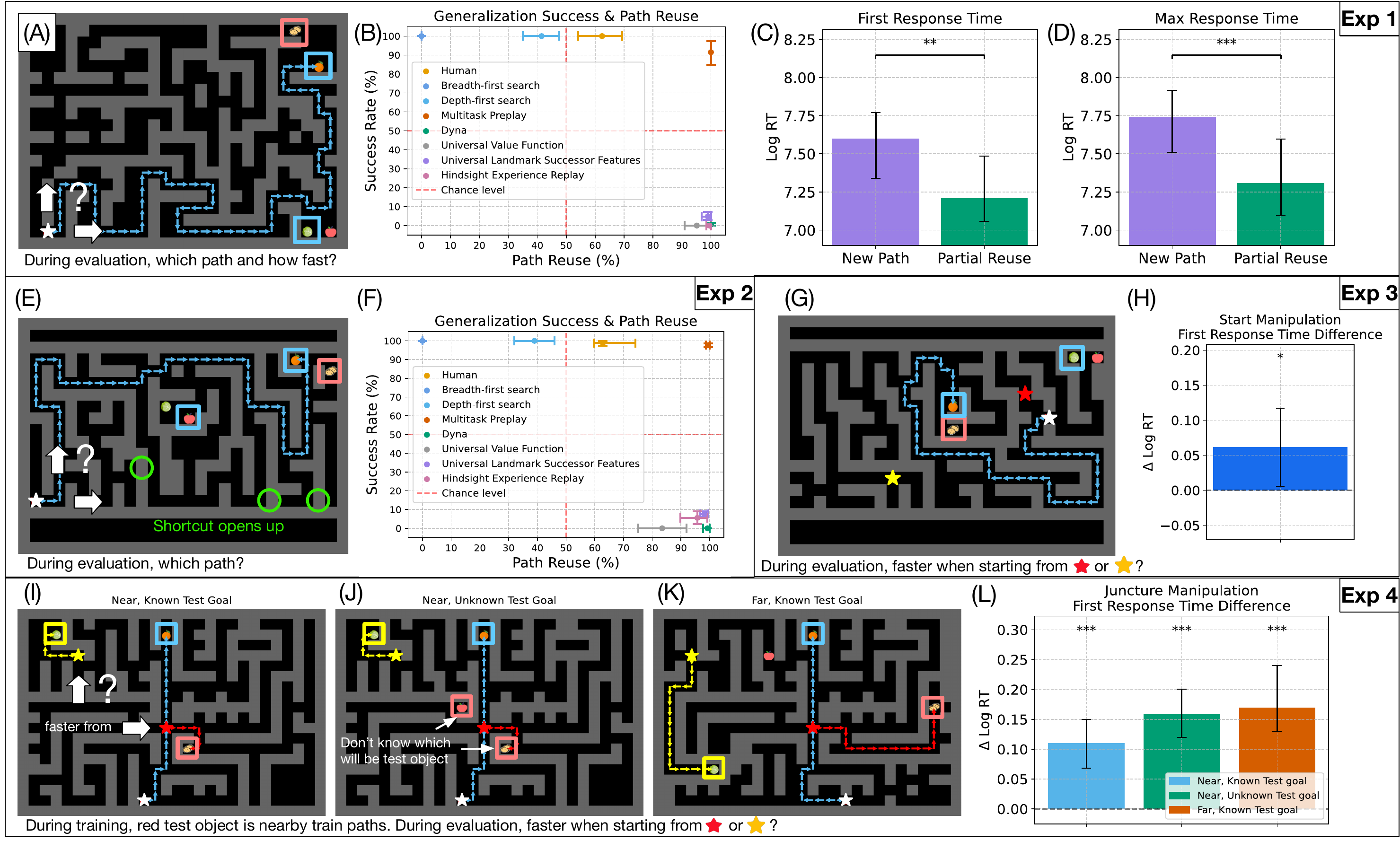}
\caption{
\textbf{Grid-world experiments and model comparisons}.
Across experiments, subjects learn to obtain training objects (blue boxes) before being evaluated on test objects (red/yellow boxes). White stars indicate spawning locations present in both training and evaluation; red/yellow stars show novel evaluation spawning locations. All figures display optimal paths from starting positions to target objects. Objects are randomized across subjects, who experience 4 trials per maze with different $90^o$ rotations. We compare mean subject and model performance across these trials. All plots show the data mean and 95\% confidence intervals.
\textbf{Experiment 1 (A-D)} tests if people ``preplay'' the evaluation task as they complete the nearby training task (A;\@ objects in top-right corner).
(B) Model comparisons show only decision-time planning methods and Multitask Preplay match human success rates. Human path reuse rates fall between decision-time planning (Depth-first search) and Multitask Preplay.
(C-D) Log maximum and first per-step response times during evaluation grouped by whether subjects took a new path or partially reused a training path.
\textbf{Experiment 2 (E-F)} tests if people continue to be biased to partially reuse a past path even when a shortcut becomes available (K;\@ note blue path crossing maze boundaries).
(L) Model comparisons show people best match generalization success and path reuse rates of Multitask Preplay.
\textbf{Experiment 3 (G-H)} tests if people are faster at evaluation task 1 (blue) vs evaluation task 2 (yellow) even if its longer. This would happen if the task was ``preplayed'' during training.
(J) Log first response time difference between evaluation tasks 2 and 1.
\textbf{Experiment 4 (I-L)} tests if people ``preplay'' eval task 1 (blue) at the closest location of a training path near the task object (blue start). In (E) and (G), subjects know the blue evaluation goal during training. In (F), it is unknown and one is randomly chosen from the two blue options. For all three, optimal paths during blue and yellow evaluation trials are equivalent but rotations of each other.
(H) Log first response time difference between evaluation tasks 2 and 1.
}\label{fig:jaxmaze}
\end{figure*}

Multitask preplay posits that 
people (1) ``preplay'' sequences which accomplish counterfactual tasks accessible from their direct experience,
and (2) cache the results of this computation into a predictive representation using RL.%
This model makes 4 predictions about how people will pursue novel goals that were accessible from their prior experience. We tested each prediction in separate behavioral experiments using simple grid-worlds (see Methods for more details). Human subjects controlled a red triangle, which could move through the grid-world and consume food items. Subjects were first exposed to a set of training tasks, and then given one or more evaluation tasks designed to test their generalization. We collected $n=100$ subjects for each experiment.~Following past work, we use response times as a proxy for how much computation someone has performed~\citep{hick1952rate,donders1969speed,sternberg1969memory,tomov2021multi} All analyses were preregistered, except where noted.

\textbf{Prediction 1: partial reuse of past paths to solve new tasks.} Past work has shown that people partially reuse past paths when solving new tasks in a small state space~\cite{huys2015interplay}. We extend this work with experiment 1, by measuring reuse in a larger grid-world domain, exploiting response time data to test whether people are taking actions based on ``preplayed'' task-completion knowledge (\figref{fig:jaxmaze}a). If people are partially reusing paths, they should follow these paths for new tasks even when doing so is suboptimal (in terms of shortest path length). If they are relying on cached predictions rather than planning, they should be \emph{faster} when reusing paths even though the path length is \emph{longer}.
Subjects were consistently successful at solving the evaluation task (\figref{fig:jaxmaze}b), a non-trivial achievement given that several sophisticated algorithms that learn generalizable predictive maps of the environment, including Universal Value Functions, Universal Successor Features, and Hindsight Experience Replay (discussed further below), have generalization success rates near 0. 
Critically, subjects partially reused a training path a mean of $62.39\%$ (95\%CI: 53.99, 69.53) of the time, significantly greater than %
    $50\%$ ($p=0.0049$, $Z=-0.15$, one-sided Wilcoxon signed-rank test;~\figref{fig:jaxmaze}b; see~\extfigref{ext:path-reuse-rates} for the path reuse distribution).
Our preregistered analysis of the first response time (RT), which we expected to be the most diagnostic of people planning an entirely new path or reusing cached knowledge, showed that RTs were significantly lower on the first step of reused paths compared to new paths %
    (linear mixed effects model: $\beta$=$-0.21$, 95\%~\text{CI: }$-0.36, -0.06$, t($133$)=$-2.81$, p=$0.005$;~\figref{fig:jaxmaze}c).
While most trials had a single RT significantly larger than all others when presumably most of the planning was happening (e.g. 1 second vs. .25 seconds), sometimes this occurred on later steps, somewhat idiosyncratically for different subjects (see \extfigref{ext:two-paths} for illustrative examples and \extfigref{ext:top10} for an analysis of RT magnitudes). 
We therefore conducted an exploratory follow-up analysis of the maximum RT on each path and identified a significant difference between partially reused and new paths %
    (linear mixed effects model: $\beta$=$-0.32$, 95\%~\text{CI: }$-0.49, -0.15$, t($133$)=$-3.64$, $p<.001$;~\figref{fig:jaxmaze}d).
    Under the premise that longer RTs indicate longer planning operations, the two RT analyses provide converging evidence in support of the hypothesis that planning is faster with reuse, even when the paths are longer (see \extfigref{ext:two-paths} for some illustrative examples).

Most algorithms we tested (see Methods for details) either failed to match human success rates, or failed to match human path reuse rates (\figref{fig:jaxmaze}b; see \extfigref{ext:jaxmaze_train} for train performance). While Universal value functions can in principle generalize to new goals, they require experience in the relevant part of the state space (which our experiments precluded). Universal landmark successor features (SFs) are predictive representations that describe whether counterfactual goals are nearby or can be completed. They can be used to compute action values for new goals, but, unlike Multitask Preplay, value computation relies entirely on caching (no planning). This approach fails to generalize in our task because planning is required to chart action sequences off the training paths (see~\extfigref{ext:sf-analysis} for analysis).
HER similarly fails because its value estimates only cover goals along paths the agent has already traversed (see~\extfigref{ext:her-comparison} for analysis).
Decision-time planning algorithms (breadth-first search and depth-first search) are able to chart such sequences, and hence achieve success rates comparable to humans, but they predict less reuse of the optimal training path compared to humans, and they do not predict shorter RTs for reused paths (planning time only depends on path length).

\textbf{Prediction 2: partial reuse of past paths to solve new tasks, even when a shortcut opens up.}
If people have preplayed a task that was previously accessible, they might forego planning when a shortcut opens up and partially reuse a prior path to the novel task (optimal length=61, turns=20). 
In contrast, if they have not, then they would plan at decision-time and more likely use the shortcut path that has an optimal length of 55 and 17 turns. 
We test for this in experiment 2 (see~\figref{fig:jaxmaze} E).
Subjects partially reused a training path a mean of $62.80\%$ (95\%CI: $59.68, 73.91$), significantly greater than $50\%$ (one-sided Wilcoxon signed-rank test: $p<.001$, $Z=1.08$; ~\figref{fig:jaxmaze} F~\figref{fig:jaxmaze}b; see~\extfigref{ext:shortcut-rates} for the path reuse distribution).
Like in our previous experiment, caching-based solutions fail to generalize: Universal Value Functions achieve $0\%$ success, while mechanisms for transfer across goals fare only marginally better (Universal Landmark SFs $7.65\%$, HER $5.56\%$).
Decision-time planning algorithms can form plans to accomplish these goals but predict significantly less path reuse than what people exhibit ($39.00\%, 95\%~\text{CI: }31.50, 46.00$ for depth-first search, $0\%$ for breadth-first search).
In contrast, Multitask Preplay, which predicts that people chart a path off their prior experience and cache the results, reproduces the human pattern of solving the task while preferentially reusing prior paths, with a mean success rate of $97.74\%$ ($95\%~\text{CI: }96.71, 98.75$) and path reuse rate of $99.50\%$ ($95\%~\text{CI: }98.74, 100.00$).

\textbf{Prediction 3: fast response times for new tasks in novel starting locations.}
In experiment 3, we look for evidence of task preplay at novel starting locations from a subject's prior experience.
We use an evaluation protocol where a subject either begins at an intermediary position of their prior experience (condition 1) or in a novel part of the map (condition 2).
Importantly, the optimal path in condition 2 (length=44, turns=14) is less than in condition 1 (length=56, turns=19; see~\figref{fig:jaxmaze} G).
This means that if they are planning in both conditions, they should be faster in condition 2. In contrast, if they have preplayed the novel task and cached the results already during training, they should be faster in condition 1, despite it requiring a longer task solution.
We measured $\deltart=\log \text{FirstRT(cond2)} - \log \text{FirstRT(cond1)}$.
Despite condition 1 being $33\%$ longer than condition 2, $\deltart$ was significantly greater than $0$, i.e. people are significantly faster to respond at a \textit{further} known location (one-sided Wilcoxon signed-rank test: $\text{median }\deltart=0.06, (95\%~\text{CI: }0.01, 0.12), p=0.015$; see~\figref{fig:jaxmaze} H).
This provides evidence that people propagate counterfactual task completion knowledge to novel starting points which were intermediary points of their previous experience.

\textbf{Prediction 4: fast selection of novel behaviors that solve new tasks.}
The previous results suggested that people use cached knowledge to quickly pursue novel goals through familiar paths.
While Universal Landmark SFs do not predict the previous behavioral results, one possibility is that people use a similar landmark-oriented algorithm to navigate towards a goal and then start planning when at a \textit{juncture point} that departs from their previous experience.
This contrasts with the prediction of Multitask Preplay that people have already ``preplayed'' actions that depart from the juncture point and require no more planning.
To test for evidence of this preplay, we spawn people either at a juncture point (condition 1) or in a new part of the map where they will need to perform decision-time planning because their previous experience cannot be reused (condition 2).
If they have preplayed the task, they should be faster in condition 1 than in condition 2. If they are performing decision-time planning once at the juncture point, then the response times should be approximately equivalent in the two conditions.

We ran $3$ preregistered experiments.
In the first (\figref{fig:jaxmaze} I), $\offtask$ is nearby and known---similar to experiment 1.
In the second (\figref{fig:jaxmaze} J), $\offtask$ is nearby but unknown---this tests how automatic preplay for tasks may be as people do not know they will be evaluated on them later on.
In the third (\figref{fig:jaxmaze} K), $\offtask$ is far and known---this tests the bounds of preplay and shows whether it is robust to longer distances.
We measured $\deltart=\log \text{FirstRT(cond2)} - \log \text{FirstRT(cond1)}$.
$\deltart$ was significantly larger than $0$, even though the distance to the goal was the same across conditions 
(one-sided Wilcoxon signed-rank test:
$\text{median }\deltart=0.11, Z=4.56, (95\%~\text{CI: }0.07, 0.15), p<.001$ when near and known; 
$\text{median }\deltart=0.17, Z=6.43$, $ (95\%~\text{CI: }0.12, 0.20), p<.001$ when near and unknown; and 
Paired t-test when when far and known:
median $ \deltart=0.172, t(99)=5.85, (95\%~\text{CI: }0.131, 0.240), p<.001$; \figref{fig:jaxmaze} L). We show examples in~\extfigref{ext:juncture} where we see that subjects can have a larger gap between the two conditions.
We emphasize that generalization tasks had no overlap with training paths, so fast performance must come from caching. We hypothesize that people cache results of preplaying paths from juncture points---even when future goals are unknown.

\section*{Signatures of Multitask Preplay in a 2D minecraft domain}
\begin{figure*}[!htb]
    \centering
    \includegraphics[width=\textwidth]{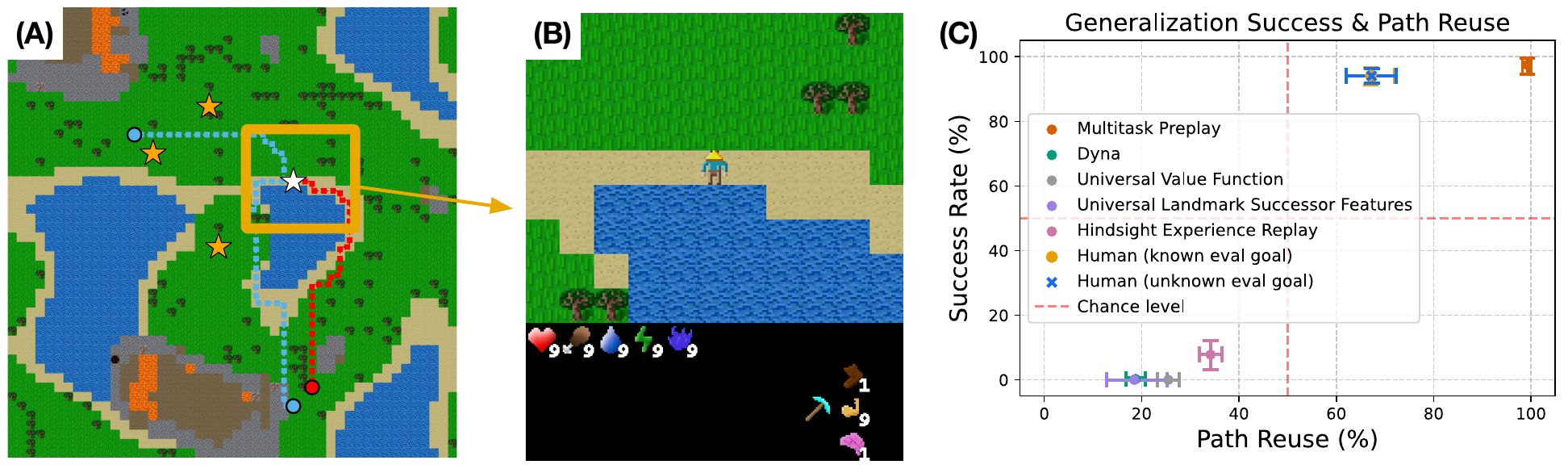}
    \caption{
        \textbf{2D minecraft experiments and model comparisons.}
        Subjects learn to obtain stones (a ruby, sapphire, or diamond) across $4$ procedurally generated maps. Two are train stones, 1 is an evaluation stone. In one condition, they are told the evaluation stone; in another, they are not. 
        \textbf{(A)} An example of the full map. Orange stars indicate spawning locations during training; the white star was a spawn location during both training and evaluation. We show the optimal path to train objects (blue) and evaluation objects (red). The evaluation object is always visibly nearby one of the training objects.  
        \textbf{(B)} A zoomed-in view around the agent. During training, subjects get the full map and local view; during evaluation, they only get the local view.
        \textbf{(C)} Model comparisons show people best match generalization success and path reuse rates of Multitask Preplay. All plots show the data mean and 95\% confidence intervals. Note that several points are overlapping and therefore not visible.}\label{fig:crafter}
\end{figure*}

Here, we test Multitask Preplay's potential as a generalizable learning theory that can predict human behavior in increasingly ecologically valid conditions.
We accomplish this by generalizing prediction $1$ from the previous section to a more complex 2D Minecraft domain ``Craftax''~\cite{matthews2024craftax}, where a subject needs to navigate to goal objects with only partial observation of a large world with many objects.
We additionally extend prediction $1$ to the setting where people are not told what object they will be evaluated on. 

\textbf{Prediction 5: partial reuse of past paths to solve new tasks in a partially observable, large world with many objects.}
This preregistered experiment conceptually replicated experiment 1 in the Craftax domain. The key design feature of this experiment is the existence of multiple paths to get to a test object, and a suboptimal path overlapping with an optimal training path.
Importantly, during training we provide subjects with the full map (\figref{fig:crafter} A) and a partial view of the world (\figref{fig:crafter} B) to mitigate the challenge of exploring this environment from only a partial view; however, during testing they only get the partial view.
We construct maps such that test objects are visible when subjects get near a train object.
We tell subjects that they are a crafter in a mining world with certain stones to mine.
In condition $1$, we tell subjects which will be their test stone goal, and in condition $2$, replicating experiment 2, we do not tell them.
This means that people are required to cache enough knowledge of this large map to support selecting actions to a new goal from a partial view of the world---like the real world.

In this more challenging domain, subjects consistently solve the evaluation task from only partial views of a large world.
Both when subjects knew their future evaluation goal (condition 1) and when they didn't (condition 2), they reused training paths to novel task objects a mean of $67.1\%$ of the time in condition 1 ($95\%~\text{CI: }[62.25, 72.00]$) and $67.3\%$ in condition 2 ($95\%~\text{CI: }[62.33, 72.33]$), significantly greater than $50\%$ (one-sided Wilcoxon signed-rank test: $p<.001$ for both conditions; \figref{fig:crafter} C; see \extfigref{ext:craftax_train_test} for train performance; see \extfigref{ext:craftax-path-reuse-rates} for the path reuse distribution).
In terms of decision-time planning approaches, no planning model we know of supports planning to novel goal objects from partial observations of a large world without assuming knowledge of the environment state space, so we do not compare to this model here.
In contrast, approaches that rely entirely on caching fall short: Universal Value Functions and Universal Landmark SFs achieve $0\%$ success, and HER reaches only $8.22\%$.
Multitask Preplay achieves a mean success rate of $97.5\%$ ($95\%~\text{CI: }94.25, 99.50$) and reuses prior paths $99.3\%$ of the time ($95\%~\text{CI: }98.25, 100.00$). As in Experiment 1, no other algorithm simultaneously achieves high success and high path reuse; Multitask Preplay is the only algorithm that, like humans, both solves the generalization task and preferentially reuses prior paths to do so.
Thus, Multitask Preplay provides a parsimonious account for what can support this kind of generalization in humans: learning a goal-oriented predictive map over many goals via counterfactual simulation.

\section*{Multitask Preplay improves generalization performance of artificial agents in novel environments}

\begin{figure*}[!htb]
    \centering
    \includegraphics[width=\textwidth]{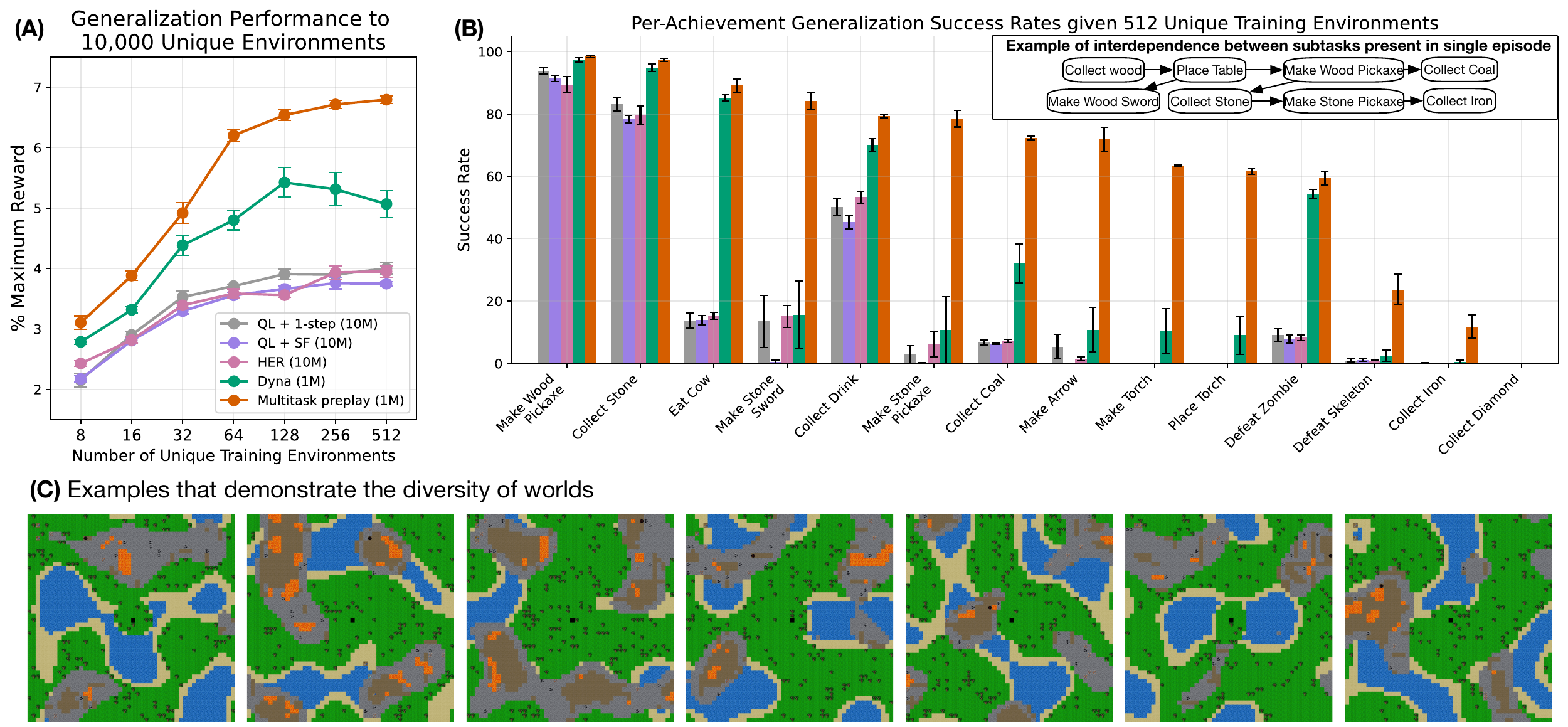}
    \caption{
        \textbf{AI simulation results studying generalization to 10,000 unique new testing environments.}
        \textbf{(A)} Each point is the mean and standard error across $5$ model initializations ($140$ total individual training runs).
        Model-based methods are run for $1$ million training steps and model-free methods are run for $10$ million training steps. 
        \textbf{(B)} Performance on individual subtasks during generalization when trained with 512 training environments.
        \textbf{(C)} Examples of procedurally-generated maps in Craftax domain. We hold-out 10,000 unique maps for evaluation.
    }
    \label{fig:crafter-ai}
\end{figure*}

So far, we have shown that when tasks co-occur, this can help with multitask generalization, as the solution for one task can be used to preemptively learn the solutions to other tasks.
However, we have only considered generalization across tasks defined by acquiring a single object; whereas real-world tasks are often hierarchical with shared subtasks (e.g., you must get a knife before either slicing fruit or vegetables). Additionally, we often generalize known task behaviors to new environments (e.g., we can find and use knives when traveling to new homes).
Here, we generalize our AI simulation results to Craftax~\cite{matthews2024craftax}. Craftax is a challenging long-horizon domain with a maximum score of 226 where (1) there is only a single hierarchical task composed of many interdependent subtasks, and (2) there are regularities in subtask co-occurrence.
We argue that subtask co-occurrence can aid in multitask generalization across environments and study the extent to which Multitask Preplay can exploit this structure.

\textbf{Prediction 6: Improved generalization to novel environments that share subtask co-occurrence structure.}
We test this prediction in an augmented variant of the standard Craftax domain where we additionally expose agents to a function $\phi_{\tt avail}(s)$ that describes what goals or \textit{subtasks} are available in a state.
Importantly, Craftax is a \textit{single-task} domain where there is a single global reward function $R_\goal(s)$ that gives different priorities to different ``subtasks'' $\{\offtask\}$. 
By design, the reward function is simply additive: $R_\goal(s)=\sum_{\offtask} R_{\offtask}(s)$.
We focus on how methods in \figref{fig:algorithm-comparison} can improve generalization of an action-value function $\hat{Q}_\theta(s,a,\goal)$ and study how different methods can leverage $\phi_{\tt avail}(s)$ to best enable generalization to 10,000 new unique environments after a variable number of training environments. We describe this protocol in more in detail in Methods: Craftax Environment Transfer Details.
Multitask Preplay leverages $\phi_{\tt avail}(s)$ to sample $\{\offtask\}$ and then performs counterfactual simulation that optimizes for the corresponding $\{R_{\offtask}\}$.
In contrast, Dyna simply uses $\phi_{\tt avail}(s)$ as input for predicting $\hat{Q}_\theta(s,a,\goal)$ but performs counterfactual simulation that tries to optimize $R_\goal$ for an environment.
We have two model-free baselines {\tt QL+1-step} and {\tt QL+SF} which both use $\phi_{\tt avail}(s)$ as either $1$-step or $n$-step prediction targets to drive representation learning---a strategy which has been shown improve generalization~\cite{agarwal2021contrastive}.
Importantly, we keep the total number of training steps constant to 1 million steps for model-based background learning methods and 10 million steps for model-free methods.

We find that as the number of training environments increases but the total steps to learn from stays fixed, Multitask Preplay best generalizes to 10,000 unique new environments, whereas Dyna and model-free methods asymptote in their generalization performance below this level (\figref{fig:crafter-ai} A). We suspect HER offers no advantage here because Craftax already rewards every reward-relevant goal a trajectory encounters, leaving nothing for relabeling to recover.
Unsurprisingly, if we remove regularities in task object co-occurrence, we see a drop in performance (see~\extfigref{ext:randomization_ablation})
If we look at individual tasks, we see that Multitask Preplay can better perform tasks with more subtasks (\figref{fig:crafter-ai} B).
That Multitask Preplay leads to improved training performance is unsurprising (see ~\extfigref{ext:train_eval}): since~\cite{kulkarni2016hierarchical}, it has been known that treating visible objects as goals can lead to faster learning on long-horizon tasks.
However, what has not been shown before is that backing up subgoal $\offtask$ knowledge to a global $\hat{Q}_\theta(s,a,\goal)$ over many environments that share subtask co-occurrence structure gives an effective ``prior'' Q-function with good generalization to new environments.
We show detailed improvements across training settings in~\extfigref{ext:all_bars} (for a comparison to state-of-the-art methods in the standard craftax setting, see ~\extfigref{ext:baseline_comparison}).
We emphasize that this is true even when keeping the number of training samples constant, leading to fewer training steps per environment (e.g., $\approx 2,000$ steps/env when using $512$ environments).

\section*{Discussion}

Our experiments provide indirect evidence that people use counterfactual simulation to learn preemptively about off-task goals. People were faster at solving new tasks when they could partially reuse old solutions, even though those solutions were longer than the optimal path. This form of reuse arises specifically when people can generate counterfactual plans to off-task goals from their current position. People exhibited a tendency to reuse past solutions even when better paths became available. Similar patterns were observed in a more complex, partially observable domain (a 2D version of Minecraft).

We argue that this reuse behavior is adaptive, despite introducing systematic biases, because it enables people to achieve flexibility to solve many different tasks without significant decision-time planning---a fundamental bottleneck for model-based algorithms.
Across machine learning, cognitive science, and neuroscience, Universal Value Functions and Successor Features have been proposed as alternative approaches that enable cheap computation of this kind of flexible behavior~\cite{schaul2015universal,momennejad2017successor,borsa2018universal,tomov2021multi}.
However, our simulation results show that this is not the case for the domains we study. 
In contrast, we show that a new algorithm, Multitask Preplay, succeeds where humans succeed---and accomplishes these successes with human-like biases for reusing previous solutions. The algorithm incorporates preemptive counterfactual simulation into a general and powerful RL architecture that can significantly improve generalization to new tasks compared to modern baselines.
While each individual experiment may admit an idiosyncratic alternative theory---for example, people may avoid unknown paths as ``risky'' for experiment 1---Multitask Preplay provides a simple and parsimonious answer across the 5 experiments we ran.
Another limitation of our work is that we currently provide only indirect evidence via response time correlations and somewhat coarse behavioral predictions. Future work can pursue complementary measurement tools such as eye tracking~\citep[as in][]{gerstenberg2017eye} or neuroimaging~\citep[as in][]{momennejad2018offline}.

In addition to our cognitive science results, our AI simulations show that learning from both real and counterfactual tasks leads to more robust reward predictions, and improves generalization to $10,000$ unique new environments that share task co-occurrence structure with the training environments. Unfortanately, this experiment would be hard to run with humans, as they would likely require prohibitively long sessions. We use the AI results to argue that this property common to real-world environments is something that can be effectively exploited by Multitask Preplay.

While past human behavioral studies have provided evidence for policy reuse~\cite{huys2015interplay} and offline simulation~\cite{gershman2014retrospective,momennejad2018offline}, these experiments used environments with small state spaces and lacked generalization tests to novel goals. A number of studies have examined feature-based generalization~\cite{norbury2018value,schulz2018putting,wu2024unifying}, but only in bandit tasks that lack sequential structure. Other studies have examined generalization to novel goals~\cite{tomov2021multi,hall2025neural}, but were not designed to test for counterfactual simulation. Ours is the first series of experiments to comprehensively test the hypothesis that people engage in preemptive counterfactual simulation of off-task goals.

Within machine learning, a common strategy for learning about alternative goals is to leverage Hindsight Experience Replay (HER)~\cite{andrychowicz2017hindsight}, which relabels failed experiences by setting reached states as counterfactual goals. This can accelerate main-task learning~\cite{yang2021mher} and improve exploration~\cite{guo2019directed,ren2019exploration}. While some work has used states from erroneous planning for relabeling~\cite{moro2022goal}, none intentionally selects available but unpursued goals for simulation. Multitask Preplay instead chooses unpursued goals and augments experience with counterfactual simulation toward these goals. Our approach falls within the ``many-goals RL'' framework that uses each experience to learn about multiple goals~\cite{kaelbling1993learning,veeriah2018many}. While prior work used model-based simulation for many goals in robotics~\cite{mendonca2021discovering}, it focused on unsupervised pretraining rather than the orthogonal issue of integrating replay and counterfactual simulation to learn about other \textit{accessible} but unpursued tasks. Kaelbling et al.~\cite{kaelbling1993learning} first described combining Dyna with counterfactual simulation for many goals, but neither they nor subsequent work demonstrated simulation results or explored transfer benefits. We are the first to detail show this enables learning of goal-oriented predictive representations that transfer to new tasks and new environments where tasks share co-occurrence structure, and the first to detail the algorithmic components needed for successful model-based all-goals learning in large worlds.

Our work raises a number of questions. First, how do people select off-task goals? We studied only the simplest possible selection algorithm, but one could sample goals based on criteria such as uncertainty, future relevance, or learning progress. This may allow agents to better allocate computational resources as the number of possible goals increases in naturalistic environments. 
Second, our simulations assumed access to an oracle world model, but in general this needs to be learned from experience. Future work could either leverage a Vision-language model such as RT-2~\cite{brohan2023rt}, or learn the model with a modern algorithm such as MuZero~\cite{schrittwieser2020mastering}. Finally, we have only studied video game applications in this paper, but Multitask Preplay could be applied to a wider range of challenging problems. For example, the Habitat AI domain~\cite{savva2019habitat,szot2021habitat} contains virtual scans of thousands of real homes which possess the task co-occurrence property we argue Multitask Preplay exploits. This domain could provide an opportunity to compare human and AI performance on naturalistic object-oriented tasks, such as finding mugs or preparing food~\cite{carvalho2020reinforcement}. 

Our work additionally opens up interesting questions for neuroscience.
Multitask Preplay is inspired by preplay activity in the hippocampus, where place cells (neurons selective for particular locations in space) are sequentially activated during sleep or quiescence. Preplay sequences are distinct from ``replay'' sequences in that they do not rigidly reflect an animal's past experience, but rather experience it has later once awake~\cite{dragoi2011preplay,dragoi2013distinct,olafsdottir2015hippocampal}. The computational function of neural preplay is currently unclear, but it may play a role enabling fast, adaptive decision making. 
Multitask Preplay may offer a new theoretical explanation, and future neurophysiology studies could use versions of our experimental protocols to more specifically test this hypothesis.

\subsection*{Acknowledgements}
We thank Andrew Lampinen, Angelos Filos, Shuze Liu, Anne Collins, Tomer Ullman, and the Computational Cognitive Neuroscience Lab at Harvard generally for helpful discussions on the topic. This work was supported by the Department of Defense Multidisciplinary University Research Initiative (MURI) program under Army Research Office (ARO) grant W911NF-23-1-0277, a Polymath Award from Schmidt Sciences, a Philip Wrightson Award from the New Zealand Neurological Foundation, and by a gift from the Chan Zuckerberg Initiative Foundation to establish the Kempner Institute for the Study of Natural and Artificial Intelligence.

\subsection*{Author contribution}
W.C., H.L., and S.G. conceived the model.
W.C., S.H., and S.G. conceived the experiments.
W.C. conducted all simulations and ran all experiments.
All authors wrote the manuscript.

\subsection*{Conflict of Interest}
H.L works at LG AI Research. All other authors declare no competing financial interests.

\subsection*{Data availability}
Data and preregistration for the JaxMaze experiments are available through the Open Science Foundation repository \url{https://doi.org/10.17605/OSF.IO/M53QH}; and for the Craftax experiments at \url{https://doi.org/10.17605/OSF.IO/B2EVM}.

\subsection*{Code availability}
Code for this study is available at \url{https://github.com/wcarvalho/multitask_preplay}.

\bibliographystyle{unsrt}
\bibliography{preplay-bib}

\clearpage
\appendix
\section*{Methods}

\subsection*{Models}

We formulate domains as Partially Observable Controlled Markov Processes $\mathcal{C}=\langle \Envstate, \mathcal{A}, \mathcal{X}, P, O\rangle$, where $\Envstate$ denotes the environment state space, $\mathcal{A}$ denotes its action space, and $\mathcal{X}$ denotes (potentially partial) observations of the environment.
When an agent takes an action $a\in\mathcal{A}$ in state $\envstate\in \Envstate$, the next state $\envstate'$ is sampled according to a next state distribution $\envstate' \sim P(\cdot|\envstate, a)$, and an observation $x'$ is generated deterministically via $O(\envstate')$.
We formulate tasks as Partially Observable Markov Decision Processes $\mathcal{M}=\langle \mathcal{C}, R_\goal\rangle$, with goal-parameterized reward functions: $r_t = R_\goal(\envstate_t)$.
Henceforth, we will use ``goal'' and ``task'' interchangeably since a goal specifies a task in this setup.
We train both humans and models using the same set of training tasks $\mathbb{M}_{\tt train} = \{\mathcal{M}\}^{n_{\tt train}}_{i=1}$ and then evaluate them on the same set of evaluation tasks $\mathbb{M}_{\tt test} = \{\mathcal{M}\}^{n_{\tt test}}_{i=1}$.

For all deep RL models, we assume access to a learned ``agent state'' representation which aggregates all information observed so far within an episode via a recurrent function of history (e.g., a recurrent neural network); to avoid proliferation of notation, we will assume here that the agent and environment states are identical, but in general they need not be. Thus, we write the recurrent state function as $s_t=f_{\theta}(x_{t}, s_{t-1}, a_{t-1})$, where $\theta$ are the model parameters.
We optimize models so that they learn a behavior policy $\pi_\theta(a|s,\goal)$ that maximizes the future reward it can expect from taking an action $a$ in state $s$, i.e. to maximize the value function (\eqref{eq:q-value}).
All models use a target network \cite{mnih2015human} to stabilize learning. A separate copy of the parameters $\theta'$ is held fixed while computing bootstrap targets and periodically synchronized with $\theta$ every 1000 environment steps, preventing the optimization targets from shifting with each gradient update.

All deep RL algorithms we consider have a ``replay'' component to training: after an agent experiences a trajectory $\tau=(x_1, a_1, \ldots, a_{T_{\tt env}-1}, x_{T_{\tt env}})$ in the environment accomplishing some task $\goal$, both are added to a memory buffer $\buffer$. For all learning algorithms, subsequences are then sampled and ``replayed'', i.e. $\tau \sim \buffer$, and processed by a history function to obtain an agent state representations for each time-point $s_t = f_{\theta}(x_t, s_{t-1}, a_{t-1})$ leading to $\tau=(s_1, \ldots, s_{T})$, where $T << T_{\tt env}$. Algorithms then differ in how they construct a learning objective with these data.

\subsubsection*{Q-learning}

The simplest form of model-free RL is to directly approximate the value function (\eqref{eq:q-value}) using a parametrized function approximator $\hat{Q}_\theta^\pi(s_t,a_t,\goal)$, here a deep neural network, trained to optimize a loss function comparing value estimates to a target function (\eqref{eq:qlearning}).
The target function is a ``bootstrapped'' estimate of the true value function.
The target in \eqref{eq:qlearning} is called a 1-step return because it combines the observed reward at one timestep, $r_t$, with the estimated return at subsequent timesteps, $\max_a \hat{Q}_{\theta'}^\pi(s_{t+1}, a, \goal)$, discounted by $\gamma$.
Although the target relies on the value function approximator itself (this is the sense in which it is bootstrapped), it can nonetheless be used to incrementally improve the approximator.

In practice, it is advantageous to optimize a generalized target, the $\lambda$-return \cite{sutton18}:
\begin{align}
    G^\lambda_t = (1-\lambda)\sum_{n=1}^\infty \lambda^{n-1} \left[ r_t + \gamma r_{t+1} + \cdots \gamma^n \max_a \hat{Q}_{\theta'}^\pi(s_{t+n}, a, \goal) \right]
    \label{eq:lambdareturn}
\end{align}
where $\lambda \in [0,1]$ is a parameter that controls the exponential weighting of different $n$-step returns (the term in brackets).
Setting $\lambda=0$ recovers the original 1-step return. All settings of $\lambda$ are valid targets for estimating the value function, but they realize different bias-variance trade-offs. In large state spaces where goals may be many steps away, high $\lambda$ is critical: the 1-step return propagates reward information only one step per update, requiring many passes before signal reaches early states, whereas multi-step returns propagate information across the full trajectory in a single update. We use $\lambda=0.9$, which achieves good results for our applications. Using the $\lambda$-return as a target, the complete loss function becomes:
\begin{align}\label{eq:qlambda}
    L_{\tt Q\lambda}(\tau, \goal) = \sum^{T-1}_{t=1}\Big(
        G^\lambda_t - \hat{Q}_\theta^\pi(s_t, a_t, \goal)
    \Big)^2,
\end{align}
where the single-transition loss is now averaged over all transitions in the replay buffer.

To map the value function to actions, we use a simple $\epsilon$-greedy policy, which chooses $a_t = \argmax_a \hat{Q}_\theta^\pi(s_t, a, \goal)$ with probability $1-\epsilon$ or a random action with probability $\epsilon$. This ensures that the agent explores actions which are currently low-value but might turn out to be high-value with more experience. Much more sophisticated exploration policies have been developed, but our focus here is on learning algorithms, which is why we deliberately adopted the simplest possible policy.

\subsubsection*{Dyna}

Q-learning relies purely on experienced trajectories for value estimation. If the agent additionally has access to a model of the environment, $\hat{P}$, it can synthesize (``preplay'') data for improving its estimates (provided the model is reasonably accurate). This is the essential idea underlying the Dyna architecture \cite{sutton1990integrated}, summarized in \figref{fig:algorithm-comparison}.

In our implementation of Dyna, synthetic data $\tau_{\tt sim}$ is sampled from the environment model by running the model forward from observations in the replay buffer. The action selection policy is the same as the one used to generate real actions. This is repeated to produce $n_{\tt sim}$ trajectories stored in a preplay buffer. The synthetic data are treated in the same way as real data, entering into the TD loss (\eqref{eq:qlambda}).

\subsubsection*{Multitask Preplay}

\begin{figure*}[!htbp]
    \centering
    $$
    \begin{array}{ll}
    \hline
    \textbf{Dyna} & \textbf{Multitask Preplay} \\
    \hline
    \text{Sample trajectory } \tau \text{ and goal } \goal \text{ via replay} & \text{Sample trajectory } \tau \text{ and goal } \goal \text{ via replay} \\
    & \\
    \textit{// On-task simulation } & \textit{// Preplay} \\
    \text{For } t \in (1, \ldots, T) & \text{For } t \in (1, \ldots, T) \\
     \quad \text{For }i \in (1,\ldots,n_{\tt{sim}})
     & \quad \text{For }i \in (1,\ldots,n_{\tt{preplay}}) \\
     & \quad \quad \text{For }j \in (1,\ldots,n_{\tt goals})\\
     & \quad \quad \quad \text{Sample counterfactual goal: }\offtask \sim p_{\tt sim}(\cdot|s_t) \\
     \quad \quad \text{Sample trajectory } \tau_{\tt sim} \text{ for } \goal \text{ starting at } s_t
     & \quad \quad \quad \text{Sample trajectory } \tau_{\tt sim} \text{ for } \offtask \text{ starting at } s_t\\
    \quad \quad {\color{black} \text{Update loss: } L \leftarrow L +L_{\tt Q\lambda}(\tau_{\tt sim}, \goal)}
    & \quad\quad\quad \color{black} \text{Update loss: } L \leftarrow L + L_{\tt OQ\lambda}(\tau_{\tt sim}, \goal) \\
    & \quad\quad\quad \color{black} \text{Update loss}: L \leftarrow L + L_{\tt OQ\lambda}(\tau_{\tt sim}, \offtask) \\
    & \\
    \textit{// Single-goal learning} & \textit{// Conservative all-goals learning} \\
    \text{Evaluate loss: }L \leftarrow L_{\tt Q\lambda}(\tau, \goal) & \text{For }j \in (1,\ldots,n_{\tt goals})\\
    & \quad \text{Sample goal: }\offtask \sim p_{\tt all}(\cdot|s_t) \\
    & \quad \text{Update loss: } L \leftarrow L + L_{\tt COQ\lambda}(\tau, \offtask) \\
    \text{Update parameters: }\theta \leftarrow \theta - \eta \nabla_{\theta} L & \text{Update parameters: }\theta \leftarrow \theta - \eta \nabla_{\theta} L
    \\
    \hline
    \end{array}
    $$
    \caption{Comparison between Dyna and Multitask Preplay.
    In our experiments, $n_{\tt sim} = n_{\tt goals}\cdot n_{\tt preplay}$, so that Dyna and Multitask Preplay have the exact same simulation budget. The difference is in how simulations are chosen and how they are used for learning. Note that the goal distribution $p_{\tt sim}(\cdot|s_t)$ includes the on-task goal $\goal$, so Multitask Preplay subsumes standard Dyna-style on-task simulation as a special case.
    }
    \label{fig:algorithm-comparison}
\end{figure*}

Multitask Preplay (summarized in \figref{fig:algorithm-comparison}) is a multitask variant of Dyna designed to leverage each collected trajectory for learning across the full range of an agent's goals, not just the one being pursued. The central idea builds on the distinction between \emph{on-task} and \emph{off-task} learning \cite{borsa2018universal}, which parallels the classical RL distinction between on-policy and off-policy learning. Off-policy learning uses data from one policy to improve another, requiring corrections because the behavior policy differs from the target. Off-task learning uses data collected for one goal to learn about a different goal, requiring corrections because trajectories for a data-generating goal often diverge from optimal trajectories for other goals.

Multitask Preplay exploits off-task learning through two mechanisms, \emph{preplay} and \emph{conservative all-goals learning}. Preplay augments trajectories with counterfactual simulations of other goals, expanding the set of goals each trajectory can inform. Conservative all-goals learning reuses collected trajectories directly to update value estimates for goals beyond the one being pursued. Together, the two mechanisms allow every trajectory to contribute learning signal for all known goals. Both require off-task corrections, and our key insight is that existing off-policy machinery---specifically Peng's $Q(\lambda)$ \cite{peng1994incremental}---can be repurposed for this purpose.

The challenge is that multi-step returns yoke the trace to the actions actually taken. High $\lambda$ is critical for propagating reward across long trajectories, but when those actions were chosen to optimize $\goal$, the $\lambda$-weighted return accumulates divergence from what is optimal for $\offtask$ step by step. The $\lambda$-return (\eqref{eq:lambdareturn}) makes this concrete once rewritten recursively:
\begin{align}\label{eq:qlambdarecursive}
    G^\lambda_t = r_{t+1} + \gamma \left[\lambda\, G^\lambda_{t+1} + (1 - \lambda)\max_{a'} \hat{Q}_{\theta'}^\pi(s_{t+1}, a', \goal)\right],
\end{align}
with $G^\lambda_T = \max_{a'} \hat{Q}_{\theta'}^\pi(s_T, a', \goal)$ at the terminal step. Here $\lambda$ acts as a per-step trace gate, which suggests a natural correction: modulate $\lambda_t$ according to whether the collected action agrees with what the off-task goal would prescribe. We call this \emph{off-task $Q(\lambda)$}, repurposing the trace-cutting mechanism of Peng's $Q(\lambda)$ \cite{peng1994incremental} from the policy axis to the goal axis:
\begin{align}\label{eq:penglambda}
    \lambda_t = \begin{cases} \lambda & \text{if } \goal = \offtask \\ \lambda \cdot \mathbf{1}\!\left[\argmax_{a'} \hat{Q}_\theta^\pi(s_t, a', \offtask) = a_t\right] & \text{if } \goal \neq \offtask \end{cases}.
\end{align}
When goals agree, the full multi-step trace flows through and the correction reduces to standard $Q(\lambda)$. When they disagree, the trace is cut and the return falls back to a bootstrapped estimate. The off-task target $G^{\tt off}_t$ is obtained by substituting $\lambda_t$ for the fixed $\lambda$ in (\eqref{eq:qlambdarecursive}), yielding:
\begin{align}\label{eq:opqlambda}
    L_{\tt OQ\lambda}(\tau, \offtask) = \sum^{T-1}_{t=1}\Big(
        G^{\tt off}_t - \hat{Q}_\theta^\pi(s_t, a_t, \offtask)
    \Big)^2.
\end{align}

With this correction in hand, preplay proceeds as follows. At each experienced state $s_t$, the agent samples $n_{\tt goals}$ counterfactual goals $G = \{g_i \sim p_{\tt sim}(\cdot|s_t)\}^{n_{\tt goals}}_{i=1}$ uniformly from accessible goals, then generates simulated trajectories as in Dyna---but under a mixed policy that interpolates between the current and counterfactual goals:
\begin{align}
    a_t = \argmax_a \alpha_\goal\hat{Q}_\theta^\pi(s_t, a, \goal) + \alpha_\offtask\hat{Q}_\theta^\pi(s_t, a, \offtask),
\end{align}
where $\alpha_\goal$ and $\alpha_\offtask$ weight the two goals. The interpolation biases simulation toward the counterfactual goal while allowing the trajectory to revert once that goal is completed. Each rollout yields learning signal for both $\offtask$ and the original goal $\goal$ via off-task $Q(\lambda)$.

\begin{figure}[!htbp]
    \centering
    \includegraphics[width=0.7\columnwidth]{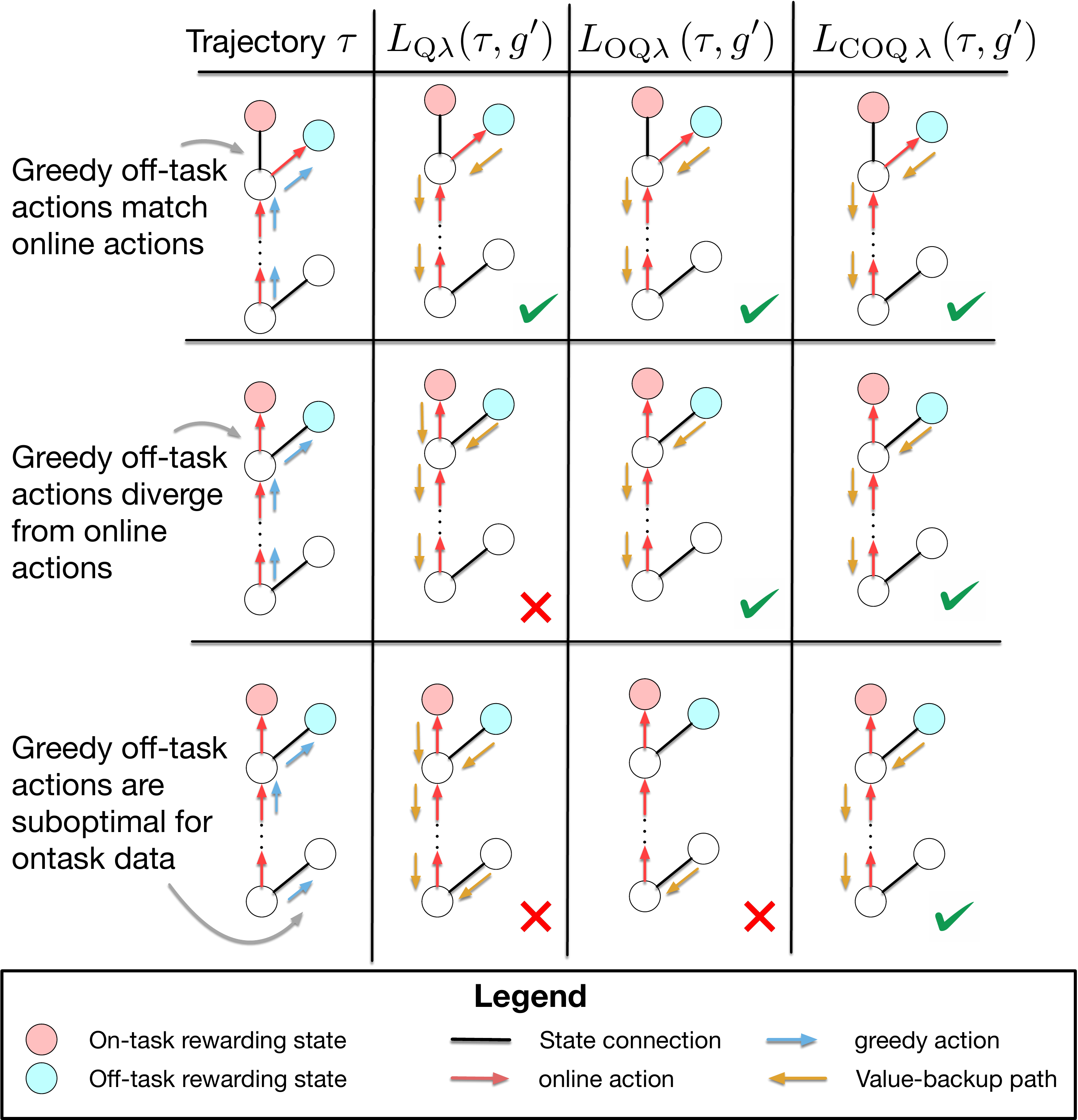}
    \caption{Comparison of value-backup strategies for off-task learning. Each row depicts a trajectory (left) and the value-backup paths under $Q(\lambda)$ ($L_{\tt Q\lambda}$), off-task $Q(\lambda)$ ($L_{\tt OQ\lambda}$), and conservative off-task $Q(\lambda)$ ($L_{\tt COQ\lambda}$) (right three columns). \textbf{Top:} When the greedy actions for the off-task goal match the collected actions, all three losses correctly propagate value from the rewarding state. \textbf{Middle:} When those greedy actions diverge from the collected actions, $Q(\lambda)$ erroneously backs up information from the on-task reward, which may be zero; off-task $Q(\lambda)$ and its conservative variant cut the trace and propagate only from the off-task rewarding state. \textbf{Bottom:} When the $Q$-value of an unseen action is overestimated, $Q(\lambda)$ backs up value from both the true reward and that inflated $Q$-value, while off-task $Q(\lambda)$ backs up from the inflated $Q$-value alone---compounding the error. Only the conservative variant, which penalizes $Q$-values for actions absent from the data, suppresses the inflated estimate while still propagating from the true reward.}
    \label{fig:value-backprop}
\end{figure}

Conservative all-goals learning addresses a complementary problem. Rather than simulating new trajectories, it reuses each collected trajectory $\tau$ directly, updating $Q$-values for goals $\offtask$ sampled from a distribution $p_{\tt all}(\cdot|s_t)$ over all known goals. Off-task $Q(\lambda)$ handles the trace correction, but a subtler issue remains: the greedy action for $\offtask$ may be suboptimal for $\goal$, so the agent selects it less and less during on-task collection. Any overestimation of that action's $Q$-value becomes self-reinforcing---never tested, never corrected. We prevent this with a conservative regularizer that penalizes $Q$-values for actions absent from the collected data \cite{kumar2020conservative}:
\begin{align}\label{eq:cl}
    L_{\tt CL}(\tau, \offtask) = \sum^{T-1}_{t=1}\left[\log \sum_a \exp \hat{Q}_\theta^\pi(s_t, a, \offtask) - \hat{Q}_\theta^\pi(s_t, a_t, \offtask)\right].
\end{align}
The complete loss, \textit{Conservative Off-task $Q(\lambda)$}, combines off-task $Q(\lambda)$ with the regularizer\footnote{A similar loss was concurrently developed outside of the all-goals setting by \cite{kimpeng} for conservative value estimation during offline RL.}:
\begin{align}\label{eq:cpqlambda}
    L_{\tt COQ\lambda}(\tau, \offtask) = L_{\tt OQ\lambda}(\tau, \offtask) + \eta_C \cdot \mathbf{1}[\offtask \neq \goal] \cdot L_{\tt CL}(\tau, \offtask).
\end{align}
\figref{fig:value-backprop} illustrates how $L_{\tt Q\lambda}$, $L_{\tt OQ\lambda}$, and $L_{\tt COQ\lambda}$ differ in their value-backup behavior when learning off-task $Q$-values and \extfigref{ext:allgoals-ablation} ablates equations \eqref{eq:penglambda}, \eqref{eq:cl}, and \eqref{eq:cpqlambda}.

\subsubsection*{Universal Successor Features}

Successor features (SFs) are a form of predictive representation that generalize the standard reward-predictive representation (\eqref{eq:q-value}) used in RL to arbitrary feature-predictive representations \cite{carvalho2024predictive}. The key assumption of SFs is that reward for goal $\goal$ can be expressed as a dot-product between state-features $\phi(s)$ and a ``preference vector'' $w_{\goal}$ for those features $R_\goal(s) = \phi(s)^{\top} w_{\goal}$, where $\phi(s),w_{\goal} \in \mathbb{R}^{d}$.
SFs are defined as the predictions of the expected discounted accumulated features $\phi$ under policy $\pi$:
\begin{equation}\label{eq:sf}
    \psi^{\pi}(s,a) = \mathbb{E}\left[\sum^{\infty}_{t=0} \gamma^t \phi(s_{t+1}) | s_0=s, a \sim \pi \right].
\end{equation}
Action-values can be expressed as a dot-product between SFs and a corresponding preference vector: $Q^{\pi}(s,a,\goal) = \psi^{\pi}(s,a)^{\top}w_{\goal}$.
To facilitate generalization across tasks and policies, we will consider \textit{Universal} Successor Feature Approximators \cite{borsa2018universal}, which are the analogue of goal-conditioned value functions described above where the function approximator takes some reference to the policy for which it is estimating SFs, such as a task parameter $\goal$, as input $\hat{\psi}_\theta^\pi(s,a,\goal)$.

We assume that the features $\phi(s)$ are additionally defined by encounters with landmark objects. In fully observable domains such as JaxMaze (\figref{fig:jaxmaze}), $\phi_k(s) = 1$ every time landmark $k$ is within some radius of the agent (0 otherwise). In partially observable domains such as Craftax (\figref{fig:crafter}), landmark features activate when the object is visible to the agent.

The structure of the learning algorithm is essentially the same as for Q-learning, with the value function replaced by SFs. An analogous $\lambda$-return can be defined for SFs (again replacing rewards with features). Parameters of the function approximator are likewise optimized to the squared difference between the empirical and predicted $\lambda$-returns.

As with Multitask Preplay, we assume that the algorithm knows about all accessible goals within a domain $G=\{g_i\}^{n_{\tt train} + n_{\tt test}}_{i=1}$, along with their associated preference vectors. Thus, a trajectory $\tau$ collected for goal $\goal$ could be used to learn from all goals by randomly sampling goals and optimizing the aggregated loss. While in principle this endows the algorithm with the same generalization capabilities as Multitask Preplay, it critically depends on the ability to \emph{directly experience} many different tasks, so that all the heavy lifting can be done by a purely model-free learning algorithm. The premise of this paper is that agents in the real world rarely have access to such abundance of direct experience.

SFs also enable another form of generalization through the reuse of previous policies for a new task. The key algorithmic tool is \emph{generalized policy iteration} (GPI)\cite{barreto2018transfer}, which selects actions for a new task $\newtask$ according to:
\begin{align}
    a(s, \newtask) = \argmax_{a\in\mathcal{A}} \max_{w \in \mathcal{C}} \{\hat{\psi}^\pi_\theta(s,a,w)^{\top}w_{\newtask}\},
\end{align}
If $\mathcal{C}=\mathcal{W}_{\tt train}$, then this corresponds to seeing if any training policy leads to features specified by $w_{\newtask}$. However, if we learned $\hat{\psi}_\theta(s,a,w_{\newtask})$ with \textit{counterfactual task learning} (i.e. collected data with $\mathcal{W}_{\tt train}$ but computed learning updates for $w_{\newtask}$), we can see if $\hat{\psi}_\theta(s,a,w_{\newtask})$ does not already contain landmark and task-completion knowledge that optimizes $w_{\newtask}$. Two options for using this knowledge is to set $\mathcal{C}=\mathcal{W}_{\tt train} \cup \{w_{\newtask}\}$ or $\mathcal{C}=\{w_{\newtask}\}$.

\subsubsection*{Hindsight Experience Replay}

Hindsight experience replay (HER) \cite{andrychowicz2017hindsight} enables off-task learning without an environment model by relabeling collected trajectories with goals achieved along the way. When an agent collects a trajectory $\tau$ while pursuing goal $\goal$, HER identifies hindsight goals from states achieved in the trajectory and replays sub-trajectories with these substituted goals, optimizing the standard $Q(\lambda)$ loss (\eqref{eq:qlambda}). This relabeling transforms failures into training signal: a trajectory that failed to reach $\goal$ may have achieved other goals along the way. HER is primarily an exploration algorithm---by building $Q$-values for incidentally achieved goals, it constructs a richer value landscape that facilitates reaching the goal of interest. However, HER can only learn about goals $\offtask$ that lie along the path the agent has taken. In large state spaces, where trajectories cover a diminishing fraction of the goal space, this limits HER's ability to propagate value information beyond directly visited goals---the same limitation faced by USFA.

\subsubsection*{HER with Conservative All-Goals Learning}

TD bootstrapping can propagate value to goals absent from the current trajectory, suggesting a direct extension: augmenting HER with the conservative all-goals loss $L_{\tt COQ\lambda}$. For each trajectory $\tau$, goals $\offtask$ are sampled from $p_{\tt all}(\cdot|s_t)$, and the off-task $Q(\lambda)$ correction handles the goal mismatch while the conservative regularizer prevents overestimation. This extension enables HER to scale to large environments where standard HER plateaus, provided the evaluation goals lie along paths the agent has previously traversed (see \extfigref{ext:her-comparison}). When goals are not along any experienced path, HER with all-goals learning faces the same limitation as USFA: without direct experience of relevant states, purely model-free methods cannot construct informative value estimates.

\subsection*{Implementation details}
We build our model closely following the Recurrent Replay Q-network architecture~\cite{kapturowski2018recurrent} from the ACME codebase~\cite{hoffman2020acme}.
We begin with default hyperparameters from ACME and change them to achieve optimal training task performance. Optimal training performance is readily attainable across all tasks we study.
We base our Jax implementation on PureJaxRL~\cite{lu2022discovered}.
We summarize the network architecture used by Q-learning, Dyna, and Multitask Preplay in~\extfigref{ext:q-network}.
We present shared hyperparameters across all models in Table~\ref{tab:shared_hyperparameters} and model-specific parameters in Table~\ref{tab:model_specific_hyperparameters}.

\textbf{Q-Networks.}
\textit{JaxMaze}. Each cell of the grid carries one of $D=6$ categorical labels (wall, empty, and the four object types that serve as goals). We embed each label into a 64-dimensional vector using the standard word-embedding architecture from NLP~\cite{mikolov2013linguistic,vaswani2017attention}; here the ``words'' are cell types, so the embedding lets the network share representations across grid positions that contain the same object. We use only the \textit{initialization} scheme from that literature---the embedding matrix is trained from scratch via the RL loss, not loaded from any pretrained model. The per-cell embeddings are flattened into a $H \times W \times 64$-dimensional vector. The goal is a one-hot vector over the $C=4$ goal-object types, processed through a 128-unit linear layer. We concatenate the observation and goal embeddings and give them as input to a 256-unit Long short-term memory (LSTM)~\cite{hochreiter1997long}.
We leverage an LSTM, because it facilitates passing gradients through long time-series data such as our replay data. This dramatically improves learning of recurrent Q-networks.
The output of the LSTM is concatenated with the goal embedding and passed through a 1024-unit MLP with $2$ hidden layers and outputs $A$ action values.
All layers use ReLU activations and omit bias terms.
\textit{Craftax}. Craftax already provides a $56$-dimensional continuous observation (normalized to $[0,1]$) that summarizes the agent's local view and inventory, so no embedding step is needed---we pass the observation directly into the LSTM alongside the goal encoding. The goal dimensionality depends on the variant: $C=3$ for the goal-conditioned Craftax used in the human experiments (three goal objects) and $C=72$ for the standard Craftax achievement space used in the AI experiments (one dimension per achievement). We do not omit bias terms in this architecture.

\textbf{Successor Feature Networks.}
These networks share the Q-network architecture but output $A \times C$ successor feature values instead of $A$ Q-values. Action values are computed via per-action dot products with the goal vector.

\textbf{Training.}
We trained all models with a protocol designed to mirror the structure of our human experiments, so that no model enters the test phase with an unfair informational disadvantage. For each JaxMaze experiment, we train on rotations of the same mazes used with human subjects (\extfigref{ext:jaxmaze-envs} and \extfigref{ext:jaxmaze-juncture-envs}), and we swap which objects serve as train and test objects across runs so that every object serves as a train object at least once. We additionally train jointly on the two-paths experiment, the shortcut experiment, and a practice environment in which the agent must simply obtain all four objects within 10 timesteps. Thus every model has direct experience attempting to obtain the test-task objects, and this is reflected in test-task success being well above zero for USFA and HER (\extfigref{ext:jaxmaze_train} and \extfigref{ext:craftax_train_test}) despite the difficulty of our long-horizon tasks. For Craftax, each block uses a distinct procedurally generated map (\extfigref{ext:craftax-envs}), and the same train/test-object swapping applies.

In JaxMaze, all models use $N=32$ parallel environments. In Craftax, model-free methods (Q-learning, USFA, HER) use $N=128$ parallel environments, while model-based methods (Dyna, Multitask Preplay) use $N=32$ parallel environments.
Models begin performing gradient updates after 10,000 data points are in the replay buffer. Model-free methods update every $160$ environment steps across both domains. Dyna and Multitask Preplay update every $160$ environment steps in JaxMaze and every $32$ steps in Craftax; we tried updating model-free methods more frequently in Craftax but found this led to degraded performance. All models select actions with $\epsilon$-greedy exploration, where each parallel environment is assigned its own $\epsilon$ that is held constant throughout training. In JaxMaze, $\epsilon$ values take the form $0.1^{x}$ with $x$ linearly spaced on $[1, 3]$, placing $\epsilon$ in $[10^{-3}, 10^{-1}]$ with a bias toward the low end. In Craftax, longer episodes required wider exploration, so $\epsilon$ values take the form $0.1^{x}$ with $x$ linearly spaced on $[0.05, 0.9]$, placing $\epsilon$ in $[{\approx}0.13, {\approx}0.89]$.

\textbf{Experience Replay.}
We employ prioritized experience replay~\cite{schaul2015prioritized}, which samples training transitions based on their TD error magnitude rather than uniformly.
This allows agents to learn high-error experiences more quickly.
The importance sampling exponent (we use $0.6$) moderately corrects for the bias introduced by non-uniform sampling, while the maximum priority weight (we use $0.9$) prevents any single high-priority transition from dominating gradient updates.
We use batch sizes with $N=32$ trajectories of length $T=40$ for both JaxMaze and Craftax. 
One difference is that we find longer trajectories of length $T=80$ improve performance of model-free methods in Craftax.

\textbf{Counterfactual simulation.} 
Both Dyna and Multitask Preplay use oracle world models. The key difference between real data and simulated data is that simulated data must begin from some starting ``replayed'' state and can only do a finite number of simulation steps.
We set this to $15$ for JaxMaze and $20$ for Craftax.
We have simulation policies use epsilon values distributed logarithmically between $1$ and $3$ using a base of $0.1$, which effectively keeps values in the range of $[0, .1]$ with a bias for smaller values.

\textbf{Optimization.}
Temporal Difference learning methods use their own predictions when constructing prediction targets. This means that as a model learns, its prediction target is changing, which can make learning unstable~\cite{mnih2015human}. 
To mitigate this, we employ ``target networks'', where we have a second set of parameters that are used to compute prediction targets.
These parameters are updated to the latest parameters every $1000$ learning updates.
We use Adam optimizer~\cite{kingma2014adam} with default values except changing the epsilon value to $10^{-5}$. This is the default in ACME and increases numerical stability by preventing division by very small numbers in the denominator of the parameter updates. 
We use learning rates of $0.001$ for model-free methods and $0.0003$ for model-based methods in JaxMaze; all methods use a learning rate of $0.0003$ in Craftax.
To prevent updates with gradients that are too large, we clip gradients to have a maximum global norm (across all parameters) of $80$.

\subsection*{Hyperparameter search}
All searches were random searches. For each sweep we drew $50$ trials, and each trial sampled one value independently from every axis's range. Training-task performance selected the winner. Sweeps followed an inheritance chain across algorithms: each sweep began from the nearest parent configuration, and any axis not re-searched kept its parent's value.

\subsubsection*{JaxMaze}
We based Q-learning on the Recurrent Replay Q-network implementation from ACME~\cite{kapturowski2018recurrent,hoffman2020acme}, replacing the default return estimator with $Q(\lambda)$~\cite{sutton2014qlambda} to better support the long horizons of our mazes. We searched the target-network update interval over $\{100, 200, 1000, 2000\}$; the learning rate over $\{10^{-2}, 3\times10^{-3}, 10^{-3}, 3\times10^{-4}\}$; the Adam numerical-stability constant over $\{10^{-4}, 10^{-5}, 10^{-6}, 10^{-7}, 10^{-8}\}$; the Q-network hidden dimension over $\{256, 512, 1024\}$; the number of Q-network hidden layers over $\{1, 2, 3\}$; the discount factor over $\{0.99, 0.991, 0.992, 0.993\}$; and $\lambda$ over $\{0.3, 0.6, 0.7, 0.8, 0.9\}$.

USFA was initialized from Q-learning and inherited every choice except those specific to the successor-feature head. We re-searched the learning rate, the successor-feature hidden dimension, the number of successor-feature hidden layers, and the discount factor over the same grids.

HER was also initialized from Q-learning, and most axes transferred. The critical addition was conservative all-goals learning, which our ablation shows is necessary to generalize on the larger mazes (\extfigref{ext:her-comparison}). Following standard practice we re-searched the learning rate, and we additionally searched the conservative Q-learning penalty coefficient over $\{0, 1, 10^{-1}, 10^{-2}, 10^{-3}, 10^{-4}, 10^{-5}\}$.

Dyna was initialized from Q-learning and uses the same per-environment $\epsilon$ values for simulation as for online acting. We re-searched the learning rate, Q-network hidden dimension, number of Q-network layers, and discount factor over the same grids.

Multitask Preplay was initialized from Dyna. We searched the conservative Q-learning penalty coefficient and the conservative all-goals coefficient, each over $\{1, 10^{-1}, 10^{-2}, 10^{-3}, 10^{-4}, 10^{-5}\}$.

\subsubsection*{Goal-conditioned Craftax}
Each algorithm began from its JaxMaze counterpart, with two changes that cut across every sweep. First, we re-searched the scaling axes that matter more in Craftax: Q-network hidden dimension over $\{256, 512, 1024\}$, number of hidden layers over $\{1, 2, 3\}$, number of parallel environments over $\{32, 64, 128\}$, trajectory length over $\{40, 80\}$, and batch size over $\{32, 64, 128\}$. Increasing hidden dimension, depth, parallel environments, and trajectory length each contributed substantially. Second, we replaced the JaxMaze per-environment $\epsilon$ scheme with the wider Craftax scheme described in the Implementation details.

USFA inherited the Q-learning improvements and we re-searched only the learning rate. HER inherited from its JaxMaze counterpart together with the Craftax Q-learning improvements; no further search was needed. Dyna inherited from Craftax Q-learning and additionally searched the simulation length over $\{10, 15, 20\}$, the Q-network hidden dimension, and the number of hidden layers. Multitask Preplay began from Dyna and searched the conservative Q-learning penalty coefficient over $\{1, 10^{-1}, 10^{-2}, 10^{-3}, 10^{-4}, 10^{-5}\}$.

\subsubsection*{Standard Craftax}
Standard Craftax shares its scaling regime with goal-conditioned Craftax, so most configurations transferred directly. Q-learning with the 1-step auxiliary task and Q-learning with the successor-feature auxiliary task each began from goal-conditioned Craftax Q-learning and searched the auxiliary-task coefficient over $\{10^{-1}, 10^{-2}, 10^{-3}, 10^{-4}, 10^{-5}, 10^{-6}\}$. HER, Dyna, and Multitask Preplay inherited their configurations from the goal-conditioned setting.

\subsection*{Environments}

\textbf{JaxMaze}.
\textit{Tasks} involve controlling a red triangle to obtain objects in the maze (see~Figure 3). The red triangle can move in any cardinal direction that is not blocked by a wall. When the red triangle moves over an object, it immediately collects it and the episode terminates.
\textit{Observations} of the environment contain the full map. Humans observe pixels and all models get symbolic representations of the entire map.
\textit{Actions}: Agents can move up, down, left, and right. 
\textit{Rewards}: When a task's goal object is collected, agents get a reward of $1$; all other timesteps get a reward of $0$.

\textbf{Goal-conditioned Craftax}. This domain is a variant of the original Craftax environment~\cite{matthews2024craftax}.
\textit{Tasks} involve obtaining diamonds, sapphires, or rubies. Once a stone is collected, the episode terminates.
\textit{Observations} of the environment contain either the full map or a zoomed-in view (see Figure 4). During training, human subjects observe the full map and the zoomed-in view; during testing, they only see the zoomed-in view. All models only see the zoomed-in view during both training and testing. Models get symbolic representations of the entire map.
\textit{Actions}: Agents can move up, down, left, right, or select a ``do'' action which will collect the object in front of an agent. 
\textit{Rewards}: When a task's goal object is collected, agents get a reward of $1$; all other timesteps get a reward of $0$.
To make the environment easier for agents to learn in 1 hour, we start agents with a large ``strength''---i.e. with the tools necessary to mine their goal object (e.g., a pickaxe), and made parts of the map that would otherwise need to be destroyed (such as non-precious stones) passable.

\subsection*{Experiment design}
All experiments were preregistered (see Data availability) and approved by the Harvard IRB.
All subjects were recruited with CloudResearch and provided informed consent.
Sample sizes were determined based on pilot experiments (see Reporting summary).
We implemented experiments using NiceWebRL\footnote{\url{https://github.com/wcarvalho/nicewebrl/}}.

All experiments used the following block design. An experiment consisted of $4$ blocks, where each block had an associated maze (see examples of environments and goal/start sampling in \extfigref{ext:jaxmaze-envs} and \extfigref{ext:jaxmaze-juncture-envs}).
For JaxMaze, each block within an experiment presented a rotation of the same maze. In Craftax, different blocks had different procedurally generated maps.
Blocks were divided into a training segment where subjects learned tasks from $\traintasks$ and a \textit{subsequent} evaluation segment where subjects were tested on tasks from $\testtasks$.
During training, subjects had $80$ attempts to successfully complete $8$ trials per train task. During evaluation, subjects had $1$ attempt for each evaluation task.
Subjects were given a bonus for completing evaluation tasks more quickly but there was no time-limit.
When a trial was completed, subjects had to press a key to move on to the next trial.
Across subjects, block ordering and object categories were randomized.
The order that tasks was presented was also randomized.

\textbf{JaxMaze: Two Paths experiment (experiment 1) and Shortcut experiment (experiment 2)}. Both experiments followed the same protocol but differed in the mazes presented to subjects. Individual blocks consisted of $2$ training tasks and $1$ evaluation task. During training, subjects were told what task they would later be evaluated on and their starting location was chosen from $8$ possible locations. Subjects were told that they would receive a bonus for completing the evaluation task in under $20$ seconds.

\textbf{JaxMaze: Intermediary start experiment (experiment 3) details.} Individual blocks consisted of $2$ training tasks and $2$ evaluation tasks. During training, subjects were told what task they would later be evaluated on. Subjects were told they would get a bonus for completing evaluation tasks in under $20$ seconds.

\textbf{JaxMaze: Juncture experiments (experiment 4)}. Individual blocks consisted of $1$ training task and $2$ evaluation tasks. During training, subject's starting location was chosen from $8$ possible locations. Across all experiments, subjects were spawned in a novel starting location during evaluation. We tested $3$ conditions in these experiments: (1) the test object is nearby a training path and known during training ($\tt{near,know}$;~Figure 3 E), (2) the test object is nearby a training path and not known during training ($\tt{near,unknown}$;~Figure 3 F), and one where the test object is ``far'' a training path and known during training ($\tt{far,know}$;~Figure 3 G). For each condition, there were 2 evaluation tasks which had the same optimal path as a solution---but in different parts of the map. For the condition when an object was nearby but unknown, we ensured that subjects could not guess which object they would be evaluated on by having two potential goal objects on opposite sides from the spawning location. When the test object was nearby, subjects were told they would get a bonus for completing evaluation tasks in under $3$ seconds; when it was further away, we changed this to $10$ seconds.

\textbf{Craftax experiment (experiment 5)}.
Individual blocks had a unique map (\extfigref{ext:craftax-envs}), where we placed $2$ train objects in distinct locations and then a test object visibly nearby $1$ of the train objects. Like in our JaxMaze experiments, we chose start and end positions such that there were two paths to the test object---where one overlapped with the optimal path to the train object and one that did not. To get maps with this property, we procedurally generated 100 worlds and manually chose from this set. To create this gap, we exploited that subjects cannot pass through lava or water and thus would have to go around, creating two paths to objects.
Blocks were divided into a training segment where subjects learned $2$ training tasks from $\traintasks$ and a \textit{subsequent} evaluation segment where subjects were tested on $1$ evaluation task from $\testtasks$.
We ran two conditions: one where subjects were told what task they would later be evaluated on and one where they were not. In each training episode, their starting location was chosen from $4$ possible locations. Subjects were told they would get a bonus for completing evaluation tasks in under $20$ seconds.

\subsection*{Statistical Analyses}
\textbf{Path reuse (experiments 1, 2, and 5)}.
To test whether people demonstrated above chance evidence of partial path reuse, we first computed the proportion of trials in which each subject reused the prior training path at evaluation. We note that the exact functions for computing path reuse for JaxMaze and Craftax was not preregistered, though the statistical tests we used were.
\textit{JaxMaze} We define path reuse as follows. We have two binary matrices $M_{\tt train}$ and $M_{\tt test}$ of the same dimension $H \times W$ that have $M(i,j)=1$ if an agent was in position $i,j$.
We compute path overlap $o$ of how much a participant's test path overlaps with their training path by computing how much of the path is shared, divided by the length of the test path, i.e.
$o_{\tt map} = \frac{M_{\tt train}\odot M_{\tt test}}{||M_{\tt test}||}$, where $\odot$ is an element-wise product and $||M||$ defines the norm of matrix $M$. Note that $o_{\tt map} \in (-1,1)$.
Path reuse is then $p_{\tt reuse} = o_{\tt map} > \alpha_{\tt map}$ where $\alpha_{\tt map}$ is a threshold. We set $\alpha_{\tt map}=.5$ for prediction 1 and $\alpha_{\tt map}=.7$ for prediction 2.

For prediction 1 our conclusions are robust to the choice of $\alpha_{\tt map}$ (\extfigref{ext:jaxmaze-threshold-robustness} and \extfigref{ext:jaxmaze-threshold-robustness-rt}), because there are essentially only two paths to the test object. For prediction 2, shortcuts open up many more possible routes, so the choice of threshold matters more. We settled on $\alpha_{\tt map}=.8$: test paths that partially took a new shortcut fall below this value (\extfigref{ext:jaxmaze-shortcut-summary}, top row), while overlap above $.8$ indicates that the subject took no shortcut at all, i.e. completely reused the prior path (\extfigref{ext:jaxmaze-shortcut-summary}, bottom row).

\textit{Craftax} is a more open-ended environment where many paths can be taken between two points. To account for this, we have $M_{\tt train}(i,j)=1$ for all points along a person's train path or points $1$ position away (i.e. $i\pm1$ or $j\pm1$). It is possible for there to be a relatively high overlap but for the participant to come in from a different angle, so we additionally measure the cosine similarity for the last $10$ positions of a person's path. Let $p$ be a vector of positions a person has taken. Then we compute direction overlap via cosine similarity as $o_{\tt dir} = \frac{v_{\tt train}^{\top} v_{\tt train}}{||v_{\tt train}||\cdot||v_{\tt test}||}$ where $v = \frac{1}{10}\sum_{t=T-11}^{T-1} (p_{t+1} - p_t)$ and $x^\top y$ is a dot-product (we repeat this for both train and test). Note that $o_{\tt dir} \in (-1,1)$.
We had path reuse be $1$ if there was a sufficient overlap for both measures or if map path reuse was very high.
That is, path reuse is then $p_{\tt reuse} = \left(o_{\tt map} > \alpha_{\tt map} \land o_{\tt dir} > \alpha_{\tt dir}\right) \lor \left(o_{\tt map} > 2\cdot\alpha_{\tt map}\right)$ where $\alpha_{\tt map}$ is a direction threshold. We used $\alpha_{\tt map}=.25$ and $\alpha_{\tt dir}=.5$.

To build intuition for this criterion, Figures~\ref{ext:craftax-overlap-tp-direction-vs-overlap}--\ref{ext:craftax-overlap-tn-vs-fn} show representative evaluation trials. Sometimes a subject follows the training path but diverges slightly, so map overlap stays modest (e.g.\ $o_{\tt map}=.316$ in \extfigref{ext:craftax-overlap-tp-direction-vs-overlap}a) even though the angle of approach to the object is preserved across train and test; the direction clause recovers these cases. In other cases the approach is nearly perpendicular across train and test, yet the paths overlap heavily (\extfigref{ext:craftax-overlap-tp-direction-vs-overlap}b), and the high-overlap clause fires. \extfigref{ext:craftax-overlap-tp-both-vs-fp} contrasts a trial where both clauses are satisfied with a false positive. \extfigref{ext:craftax-overlap-tn-vs-fn} shows a true negative alongside a near-miss false negative, in which the subject broadly reuses the training path but falls short on both overlap and approach angle.

One way to check the geometric criterion is to replace it with a method that shares none of its assumptions. We re-classified every evaluation trial with a vision-based language model (Claude Sonnet 4.5), which inspected side-by-side renderings of the train and test paths (\extfigref{ext:claude-path-reuse-prompt}). The model is more conservative --- $63.5\%$ reuse when the goal was known and $63.7\%$ when it was unknown, against roughly $70\%$ under the geometric criterion on the same cohort ($n=359$ and $n=344$ trials, $100$ subjects per condition). Critically, the two methods' $95\%$ confidence intervals overlap in both conditions (\extfigref{ext:craftax-path-reuse-method-comparison}), placing the two estimates within sampling error of one another.

We then tested whether the resulting distribution was normally distributed using a Shapiro-Wilk test. As the data were not normally distributed in all cases (all W-values $>$ 2964 and all p-values $< 9.4\times10^{-9}$ for JaxMaze; all W-values $> 2966$ and all p-values $< 1.3\times10^{-10}$ for Craftax), we tested whether the proportion of evaluation trials with partial reuse was significantly greater than $0.5$, using non-parametric, one-sided Wilcoxon signed-rank tests.

\textbf{RTs during partial reuse (Figure 3 C\&D)}.
An episode of $T$ steps contains $T$ response time (RT) measurements. Each RT measurement corresponds to the time between when an image is seen by a subject to when they hit a key in response. 
RT measurements have a skewed distribution. A standard method to deal with this is to log-transform RTs so that they approximately satisfy Gaussian distributional assumptions~\citep{whelan2008effective,ulrich1993information}. Thus, we applied all statistics to log-transformed RT measurements.
In this analysis, we either look at the maximum RT value across all $T$ steps or the very first one.
To evaluate log RTs on reuse and non-reuse choices, we used linear mixed effects models (LMEs) with the following equation: $RT=B0 + B1 \times p_{\tt reuse} + B2 \times (1 | \text{subject ID}) + \epsilon$.
In this equation, B0 is a constant and $\epsilon$ is residual error.
The fixed effect (path score) is a binary variable indicating whether subjects partially reused a training path (1) or followed a new path (0) on a given evaluation trial.
The random effect $(1 | \text{subject ID})$ models random intercepts for each subject.
LMEs were used in these cases because the path score measure was choice dependent, as opposed to being manipulated experimentally.
A significant B1 coefficient with a negative sign indicates significantly lower RTs when subjects partially reused the training path.

\textbf{RTs during juncture and starting point manipulations (Figure 3 H\&J)}.
To test whether people were faster when placed along the training path at evaluation, we first computed the log RT for the training junction and non-training junction evaluation trials.
We then took the difference in log RT between these evaluation trials for each subject and tested whether the resulting distribution was normally distributed using a Shapiro-Wilk test.
As the data were not normally distributed, we used one-sided Wilcoxon signed-rank tests to test whether subjects were significantly faster when placed along the training path at evaluation.
An equivalent statistical analysis was used to compare RTs for on-path and off-path evaluation trials.

\subsection*{Craftax Environment Transfer Details}
This is a procedurally generated environment where different initiation seeds spawn different environments. In the normal setup, all initiation seeds are used for training. To enable the study of transfer across environments, we followed a standard protocol in the literature~\cite{cobbe2019procgen,kuttler2020nethack} and designated $n_{\tt train}$ train initialization seeds and $n_{\tt test}$ test initialization seeds. For each evaluation we subsample a fixed set of seeds from the held-out test pool; the specific seeds are open-source and the subsampling is visible in the released analysis scripts on GitHub.
Craftax is a \textbf{single-task} environment where there is a \textbf{single global reward function} that gives different priorities to different goals. Specifically, for $m$ goals, $r_{\goal} = (\phi_{\tt achieved}(s_{t}) - \phi_{\tt achieved}(s_{t-1}))^{\top} w_{\goal}$ where $w_{\goal} \in \mathbb{R}^m$ is a vector that gives different preferences to different goals. We do not change this.
In this article, our goal is to show that \textit{preplay towards accessible goals in one environment improves generalization to a novel environment}. As such, we focus on the setting where agents knows about what goals exist and when they are available. For example, if you see a diamond, this becomes an available goal for you when you already have the necessary tool (i.e., a pickaxe). We used an LLM to create a function which goes through the logic of Craftax and automatically computes this, showing that it is possible to automate these kinds of computations today with modern language models. We expose this to the agent with a function $\phi_{\tt avail}(s) \in \mathbb{R}^{m}$, which is $1$ for a goal if that goal can be completed. Index $i$ of $\phi_{\tt avail}(s)$ corresponds to the availability of the goal specified by index $i$ of $\phi_{\tt achieved}(s)$. We experimented with also exposing subgoals as accessible only if they were visible and not if preconditions were met. Multitask Preplay still outperformed Dyna in this setting (see~\extfigref{ext:preplay_ablations}) but chose to focus our results in the setting where---as is common with people---an agent is aware of possible subgoals and whether they are currently possible. This is because our focus is on how this \textit{improves generalization to new environments}; not on how it improves learning in a training environment---something already shown in prior work~\cite{kulkarni2016hierarchical}.

To expose this knowledge to all agents, \textbf{observations} are defined as $x=(x_{\tt image}, \phi_{\tt avail}(s), \phi_{\tt achieved}(s))$ where $x_{\tt image}$ is a zoomed-in view of the agent in the map, $\phi_{\tt avail}(s)$ describes goals are achievable, and $\phi_{\tt achieved}(s)$ describes which goals have been achieved thus far. All agents get all of this information.
\textbf{Actions}. There are 74 actions which involve object-interaction actions like $\tt{PLACE\_PLANT}$ or $\tt{MAKE\_WOOD\_PICKAXE}$ which work successfully once relevant preconditions are met.

\paragraph{Model implementations}
We consider $4$ models that each take advantage of $\phi_{\tt avail}(s)$ in different ways. All models, including Dyna, use $\phi_{g}(s) = [\phi_{\tt avail}(s), \phi_{\tt ach}(s)]$ as input for their action-value predictions. 

\textbf{Q-learning + $1$-step model}. Here, we leverage $\phi_{g}(s)$ to define an auxiliary task where the agent learns to predict $\phi_{g}(s_{t+1})$. We use the same neural network as the Q-network architecture from the Craftax human experiments with the following change. We also produces $\hat{\phi}_g(s')$ with a 512-unit MLP that takes the LSTM state as input.

\textbf{Q-learning + SF}. Since this is a single-task environment, successor features can only be used as an auxiliary task and not as a method for transfer across tasks. Here, $\phi_{g}(\cdot)$ defines the features for successor features. Like with the previous baseline, we change the Craftax human experiments Q-network to additionally produce $\hat{\psi}^{\pi}(s, a, \goal)$ with a projection from the the LSTM state as input.

\textbf{Dyna}. Dyna uses the learned world model to generate simulations that optimize the main reward $R_\goal$. Unlike Multitask Preplay, Dyna does not sample subgoals---all simulated experience is used to improve the policy for the single task objective. We also ablate a variant where the simulation goal is randomly replaced with a uniformly sampled goal from $\phi_{\tt avail}(s)$ at varying rates. This tests whether naively introducing off-task simulations (without the structured all-goals learning of Multitask Preplay) can improve generalization (see~\extfigref{ext:dyna-multigoal}).

\textbf{Multitask Preplay}. Since this is a single task setting, we define the main goal as $\goal$ and then sample subgoals $\offtask$ during preplay. 
This algorithm works equivalently to in our previous experiments. The main difference is that we sample $\offtask \sim \operatorname{Uniform}(\phi_{\tt avail}(s))$. 
Note that a sampled $\offtask$ will often correspond to a goal that was \textit{not} completed in the real-life experience.
We learn a single action-value function for both $\goal$ and $\offtask$ and simply change which is input depending on the relevant computation.
This effectively allows Multitask Preplay to preplay simulations towards $\offtask$ and then update action-values for both $r_\goal$ and $r_\offtask$.
We find that this is key to maximizing the benefits of multitask preplay (see~\extfigref{ext:preplay_ablations}). 

\newpage

\clearpage
\begin{table}[!htp]
\centering
\caption{Shared hyperparameters across all models. These parameters are consistent across Q-learning, Successor Features (SF), Dyna, and Multitask Preplay algorithms.}
\label{tab:shared_hyperparameters}
\begin{tabular}{ll}
\hline
Hyperparameter & Value \\
\hline
Discount factor ($\gamma$) & 0.992 \\
Replay buffer capacity & 50,000 \\
Batch size & 32 \\
Gradient clipping & 80 \\
Activation function & Leaky ReLU \\
LSTM hidden size & 256 \\
MLP hidden dimensions & 1024 \\
\hline
\end{tabular}
\end{table}

\begin{table}[!htp]
\centering
\caption{Model-specific hyperparameters}
\label{tab:model_specific_hyperparameters}
\begin{tabular}{lccccc}
\hline
 & Q-learning & SF & HER & Dyna & Multitask Preplay \\
\hline
\multicolumn{6}{l}{\textit{Number of parallel environments}} \\
\quad JaxMaze & 32 & 32 & 32 & 32 & 32 \\
\quad Craftax & 128 & 128 & 128 & 32 & 32 \\
\multicolumn{6}{l}{\textit{Batch length}} \\
\quad JaxMaze & 40 & 40 & 40 & 40 & 40 \\
\quad Craftax & 80 & 80 & 80 & 40 & 40 \\
\multicolumn{6}{l}{\textit{Learning rate}} \\
\quad JaxMaze & $10^{-3}$ & $10^{-3}$ & $10^{-4}$ & $3 \times 10^{-4}$ & $3 \times 10^{-4}$ \\
\quad Craftax & $10^{-3}$ & $10^{-3}$ & $3 \times 10^{-4}$ & $3 \times 10^{-4}$ & $3 \times 10^{-4}$ \\
\multicolumn{6}{l}{\textit{MLP hidden layers}} \\
\quad JaxMaze & 2 & 3 & 2 & 2 & 2 \\
\quad Craftax & 3 & 3 & 3 & 2 & 2 \\
\multicolumn{6}{l}{\textit{Number of simulations}} \\
\quad JaxMaze & - & - & - & 2 & 2 \\
\quad Craftax (Human) & - & - & - & 4 & 4 \\
\quad Craftax (AI) & - & - & - & 10 & 10 \\
\multicolumn{6}{l}{\textit{Auxiliary task coefficients}} \\
\quad 1-step prediction & - & $10^{-5}$ & - & - & - \\
\quad SF auxiliary & - & $10^{-6}$ & - & - & - \\
\multicolumn{6}{l}{\textit{Simulation length}} \\
\quad JaxMaze & - & - & - & 15 & 15 \\
\quad Craftax & - & - & - & 20 & 20 \\
\multicolumn{6}{l}{\textit{Conservative Q-learning coefficient}} \\
\quad JaxMaze & - & - & $10^{-3}$ & - & $10^{-3}$ \\
\quad Craftax (Human) & - & - & $10^{-2}$ & - & $10^{-2}$ \\
\quad Craftax (AI) & - & - & $10^{-3}$ & - & $10^{-2}$ \\
\multicolumn{6}{l}{\textit{Conservative all-goals coefficient}} \\
\quad All settings & - & - & $1$ & - & $1$ \\
\multicolumn{6}{l}{\textit{Goal weight ($\alpha_\goal$)}} \\
\quad JaxMaze & - & - & - & - & $0$ \\
\quad Craftax (Human) & - & - & - & - & $0$ \\
\quad Craftax (AI) & - & - & - & - & $1$ \\
\multicolumn{6}{l}{\textit{Off-task goal weight ($\alpha_\offtask$)}} \\
\quad JaxMaze & - & - & - & - & $1$ \\
\quad Craftax (Human) & - & - & - & - & $1$ \\
\quad Craftax (AI) & - & - & - & - & $2$ \\
\hline
\end{tabular}
\end{table}
\begin{extfigure}[htbp]
    \centering
    \includegraphics[width=0.32\textwidth]{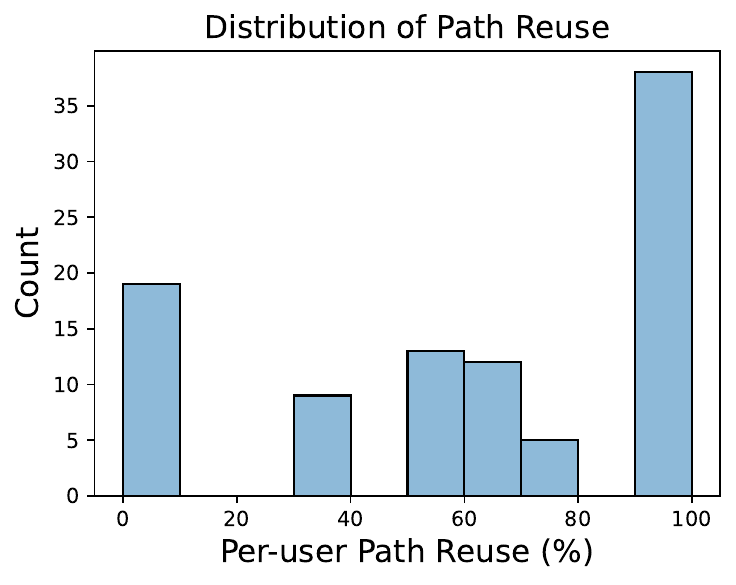}%
    \hfill
    \includegraphics[width=0.32\textwidth]{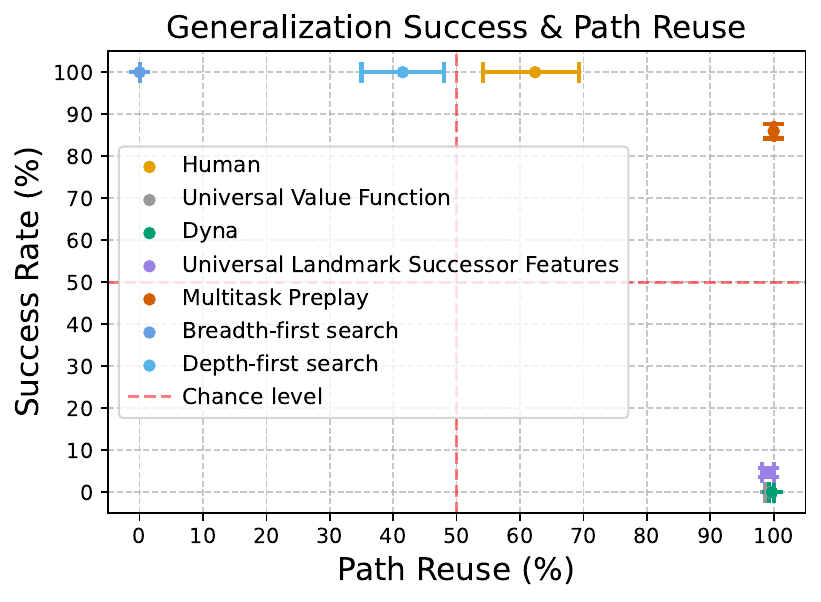}%
    \hfill
    \includegraphics[width=0.32\textwidth]{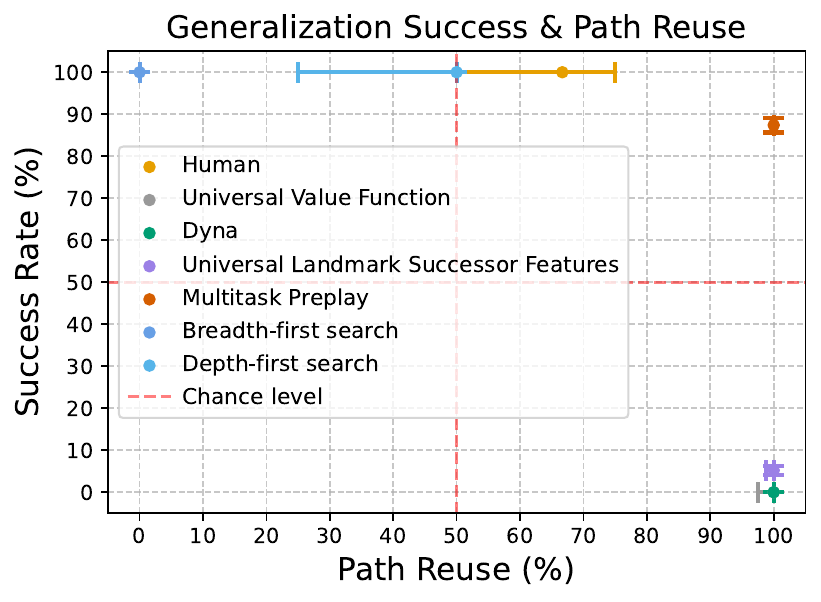}
    \caption{\textbf{JaxMaze Path Reuse supplementary results.} (Left) Human rate distributions. (Center) Mean success rates. (Right) Median success rates.}
    \label{ext:path-reuse-rates}
\end{extfigure}

\begin{extfigure}[htbp]
    \centering
    \includegraphics[width=\textwidth]{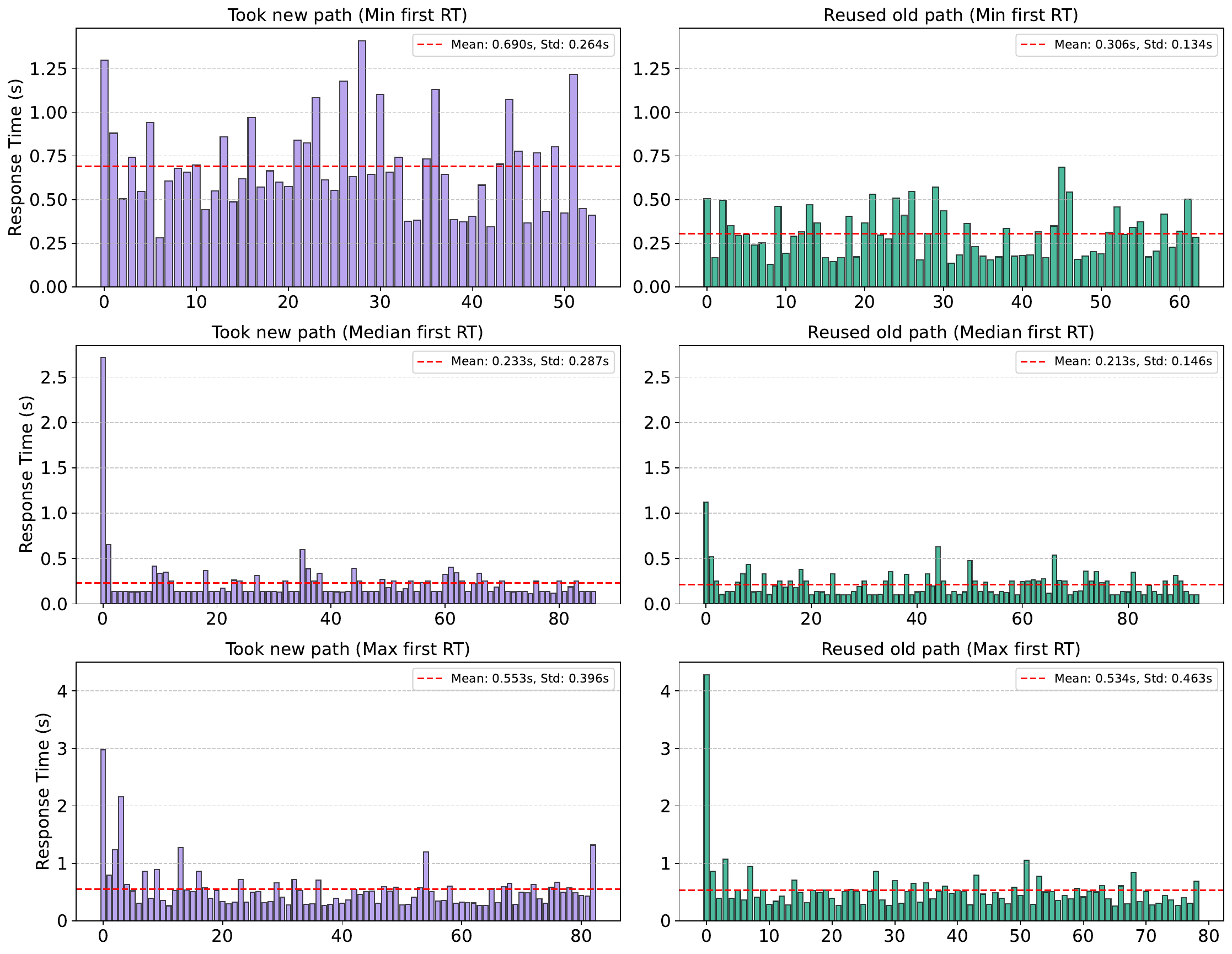}
    \caption{
        \textbf{Example differences in RTs when people take a new path (left) or reuse an old path (right)}.
    }
    \label{ext:two-paths}
\end{extfigure}

\begin{extfigure}[htbp]
    \centering
    \begin{subfigure}[c]{0.48\textwidth}
        \centering
        \includegraphics[height=0.3\textheight]{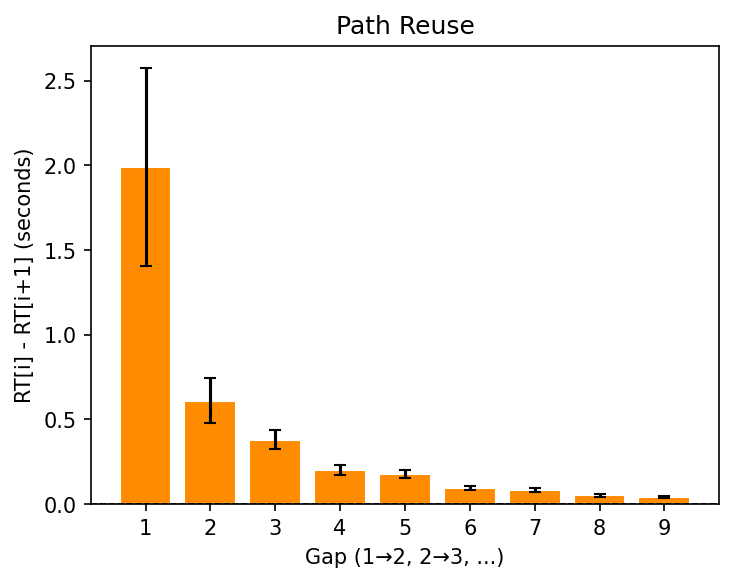}
        \caption{{The relative difference between RT[i] - RT[i+1] for the top-10 RTs}, averaged over both participants and trials. For every individual trial (i.e. episode) a participant completes, we compute this metrics. We then average over both all trials and all participants. For each index, we present mean and standard error.}
    \end{subfigure}%
    \hfill
    \begin{subfigure}[c]{0.48\textwidth}
        \centering
        \includegraphics[height=0.3\textheight]{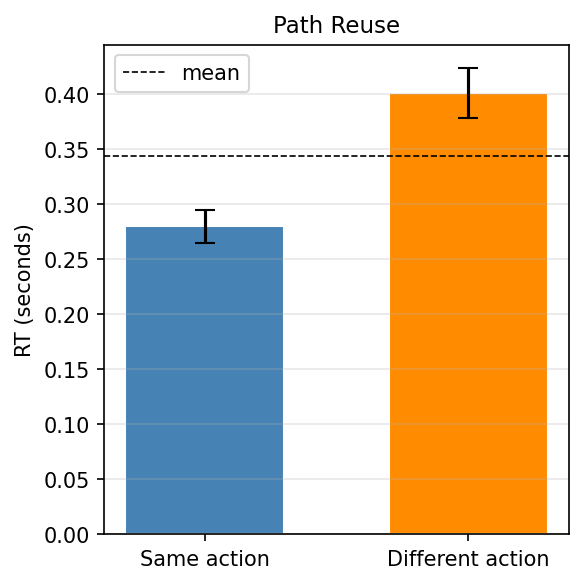}
        \caption{The average response time when people take the same action as in the previous timestep or a different action. We sort RTs into whether the action take is the same as the previous or different and present the mean and standard error over these sets.}
    \end{subfigure}
    \caption{
        \textbf{Characterizing the response time distributions}. In (a) we show that, on average for every participant, the highest response times is \textbf{greater} than 2 seconds longer than the next highest response times. In (b) we show that participants take longer to select action when the action is different from the previous timestep. Together, these results that suggest that the bulk of ``computation'' (i.e. cognitive work) is concentrated in one or a few timesteps, but that there are other timesteps where participants consistently think longer (e.g. when they need to select a new action).
    }
    \label{ext:top10}
\end{extfigure}

\begin{extfigure}[htbp]
    \centering
    \includegraphics[width=\textwidth]{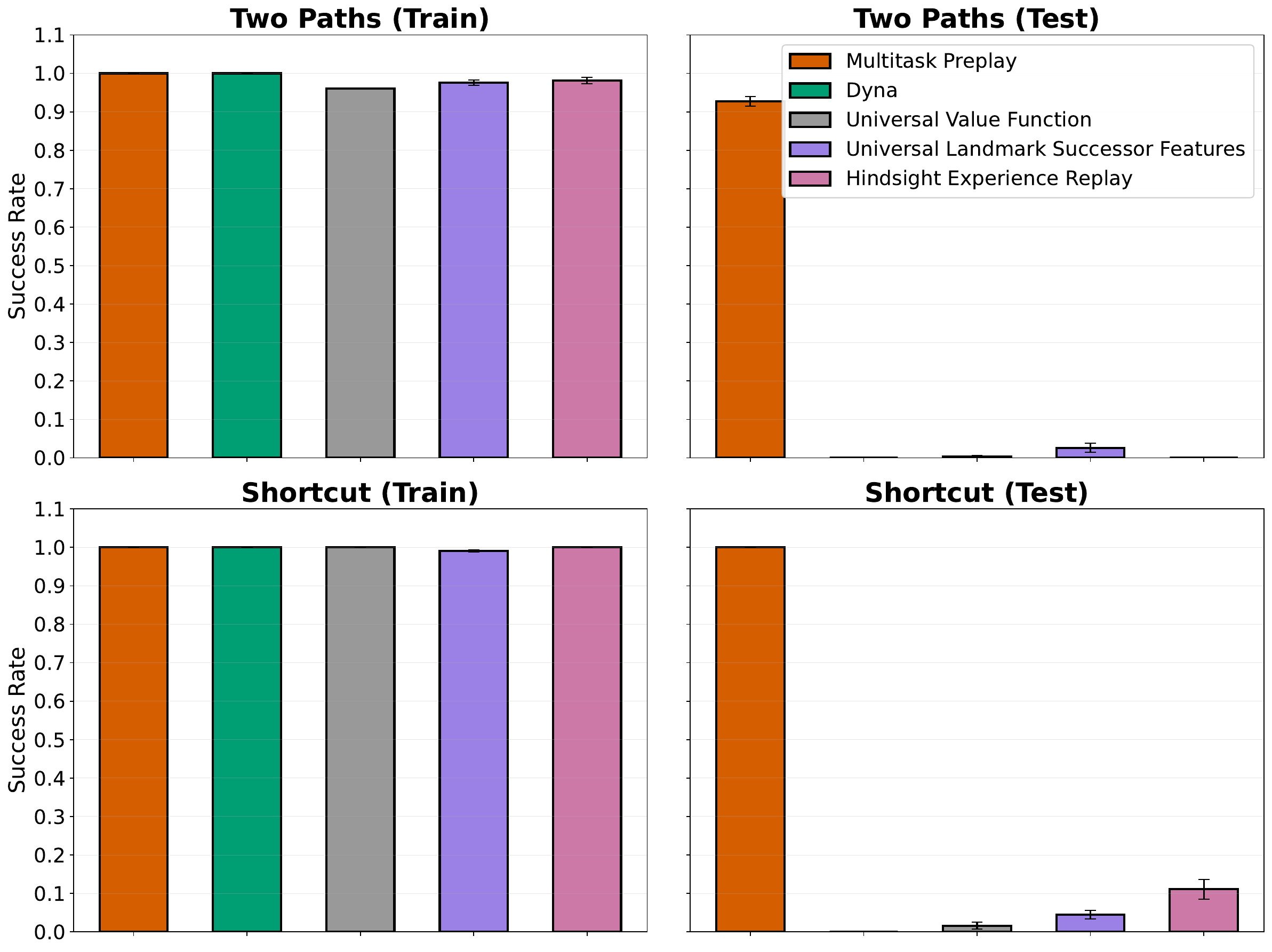}
    \caption{
        \textbf{JaxMaze training performance.} Train and test success rates for all models on the JaxMaze environment.
    }
    \label{ext:jaxmaze_train}
\end{extfigure}

\begin{extfigure}[htbp]
    \centering
    \includegraphics[width=\textwidth]{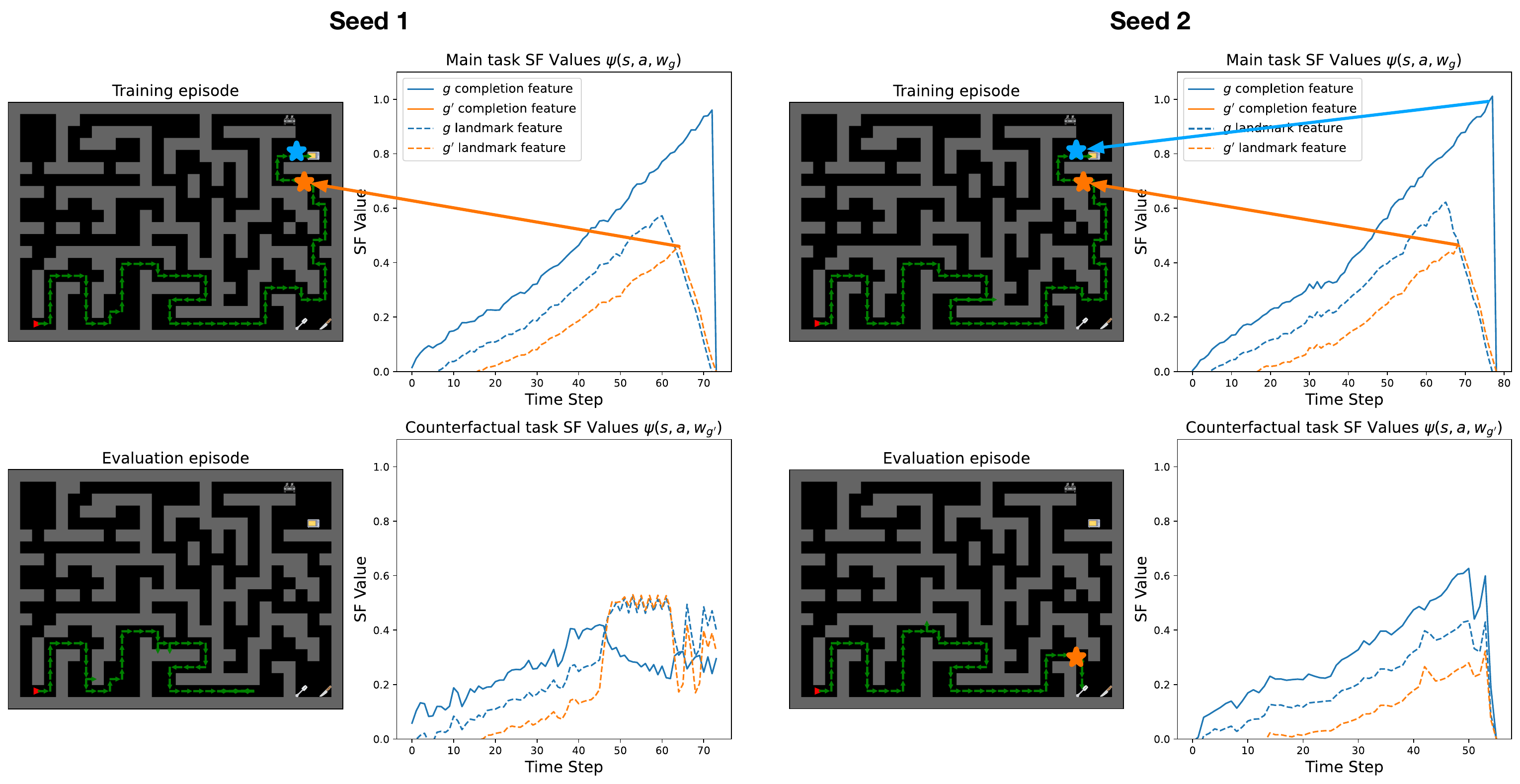}
    \caption{
        \textbf{Landmark successor features fail to generalize to ``off-task'' goals $\offtask$ nearby main task goal $\goal$ (Figure 3 A\&B)}.
        Shown are two typical failure modes when $\psi(s,a, \cdot)$ predicts goal completion features and landmark features.
        \textbf{Top row}. The first mechanism for generalization is to select actions with high $\hat{\psi}^\pi_\theta(s,a,\goal)^{\top}w_{\offtask}$. We can see that, counterintuitively, landmark features lead a corresponding ``successor feature'' (orange) with a maximum value at a position along a previous path that is \textit{not closest} to $\offtask$ but instead a position from which you will most often be ``nearby'' $\offtask$.
        To complete $\offtask$, some form of planning is required.
        \textbf{Bottom row}. The other mechanism for generalization is to directly learn $\psi(s,a,\offtask)$ with trajectories collected while pursuing $\goal$ via off-task learning~\cite{borsa2018universal}.
        While this is possible in principle, we identify two common failure modes: either the resulting estimate is noisy and uninformative (left) or $\hat{\psi}^\pi_\theta(s,a,\offtask)$ ends up converging to the $\psi(s,a,\goal)$ for a high signal training goal $\goal$ (right).
    }
    \label{ext:sf-analysis}
\end{extfigure}

\begin{extfigure}[htbp]
    \centering
    \includegraphics[width=\textwidth]{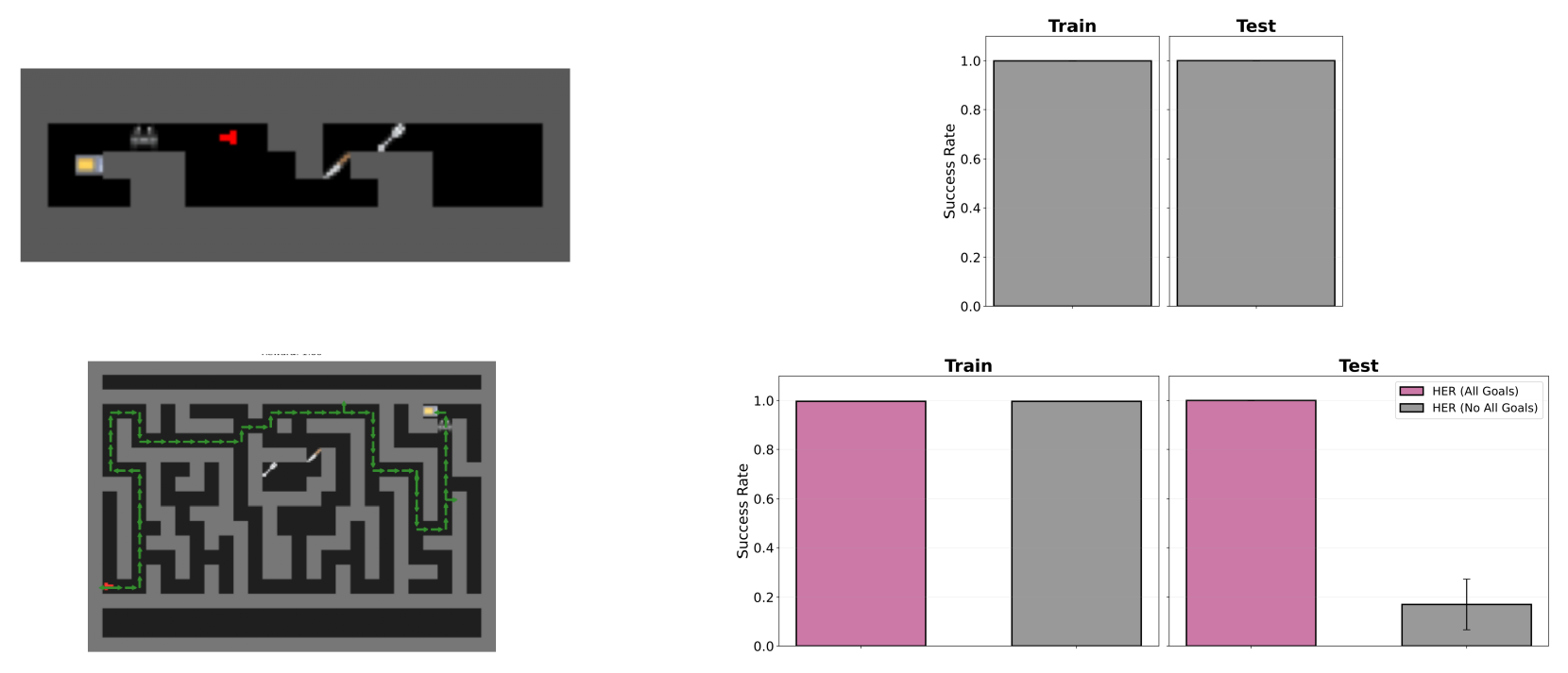}
    \caption{
        \textbf{HER with and without all-goals learning.} Train and test success rates for Hindsight Experience Replay (HER) on the small (top) and large (bottom) JaxMaze environments. In the large environment, HER with all-goals learning substantially outperforms HER without all-goals learning on held-out test mazes, while both achieve near-perfect training performance.
    }
    \label{ext:her-comparison}
\end{extfigure}

\begin{extfigure}[htbp]
    \centering
    \includegraphics[width=0.32\textwidth]{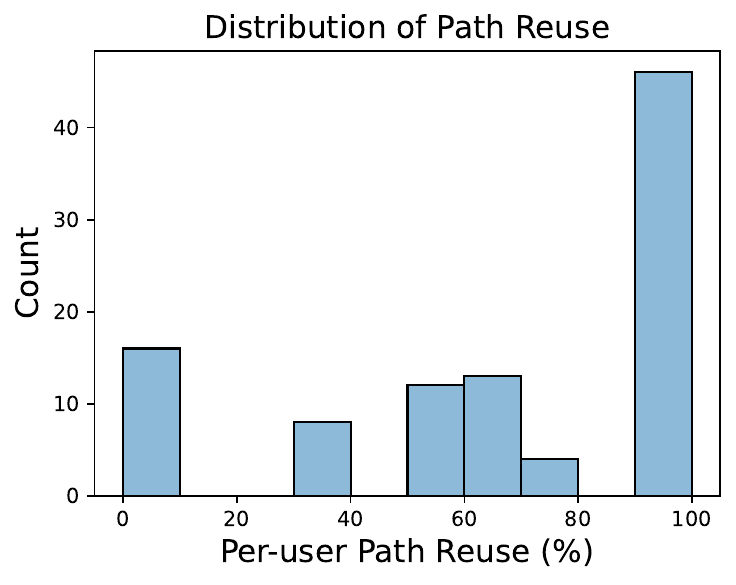}%
    \hfill
    \includegraphics[width=0.32\textwidth]{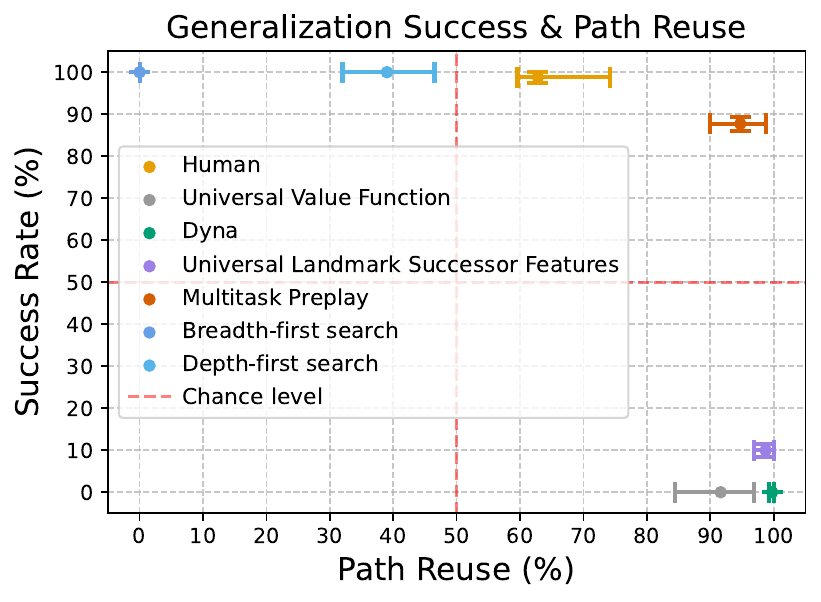}%
    \hfill
    \includegraphics[width=0.32\textwidth]{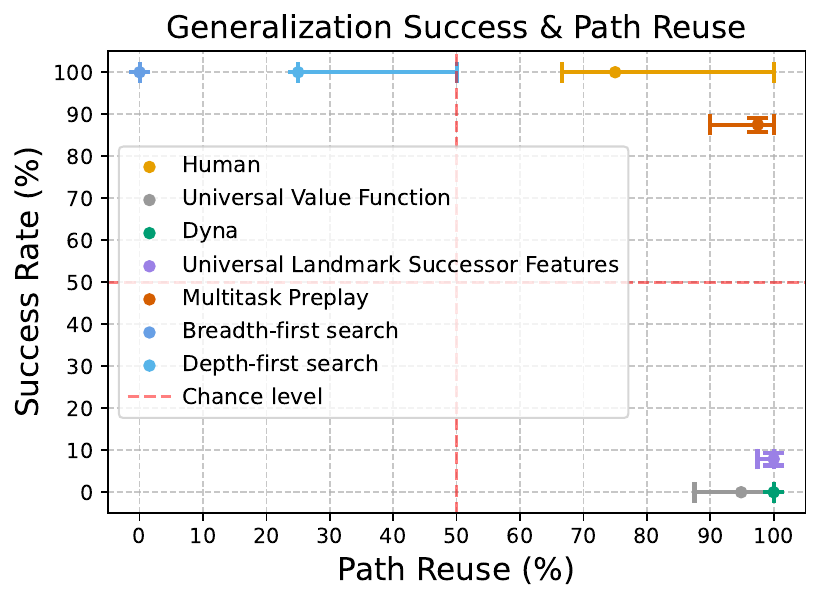}
    \caption{\textbf{JaxMaze Shortcut supplementary results.} (Left) Human rate distributions. (Center) Mean success rates. (Right) Median success rates.}
    \label{ext:shortcut-rates}
\end{extfigure}

\begin{extfigure}[htbp]
    \centering
    \includegraphics[width=\textwidth]{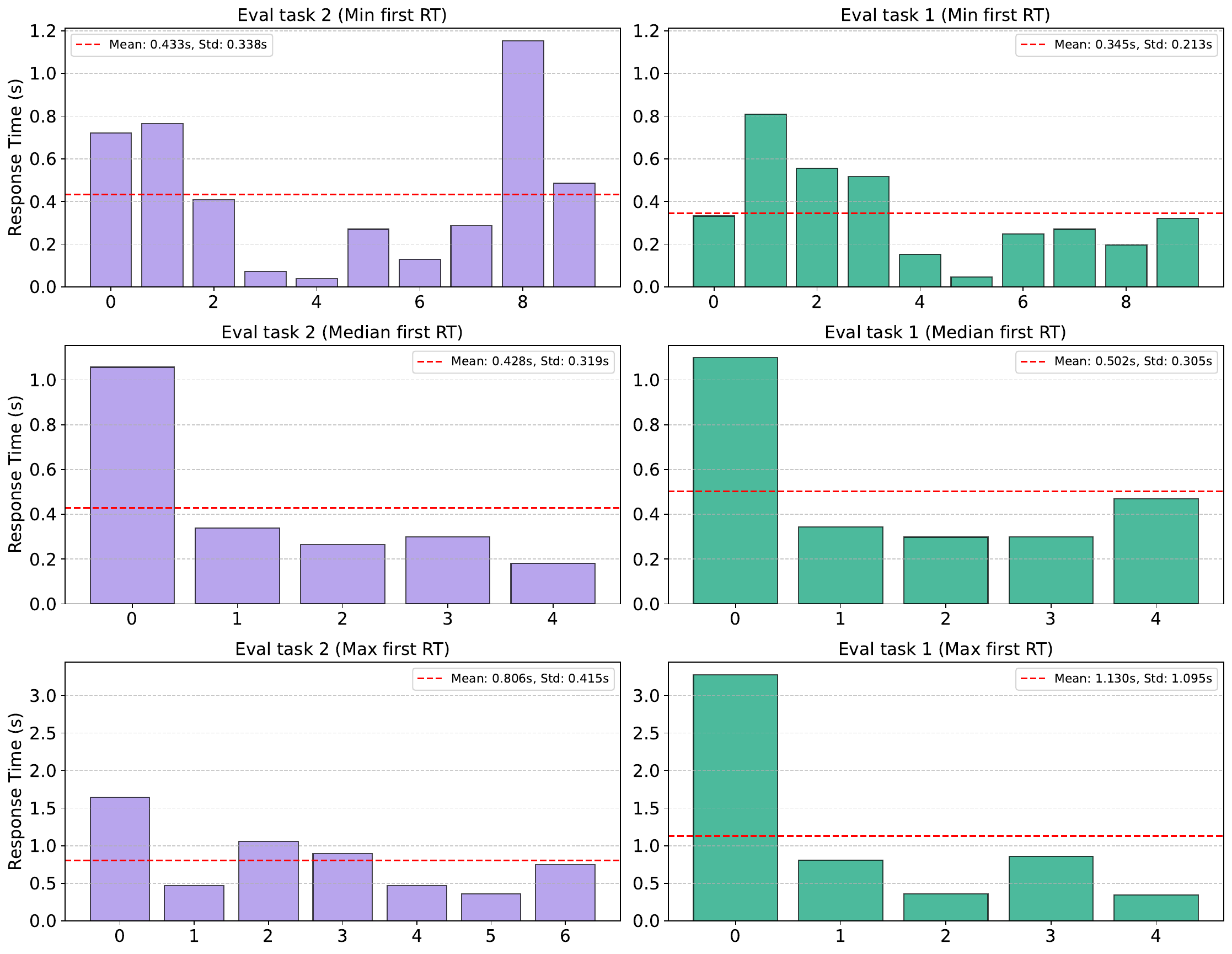}
    \caption{
        \textbf{Differences in RTs when people take a new path in new part of maze (left) or took a new path in a part of the maze where they could have used preplay (right)}.
        Example RTs over an episode during eval task 2 (left) or eval task 1 (right) in a setting where the goal object is unknown (Figure 3 E\&H).
    }    \label{ext:juncture}
\end{extfigure}

\begin{extfigure}[htbp]
    \centering
    \includegraphics[width=\textwidth]{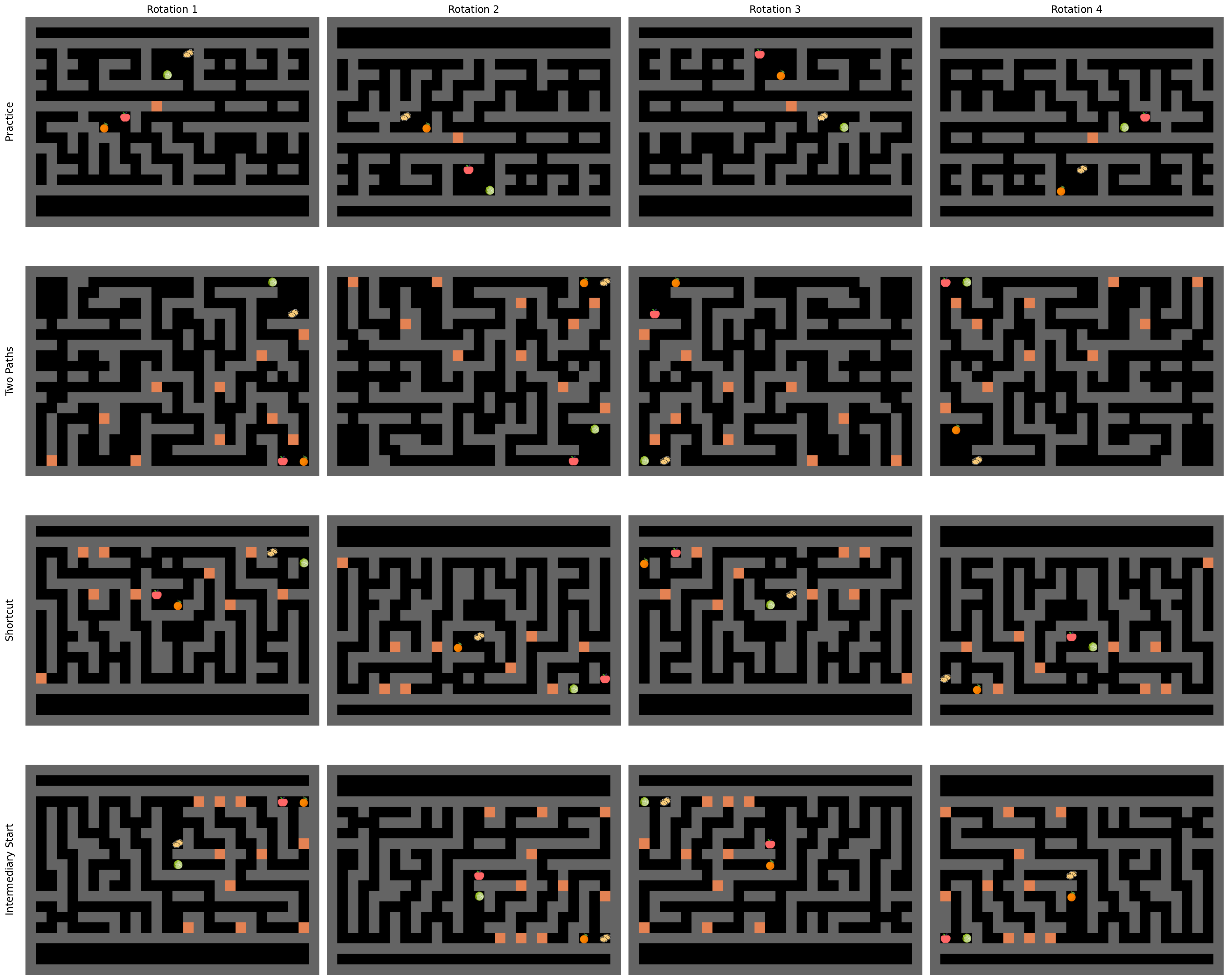}
    \caption{
        \textbf{JaxMaze environments used in experiments 1--3}.
        Each column shows a different rotation of the same maze. Rows correspond to (top to bottom): the practice maze, the Two Paths experiment, the Shortcut experiment, and the Intermediary Start experiment. Object icons indicate goal locations and red squares indicate possible starting locations.
    }
    \label{ext:jaxmaze-envs}
\end{extfigure}

\begin{extfigure}[htbp]
    \centering
    \includegraphics[width=\textwidth]{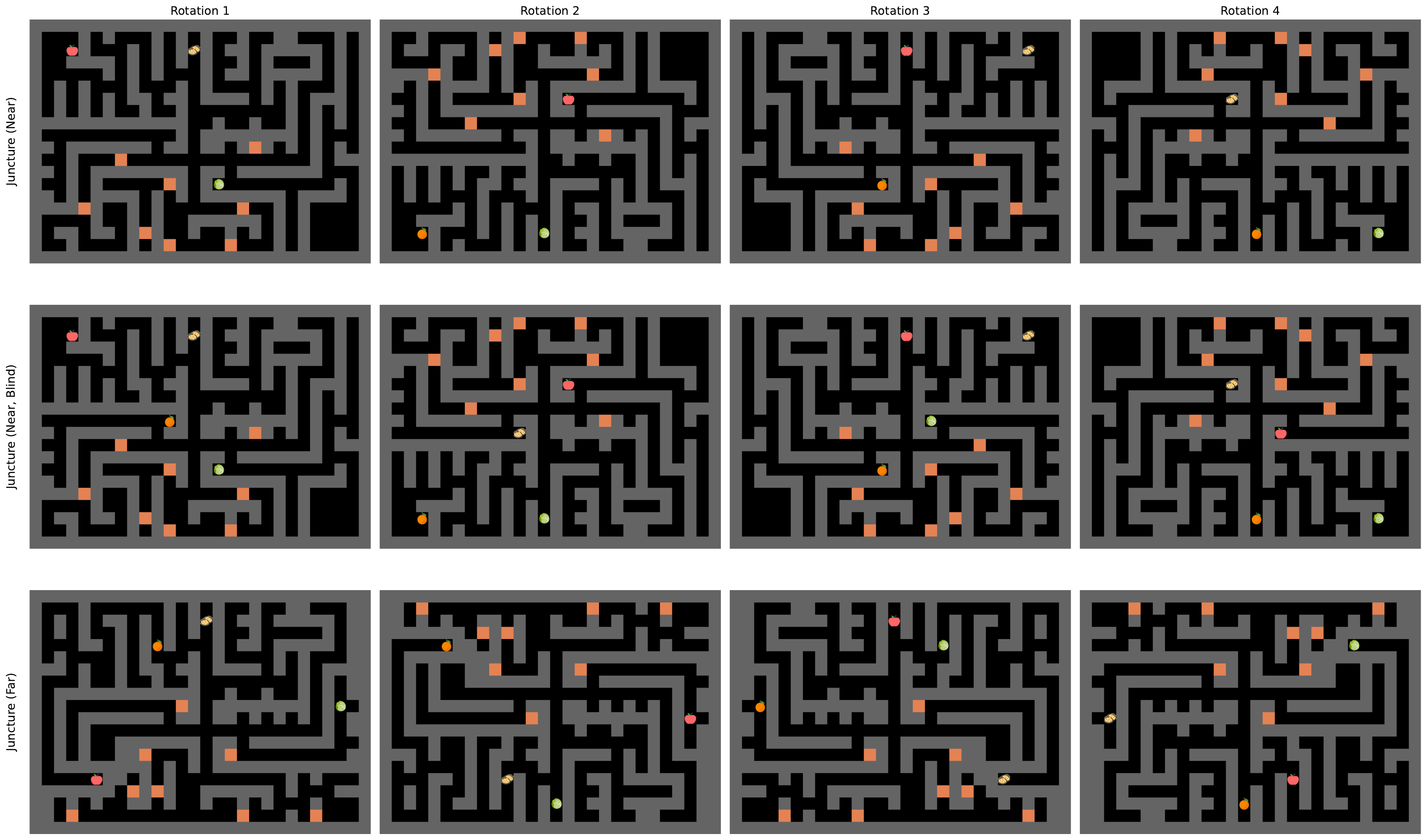}
    \caption{
        \textbf{JaxMaze environments used in the Juncture experiment (experiment 4)}.
        Each column shows a different rotation. Rows correspond to the three conditions: Near with known test goal, Near with unknown test goal, and Far with known test goal. Object icons indicate goal locations and red squares indicate possible starting locations.
    }
    \label{ext:jaxmaze-juncture-envs}
\end{extfigure}

\begin{extfigure}[htbp]
    \centering
    \includegraphics[width=\textwidth]{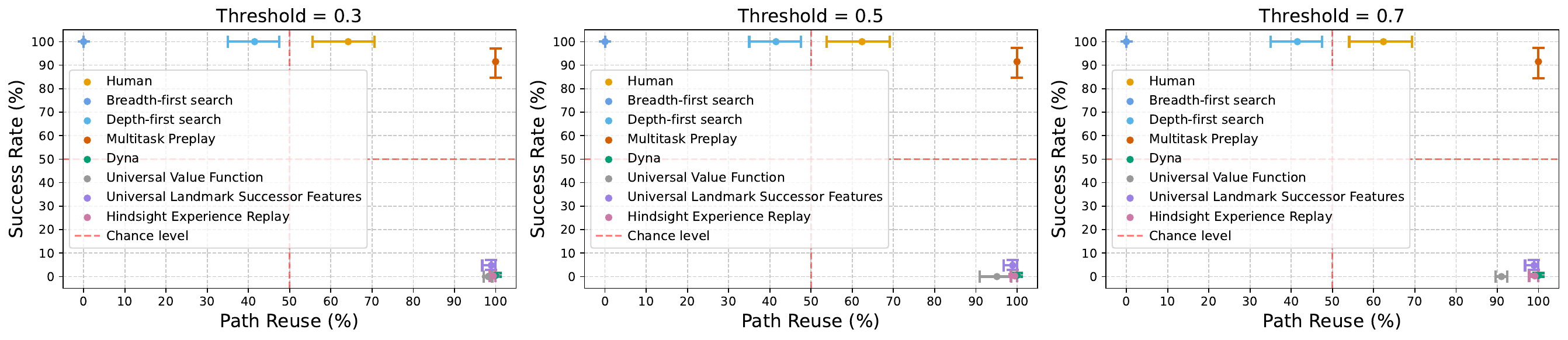}
    \caption{
        \textbf{JaxMaze prediction-1 path-reuse proportions are robust to the threshold $\alpha_{\tt map}$.}
        Proportion of evaluation trials classified as path reuse across different $\alpha_{\tt map}$ values. Because the JaxMaze environments used for prediction 1 admit essentially only two viable paths to each test object, the qualitative conclusions do not depend on the exact threshold choice.
    }
    \label{ext:jaxmaze-threshold-robustness}
\end{extfigure}

\begin{extfigure}[htbp]
    \centering
    \includegraphics[height=.9\textheight,keepaspectratio]{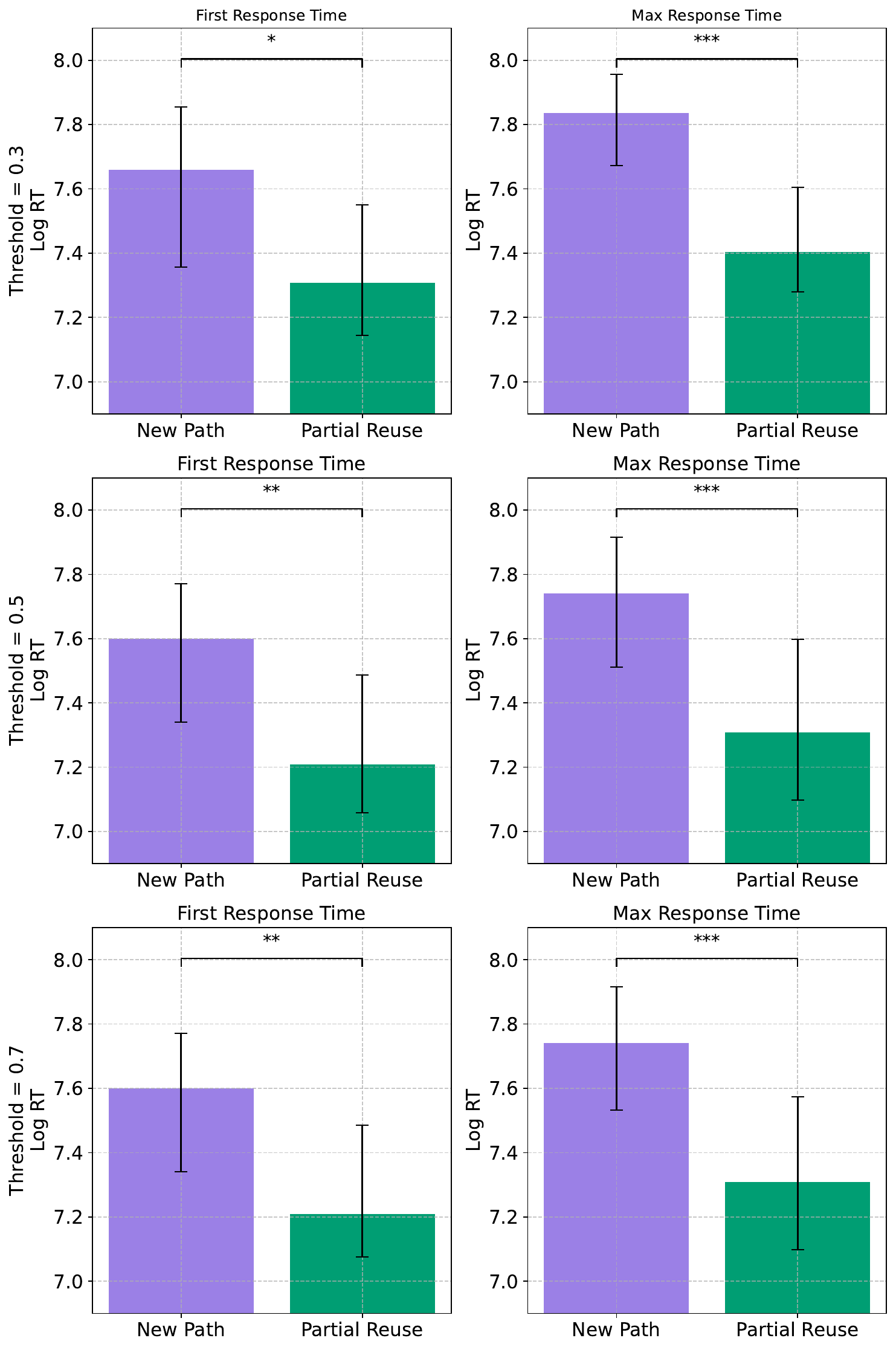}
    \caption{
        \textbf{JaxMaze prediction-1 RT effects are robust to the threshold $\alpha_{\tt map}$.}
        Log-RT comparisons between path-reuse and new-path trials across different $\alpha_{\tt map}$ values. The RT advantage for reused paths is preserved across thresholds.
    }
    \label{ext:jaxmaze-threshold-robustness-rt}
\end{extfigure}

\begin{extfigure}[htbp]
    \centering
    \includegraphics[width=\textwidth]{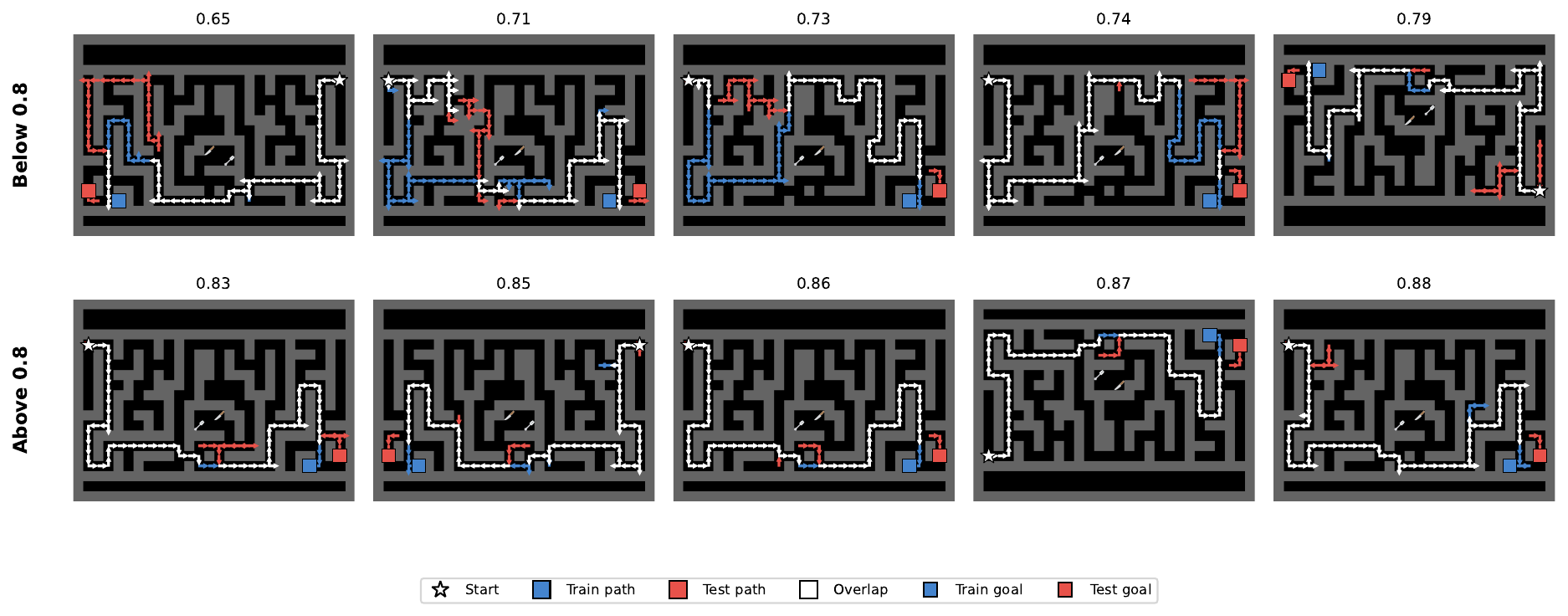}
    \caption{
        \textbf{The $\alpha_{\tt map}=.8$ threshold separates shortcut-taking from non-shortcut behavior in JaxMaze prediction 2.}
        Top row: evaluation paths that partially take a newly-opened shortcut produce overlap values that fall below $.8$. Bottom row: overlap values above $.8$ correspond to trials where the subject took no shortcut and reused the training path.
    }
    \label{ext:jaxmaze-shortcut-summary}
\end{extfigure}

\begin{extfigure}[htbp]
    \centering
    \begin{subfigure}[t]{0.7\textwidth}
        \centering
        \includegraphics[width=\textwidth]{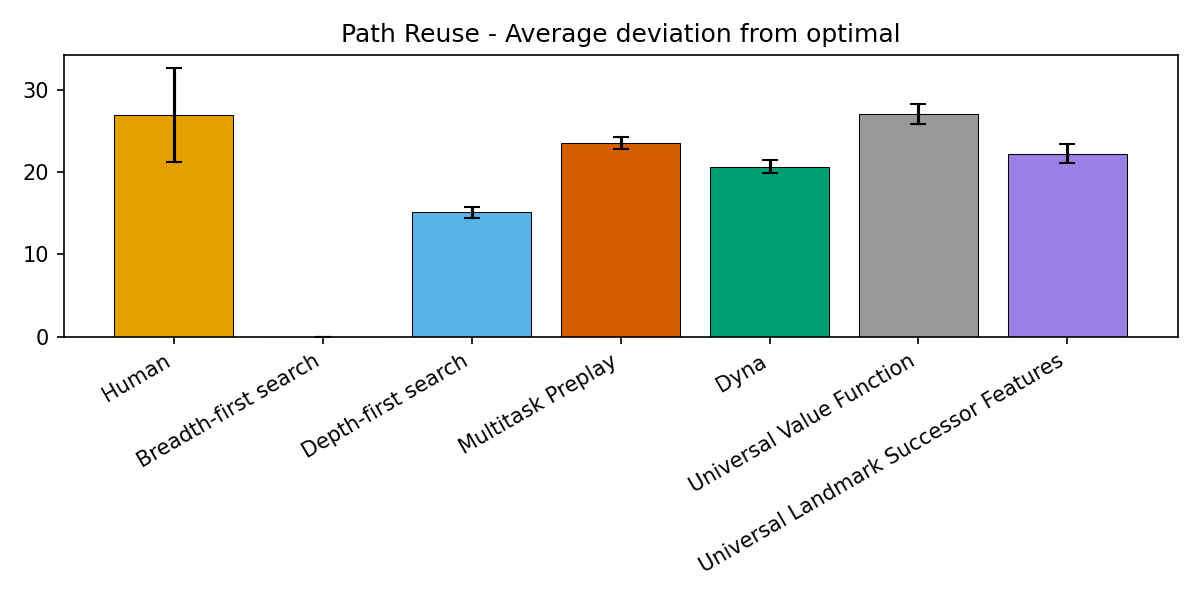}
        \caption{Experiment 1 (Path Reuse)}
    \end{subfigure}
    \vspace{1em}
    \begin{subfigure}[t]{0.7\textwidth}
        \centering
        \includegraphics[width=\textwidth]{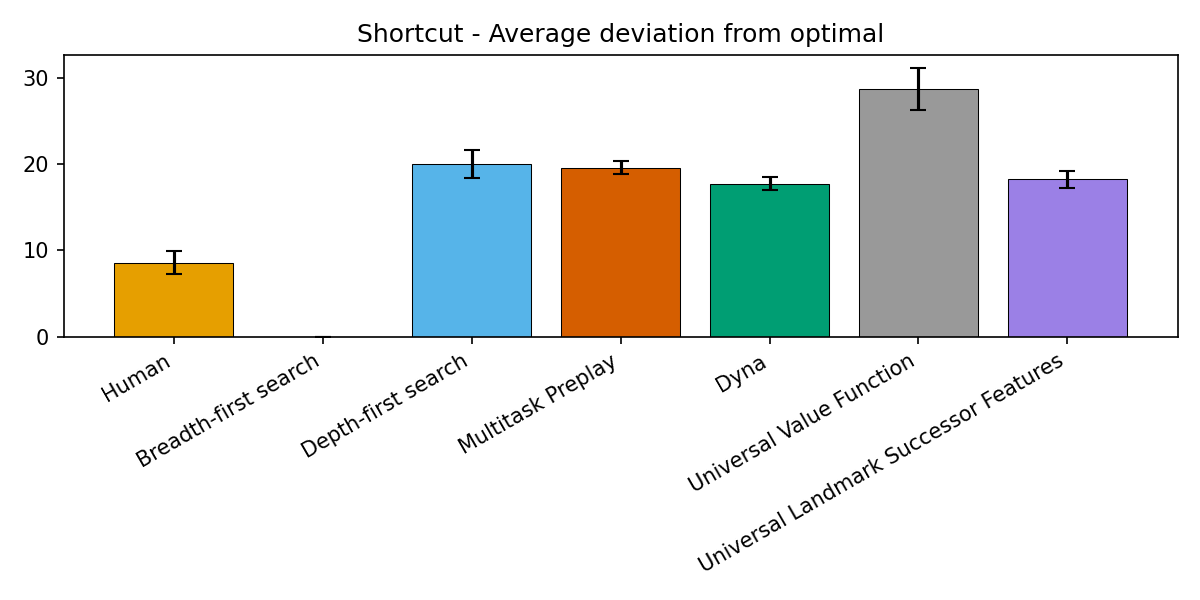}
        \caption{Experiment 2 (Shortcut)}
    \end{subfigure}
    \caption{\textbf{Optimal length deviations across experiments 1 and 2.} Distribution of deviations from optimal path length for (A) the path reuse experiment and (B) the shortcut experiment.}
    \label{ext:optimal-length-deviations}
\end{extfigure}

\begin{extfigure}[htbp]
    \centering
    \includegraphics[width=\textwidth]{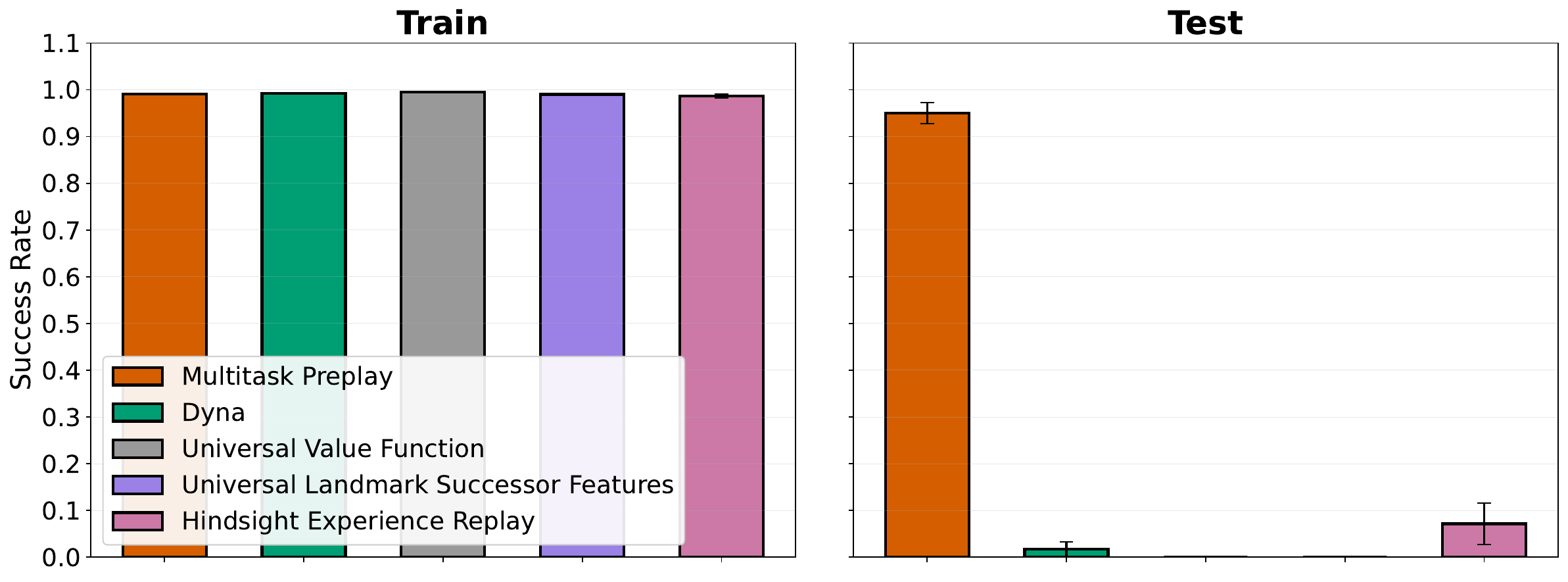}
    \caption{
        \textbf{Craftax training performance.} Train and test success rates for all models on the Craftax environment.
    }
    \label{ext:craftax_train_test}
\end{extfigure}

\begin{extfigure}[htbp]
    \centering
    \includegraphics[width=0.75\textwidth]{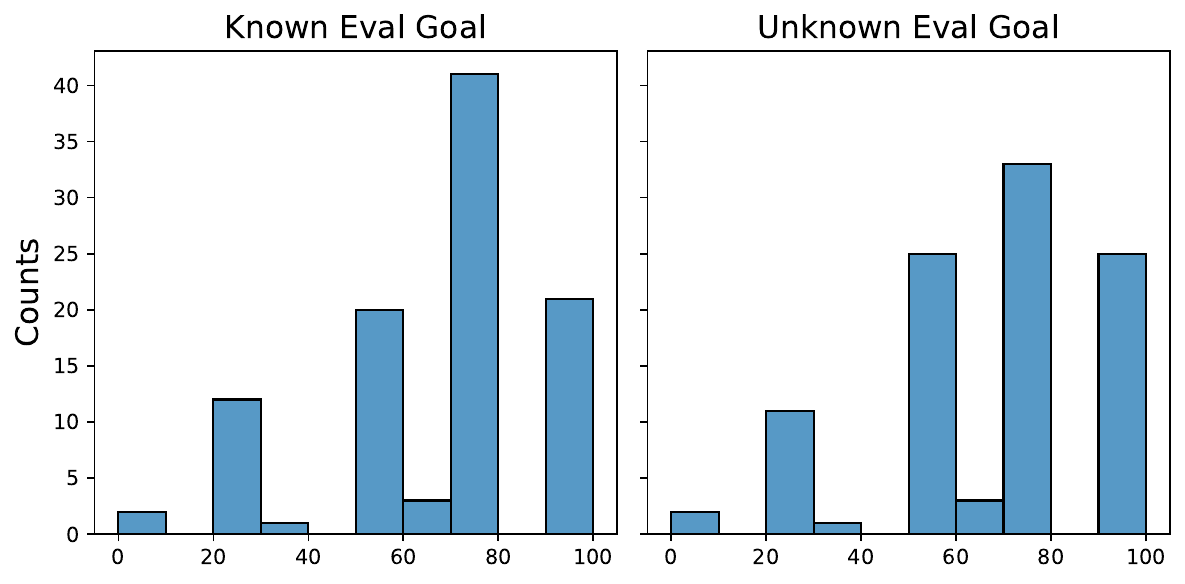}\\[1ex]
    \includegraphics[width=0.48\textwidth]{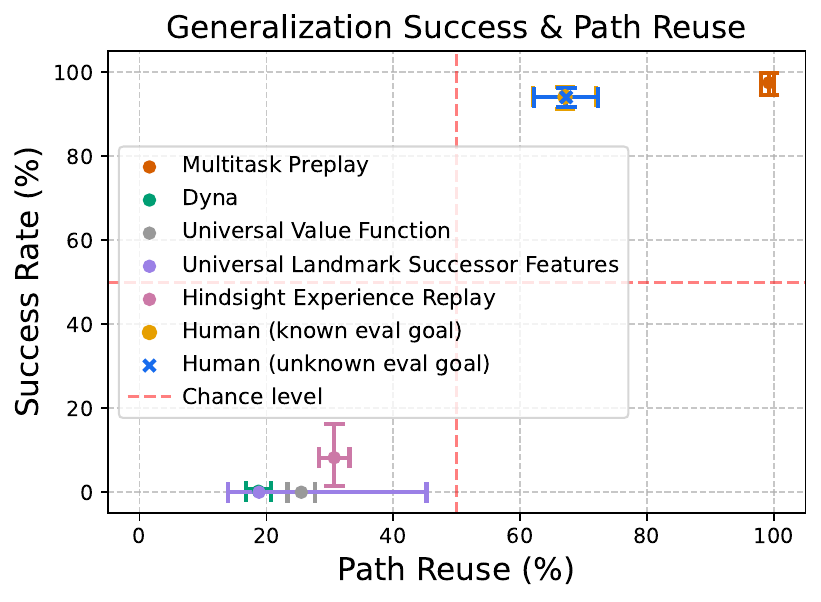}%
    \hfill
    \includegraphics[width=0.48\textwidth]{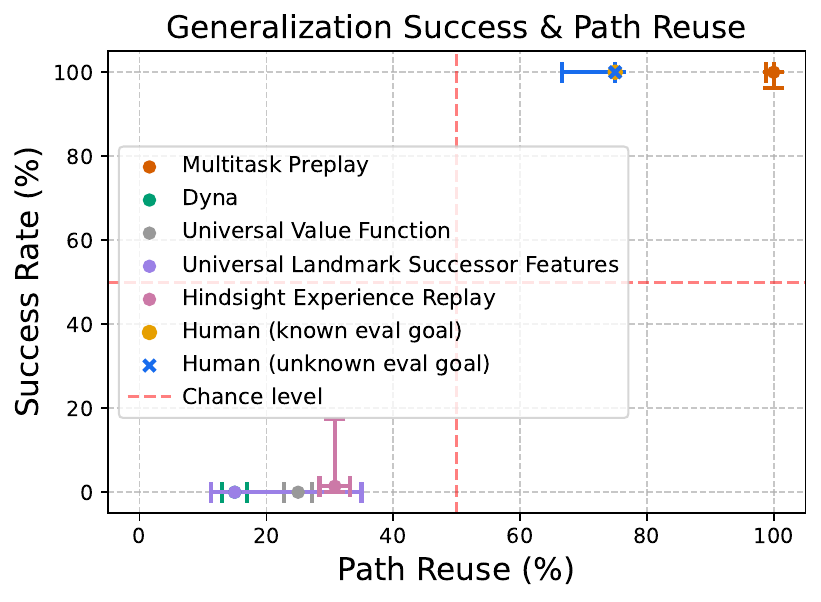}
    \caption{\textbf{Craftax Path Reuse supplementary results.} (Top) Human path reuse rate distributions. (Bottom left) Mean success rates. (Bottom right) Median success rates.}
    \label{ext:craftax-path-reuse-rates}
\end{extfigure}

\begin{extfigure}[htbp]
    \centering
    \includegraphics[width=0.6\textwidth]{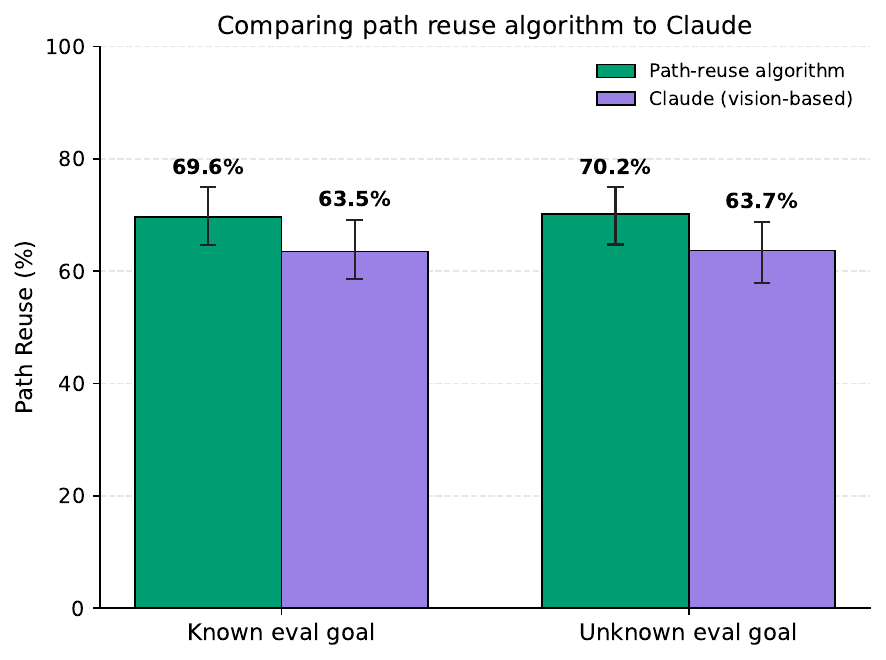}
    \caption{\textbf{Validating the Craftax path-reuse criterion with a vision-based language model.} Path-reuse rates from the geometric criterion (green) and from a vision-based LLM (Claude Sonnet 4.5; purple) agree within error: the $95\%$ confidence intervals overlap in both conditions, placing the two estimates within sampling error of one another. The LLM's classifications are uniformly more conservative (about $6$--$7$ percentage points lower in point estimate) but remain well above the $50\%$ chance baseline. Bars reflect the canonical cohort (first $100$ subjects per condition; $n=359$ trials with known goal, $n=344$ with unknown goal).}
    \label{ext:craftax-path-reuse-method-comparison}
\end{extfigure}

\begin{extfigure}[htbp]
    \centering
    \begin{tcolorbox}[
        colback=gray!5,
        colframe=gray!50!black,
        title=\textbf{Path-reuse classification prompt (Claude Sonnet 4.5)},
        fonttitle=\bfseries,
        width=0.95\textwidth,
        boxrule=0.5pt,
        arc=2pt,
        left=6pt, right=6pt, top=6pt, bottom=6pt,
        fontupper=\small\ttfamily
    ]
\begin{verbatim}
You will see two images of the same Craftax world. In each, a red line shows
the path a participant took from the yellow star (start) to the yellow circle
(goal). Both paths are from the same person on different trials -- first
reaching a training goal, then reaching a test goal placed nearby. They
started in the same position both times.

Read both images first:
1. Read {train_path}
2. Read {test_path}

Did the participant reuse the training path on the test trial? Answer two
questions:

1. SAME SIDE OF WATER: For each meaningful body of water (blue), did both
   paths go on the same side of it?

2. SAME ROUTE: Do both paths follow roughly the same route to the goal and
   approach from roughly the same direction? (A brief detour where the test
   path wanders off and then rejoins the training route is fine.)

Output exactly ONE JSON object as the LAST line of your response, no code
fences. Label "path_reuse" only if both answers are yes; otherwise
"no_path_reuse". Include a brief reasoning (1-2 sentences) describing what
you saw on each side of each path:
{"same_side": "yes|no", "same_approach": "yes|no",
 "reasoning": "<1-2 sentences>",
 "label": "path_reuse|no_path_reuse"}
\end{verbatim}
    \end{tcolorbox}
    \caption{\textbf{Prompt used to classify Crafter path reuse with Claude Sonnet 4.5.} Each evaluation trial was classified by sending Claude two rendered images of the participant's training and test paths and the prompt above. The placeholders \texttt{\{train\_path\}} and \texttt{\{test\_path\}} are replaced at inference time with file paths to the rendered images, which Claude reads via its vision-enabled file-reading tool. A trial is labeled \texttt{path\_reuse} only if Claude answers ``yes'' to both the same-side-of-water and same-route questions.}
    \label{ext:claude-path-reuse-prompt}
\end{extfigure}

\begin{extfigure}[htbp]
    \centering
    \includegraphics[width=\textwidth]{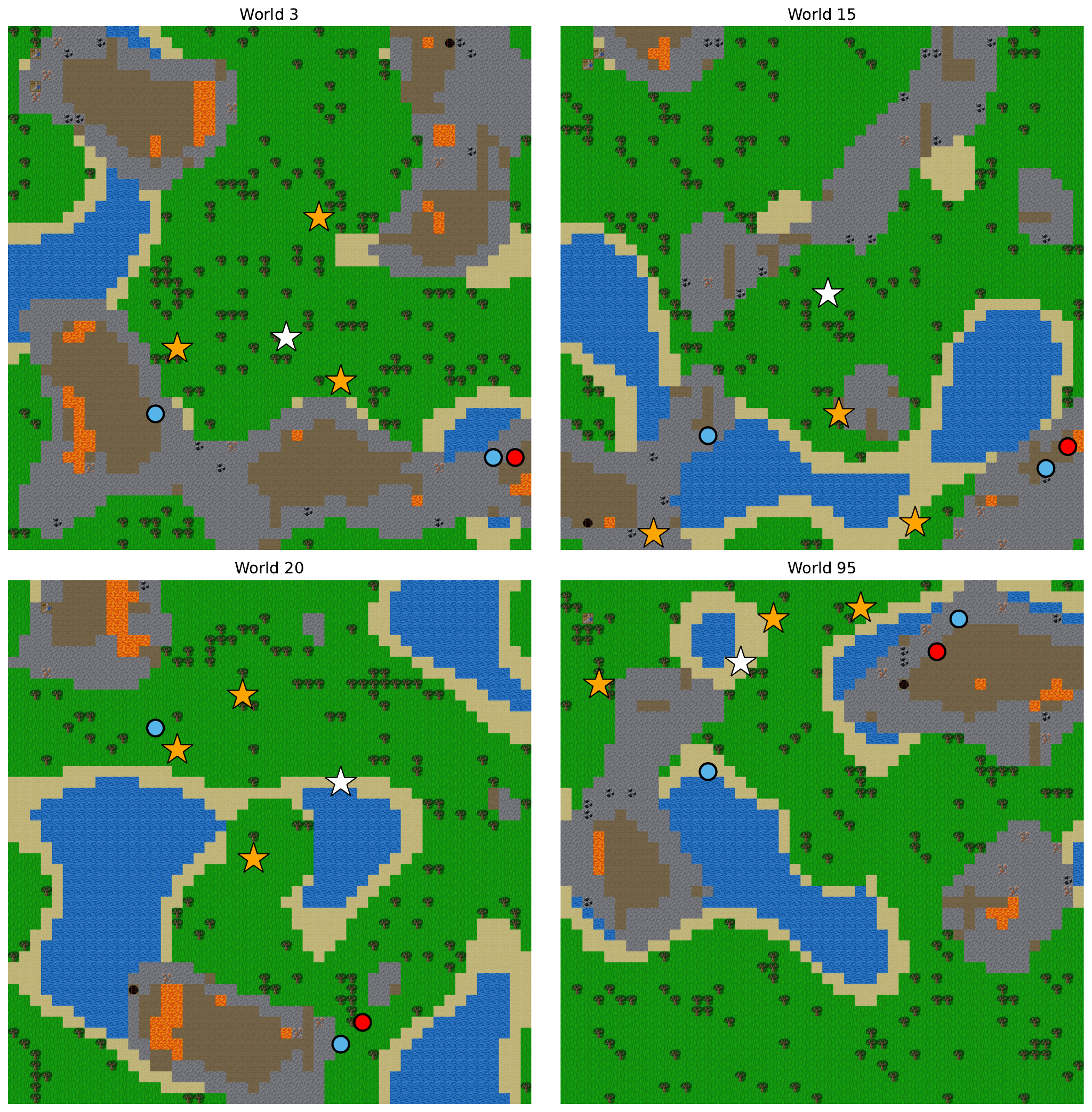}
    \caption{
        \textbf{Craftax environments used in the human experiment (experiment 5)}.
        Each panel shows one of the four procedurally generated worlds used across blocks. Stars indicate goal object locations (orange for training objects, white for the test object). Circles indicate possible starting positions (blue and red).
    }
    \label{ext:craftax-envs}
\end{extfigure}

\begin{extfigure}[htbp]
    \centering
    \begin{minipage}[t]{0.48\textwidth}
        \centering
        \includegraphics[width=\textwidth]{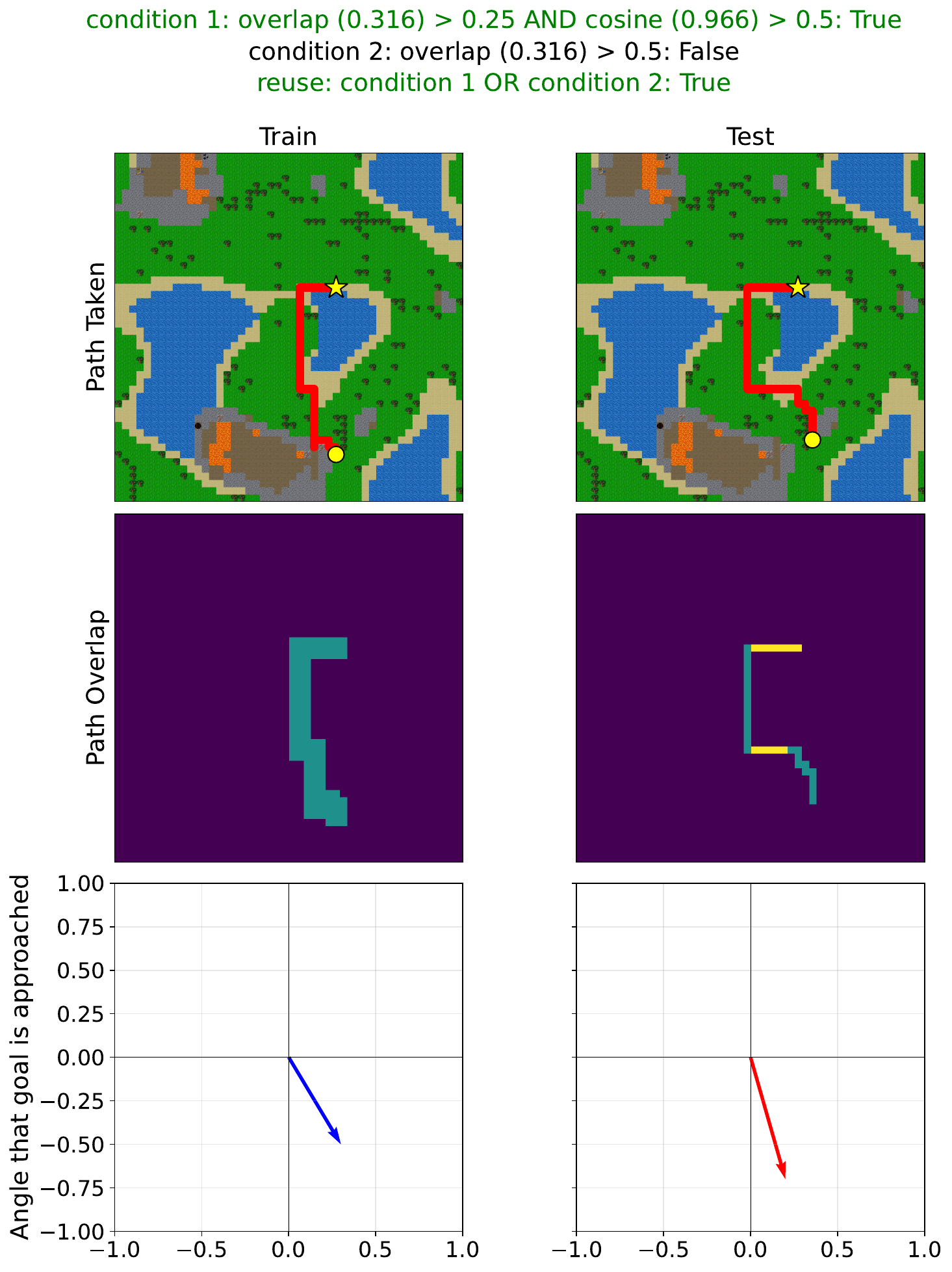}
        \subcaption{Direction-clause true positive.}
    \end{minipage}\hfill
    \begin{minipage}[t]{0.48\textwidth}
        \centering
        \includegraphics[width=\textwidth]{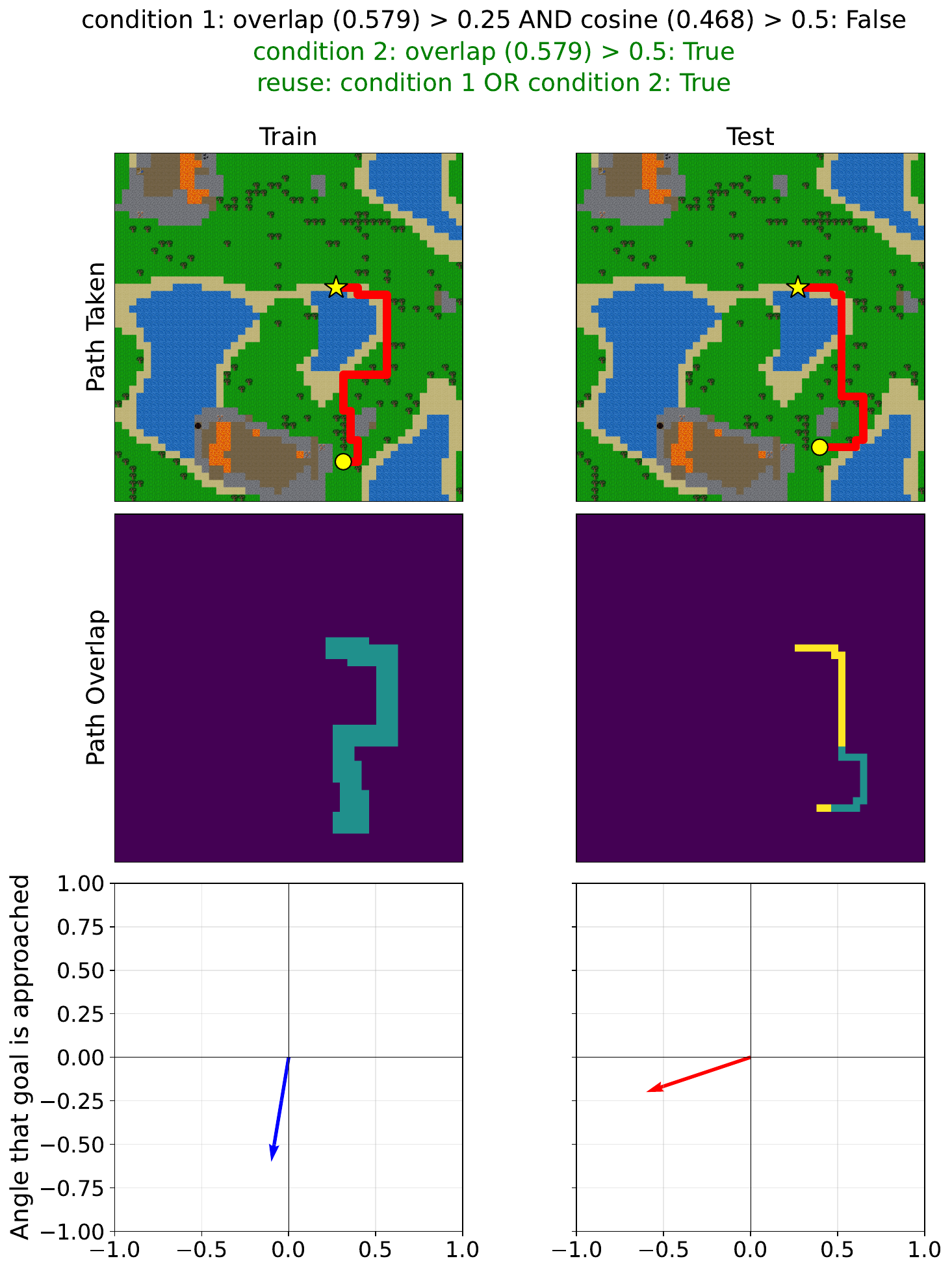}
        \subcaption{High-overlap-clause true positive.}
    \end{minipage}
    \caption{
        \textbf{Craftax path-reuse true positives under each clause.}
        (a) Path overlap is modest ($o_{\tt map}\approx.316$), but the angle of approach to the test object is consistent across train and test, so the direction clause fires. (b) Approach angles are near-perpendicular across train and test, yet path overlap is large enough that the high-overlap clause fires on its own.
    }
    \label{ext:craftax-overlap-tp-direction-vs-overlap}
\end{extfigure}

\begin{extfigure}[htbp]
    \centering
    \begin{minipage}[t]{0.48\textwidth}
        \centering
        \includegraphics[width=\textwidth]{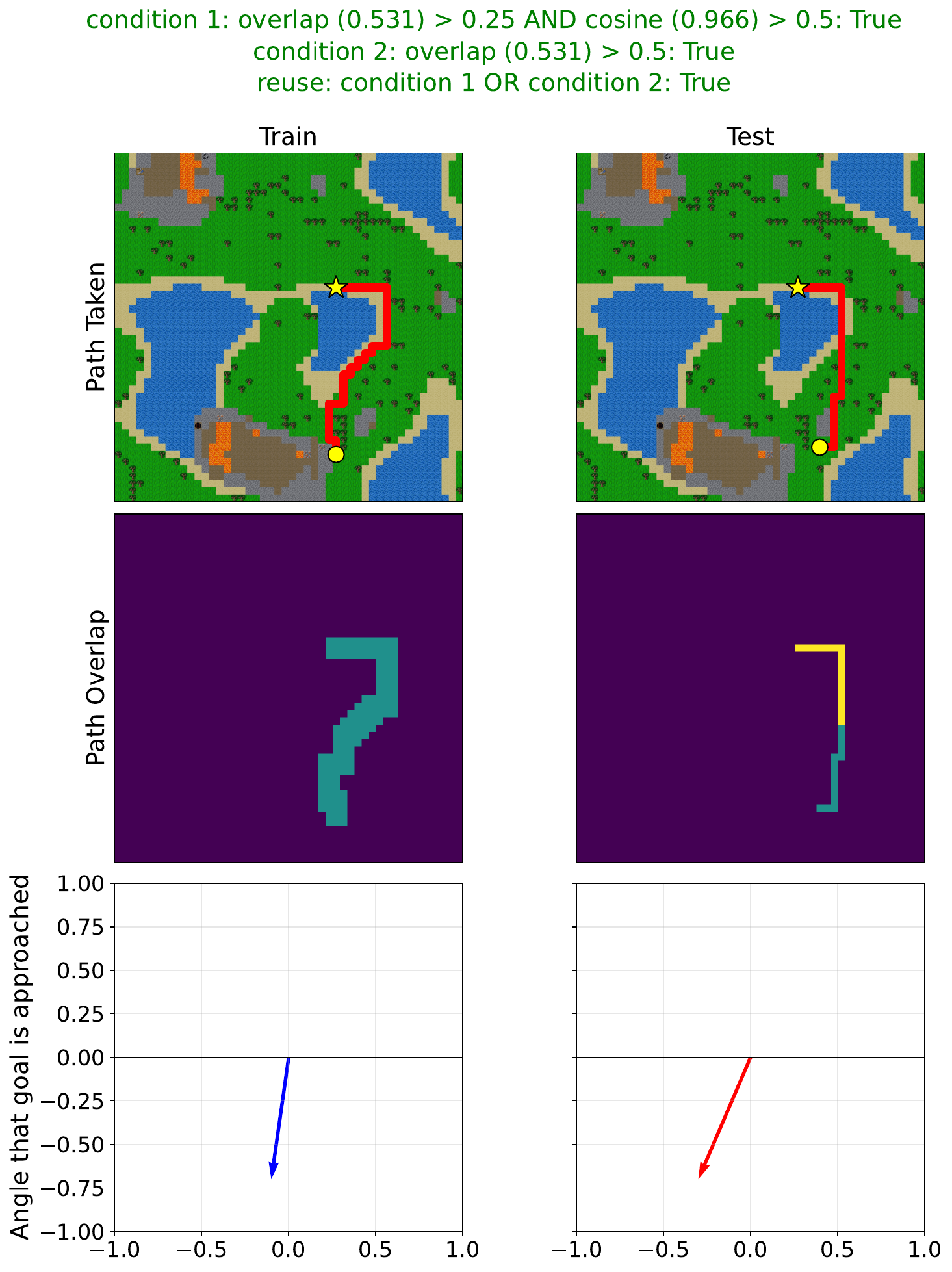}
        \subcaption{Both-clause true positive.}
    \end{minipage}\hfill
    \begin{minipage}[t]{0.48\textwidth}
        \centering
        \includegraphics[width=\textwidth]{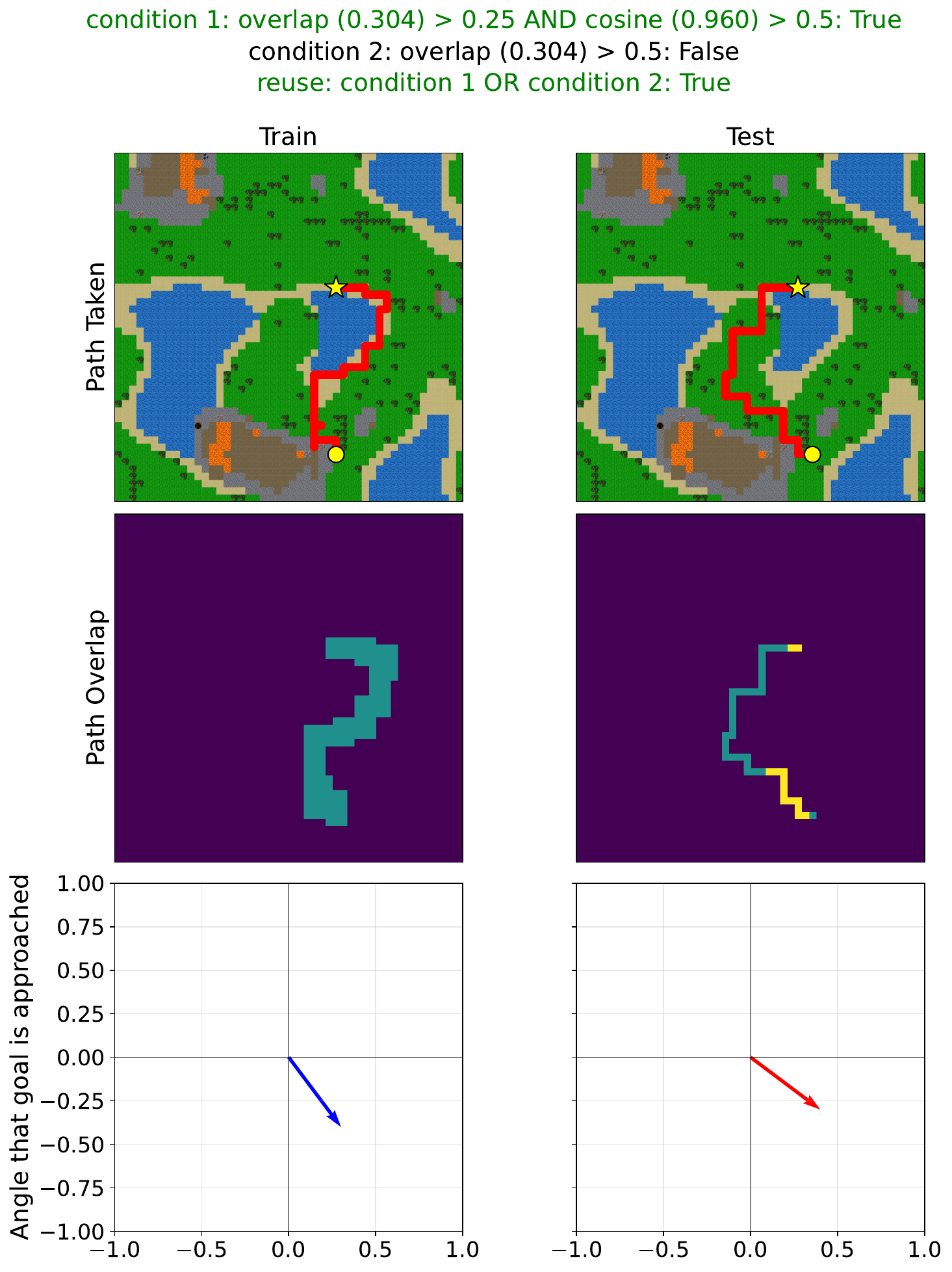}
        \subcaption{False positive.}
    \end{minipage}
    \caption{
        \textbf{Craftax path-reuse: both-clause success and a false positive.}
        (a) Both clauses are satisfied: the subject's test path overlaps the training path heavily and approaches the object from the same angle. (b) The criterion fires but the test path does not meaningfully reuse the training path.
    }
    \label{ext:craftax-overlap-tp-both-vs-fp}
\end{extfigure}

\begin{extfigure}[htbp]
    \centering
    \begin{minipage}[t]{0.48\textwidth}
        \centering
        \includegraphics[width=\textwidth]{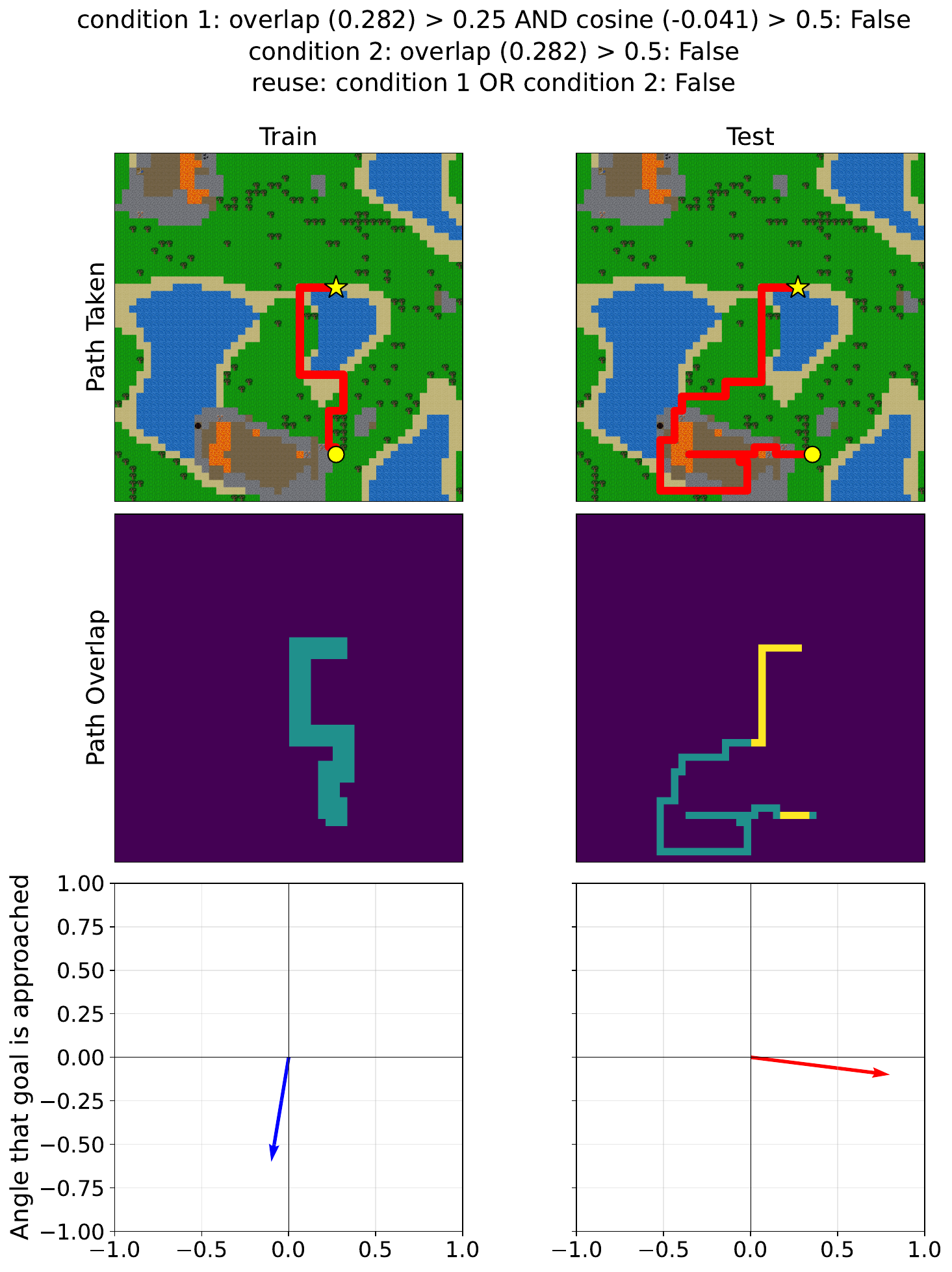}
        \subcaption{True negative.}
    \end{minipage}\hfill
    \begin{minipage}[t]{0.48\textwidth}
        \centering
        \includegraphics[width=\textwidth]{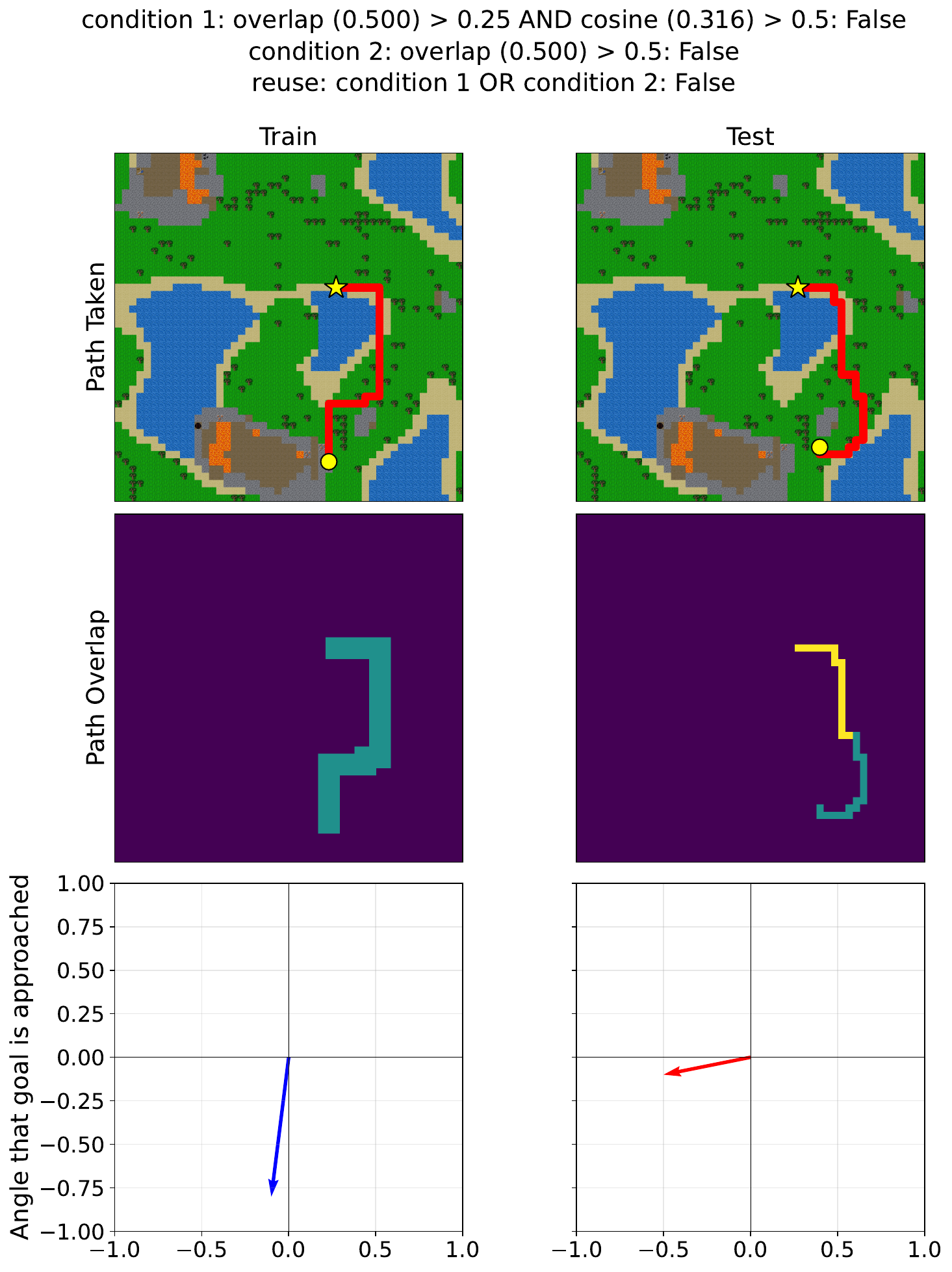}
        \subcaption{False negative.}
    \end{minipage}
    \caption{
        \textbf{Craftax path-reuse: true negative and a near-miss false negative.}
        (a) The subject takes a distinct route at evaluation and the criterion correctly returns zero. (b) The subject broadly reuses the training path but falls short on both map overlap and approach angle, so the criterion misses a partial reuse a human observer would credit.
    }
    \label{ext:craftax-overlap-tn-vs-fn}
\end{extfigure}

\begin{extfigure}[htbp]
    \centering
    \includegraphics[width=.75\textwidth]{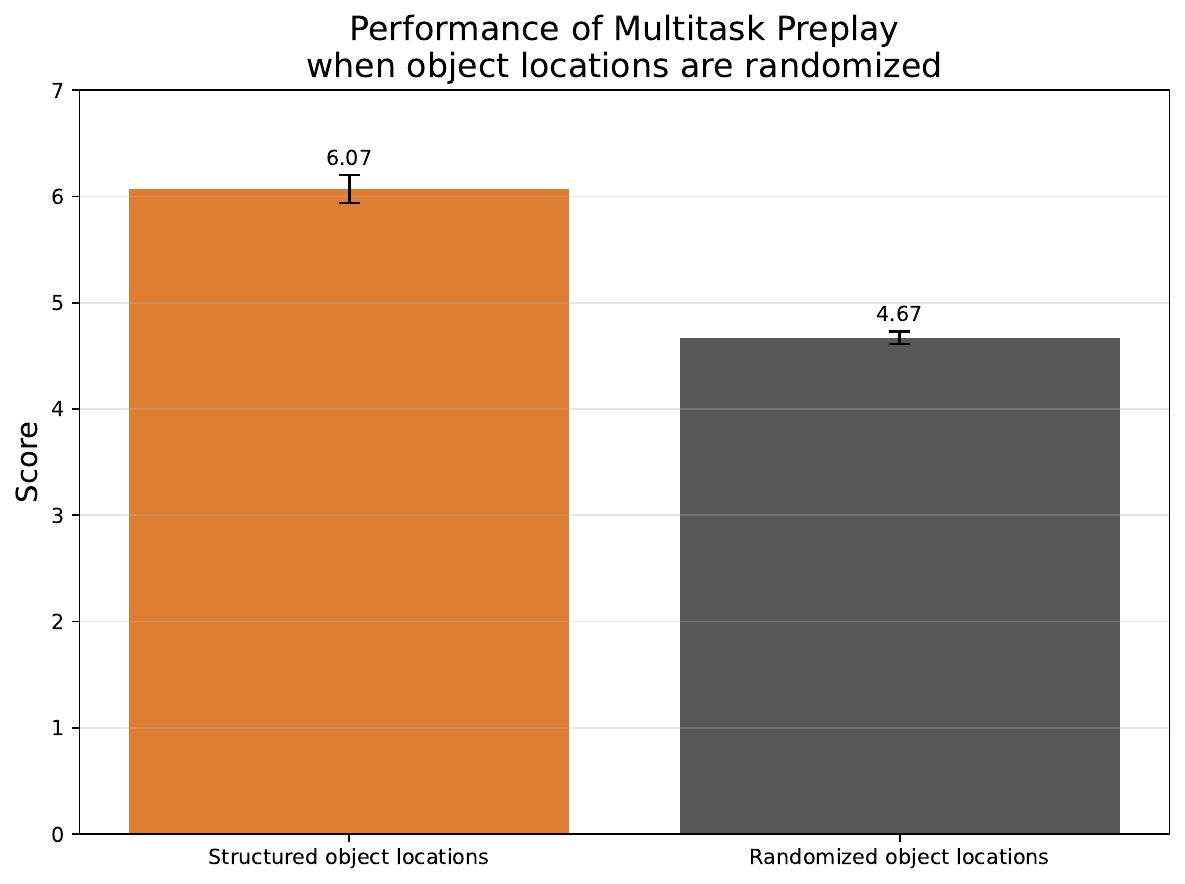}
    \caption{\textbf{Performance of Multitask Preplay when object locations are randomized instead of following default object co-occurrence structure}.}
    \label{ext:randomization_ablation}
\end{extfigure}

\begin{extfigure}[htbp]
    \centering
    \includegraphics[width=\textwidth]{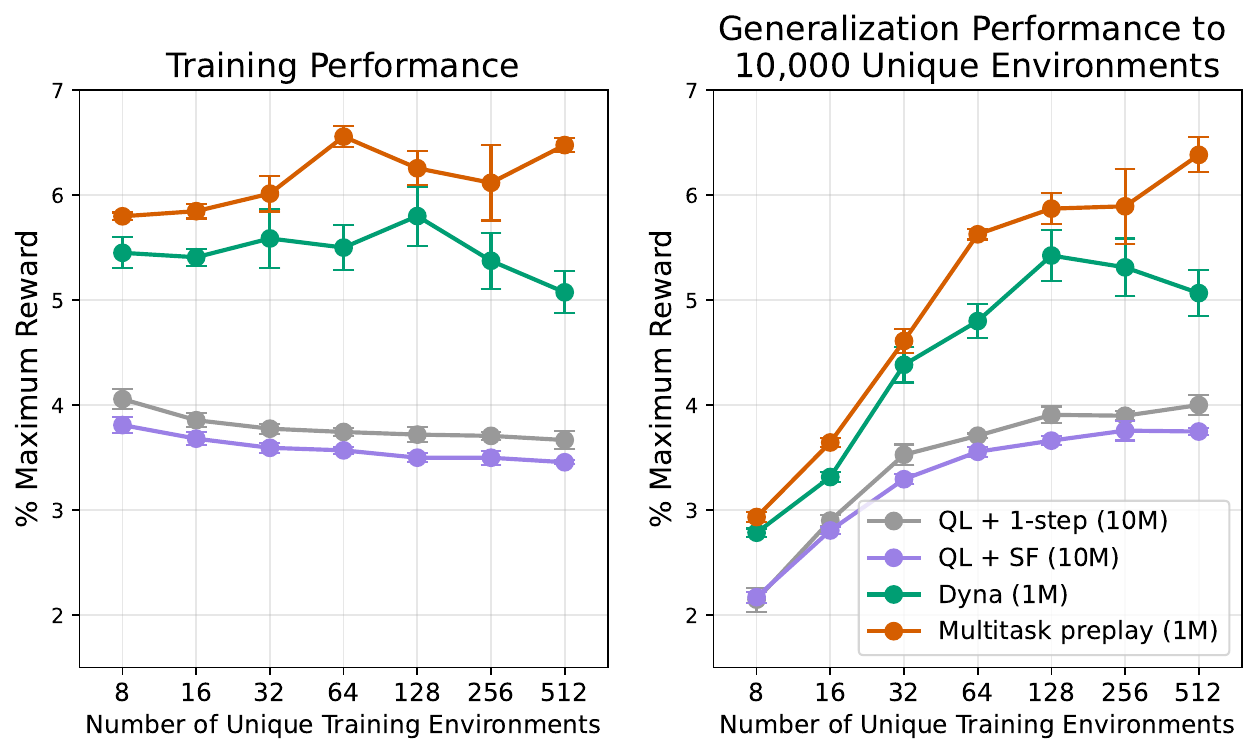}
    \caption{\textbf{Multitask Preplay consistently benefits both training and generalization from preplaying offtask goals $\offtask$}. Training and generalization performance for data in Figure 5.}
    \label{ext:train_eval}
\end{extfigure}

\begin{extfigure}[htbp]
    \centering
    \includegraphics[width=\textwidth]{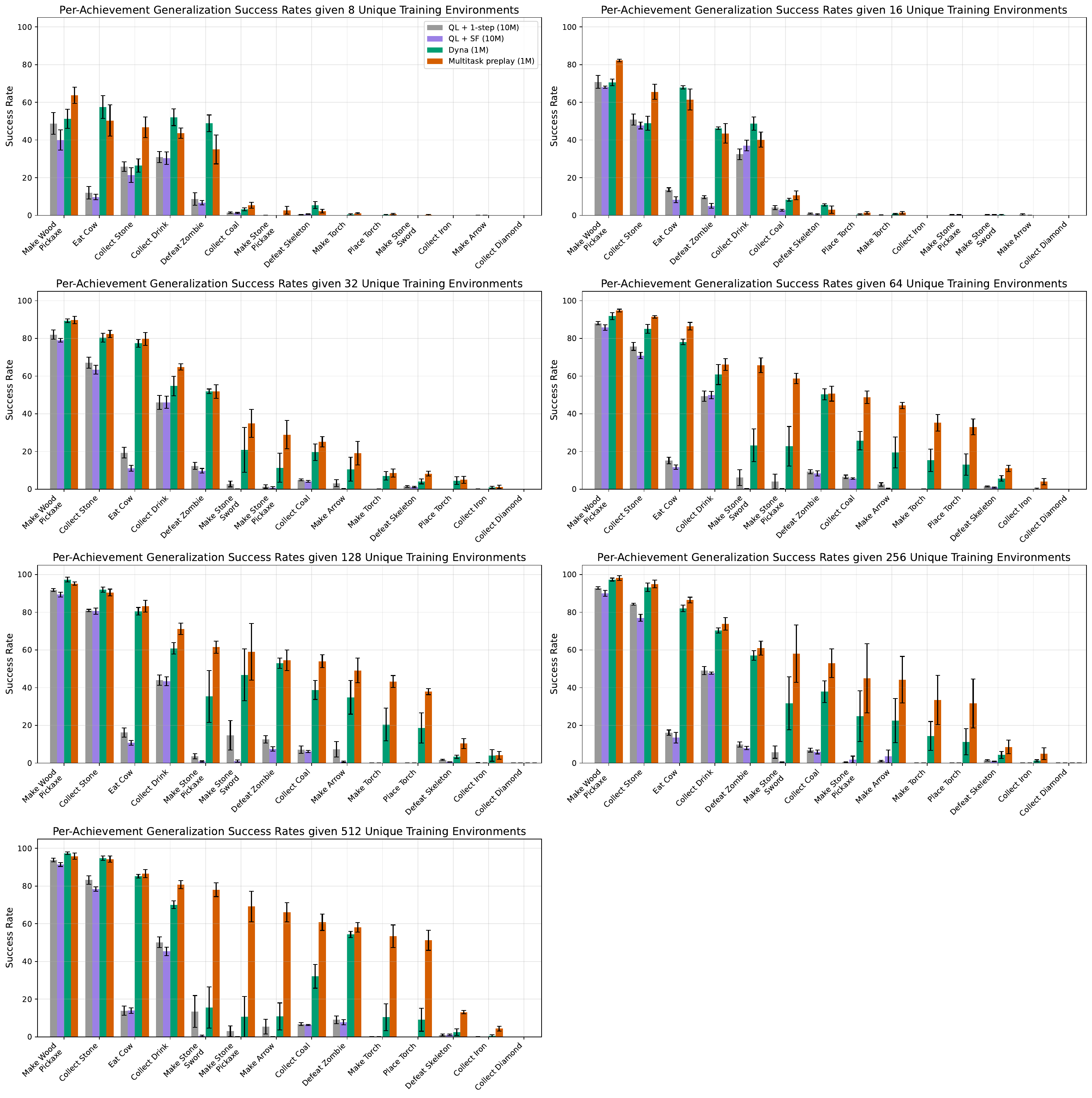}
    \caption{\textbf{Generalization performance of Multitask Preplay vs. Dyna}. As the number of training environments increases, Multitask Preplay consistently improves its generalization performance, despite learning on fewer samples per environment. In contrast, Dyna, which uses \textit{the exact same simulation budget} as Multitask Preplay begins to decrease in its generalization performance. }
    \label{ext:all_bars}
\end{extfigure}

\begin{extfigure}[htbp]
    \centering
    \includegraphics[width=.8\textwidth]{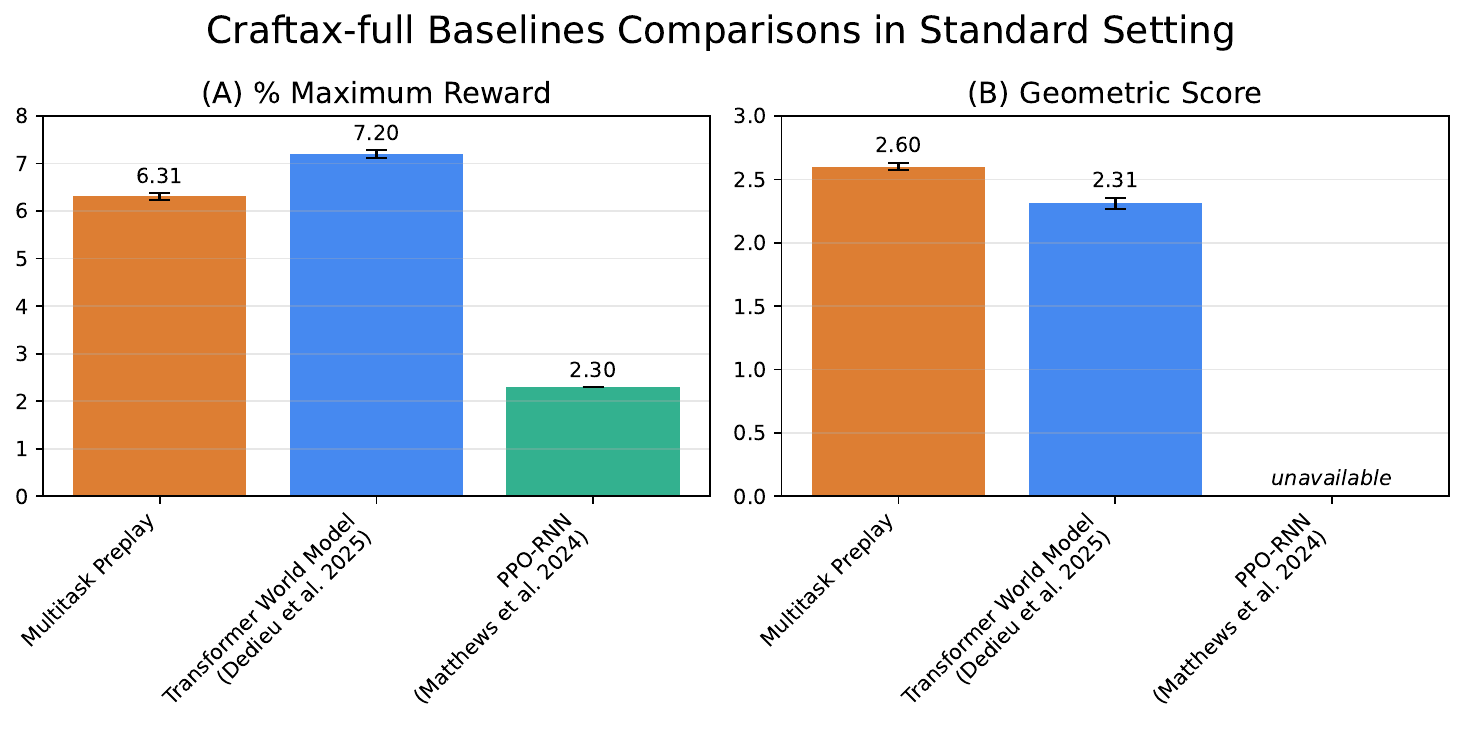}
    \caption{%
        We compare Multitask Preplay to the state-of-the-art background planning method (Transformer World Model; TWM~\cite{dedieu2025improving}) and the best model-free method (Proximal Policy Optimization; PPO~\cite{schulman2017proximal}) in the standard craftax setting (Craftax-Full~\cite{matthews2024craftax}).
        This setting differs from ours in that we study generalization as a function of the number of training environments, whereas the standard setting has a new environment sampled in every episode and measures performance within this episode.
        Multitask Preplay uses a less expressive architecture for modelling sequential dependencies (an RNN in Multitask Preplay vs. a Transformer~\cite{vaswani2017attention} in TWM), which may lead to less effective representation learning.
        We find that it can accomplish a wider range of tasks across episodes (see B), but fewer tasks within an episode (see A).
        \textbf{(A) \% Maximum Score} presented the average percentage of all possible tasks completed within an episode. The best background planning methods complete more tasks than the best purely model-free method. We find that TWM completes more tasks with an episode than Multitask Preplay on average.
        We hypothesize that this is due to its less expressive architecture for modelling sequential dependencies.
        \textbf{(B) Geometric Means} places more weight on occasionally solving many achievements than on consistently solving a subset. It is given by $S=\exp \left(\frac{1}{N} \sum_{i=1}^N \ln \left(1+s_i\right)\right)-1$, where $s_i \in[0,100]$ is the success percentage for achievement $i$.
        This measure was used to measure the performance of TWM but not PPO on Craftax-full.
        We find that Multitask Preplay has a higher score, indicating that is accomplishes a wider range of tasks.
        Given that these are generalization scores, we hypothesize that this is due to Multitask Preplay learning about subtasks outside of the ones it experienced during background planning---this in turn could lead it to be perform a wider array of tasks at generalization  (on average).
        Future work can look to integrate the more versatile background planning algorithm of Multitask Preplay with the more sophisticated architectural choices of TWM.
    }
    \label{ext:baseline_comparison}
\end{extfigure}

\begin{extfigure}[htbp]
    \centering
    \includegraphics[width=\textwidth]{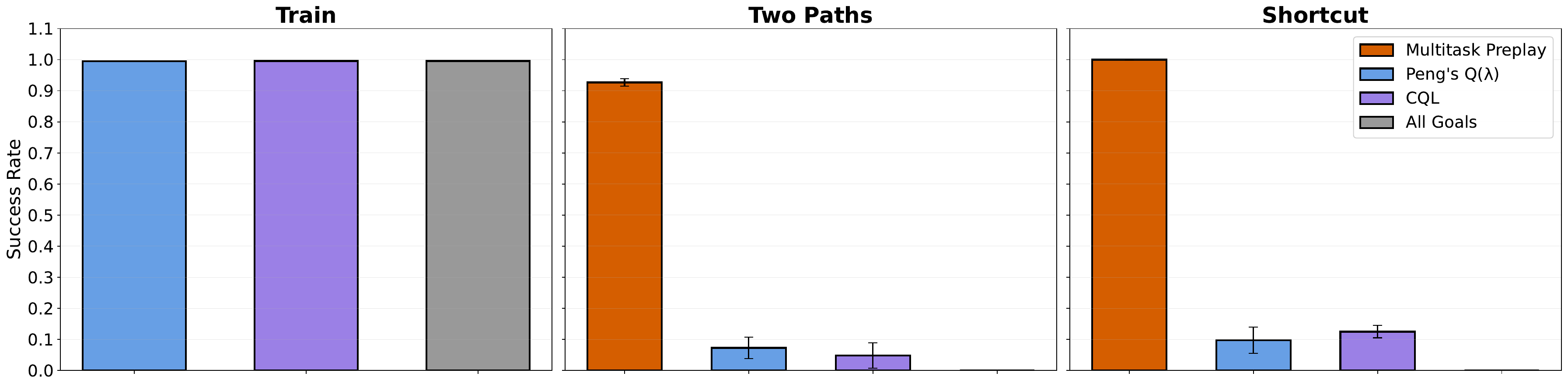}
    \caption{
        \textbf{Ablation of all-goals learning components.} Ablation of off-task $Q(\lambda)$, conservative Q-learning (CQL), and the combined conservative all-goals loss.
    }
    \label{ext:allgoals-ablation}
\end{extfigure}

\begin{extfigure}[htbp]
    \centering
    \includegraphics[width=.3\textwidth]{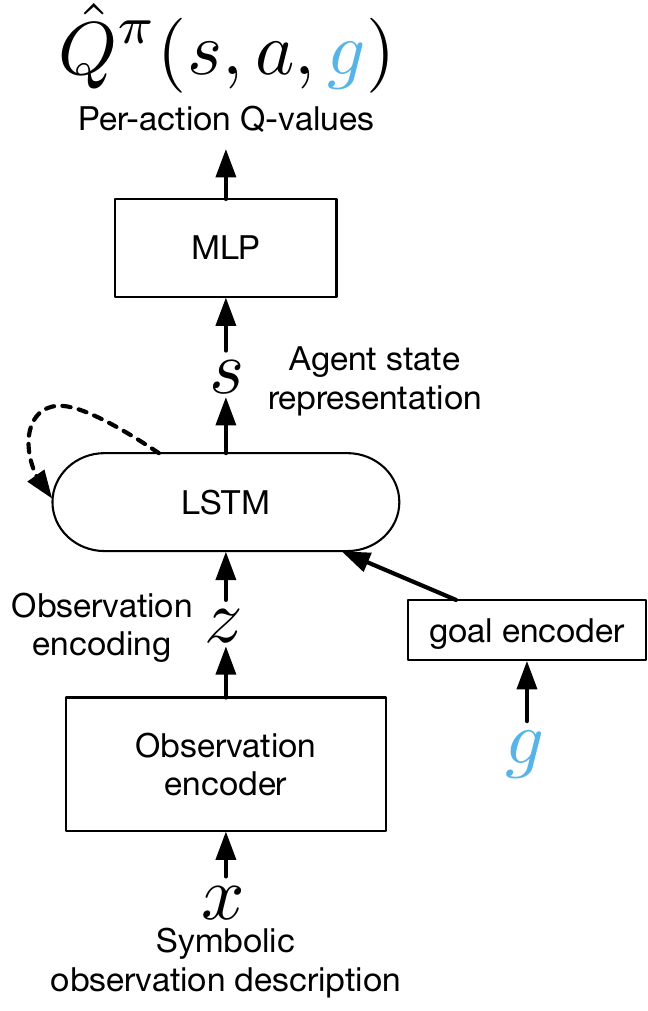}
    \caption{
        \textbf{Diagram of neural network architecture used by all methods that learn Q-values}.
    }
    \label{ext:q-network}
\end{extfigure}

\begin{extfigure}[htbp]
    \centering
    \includegraphics[width=\textwidth]{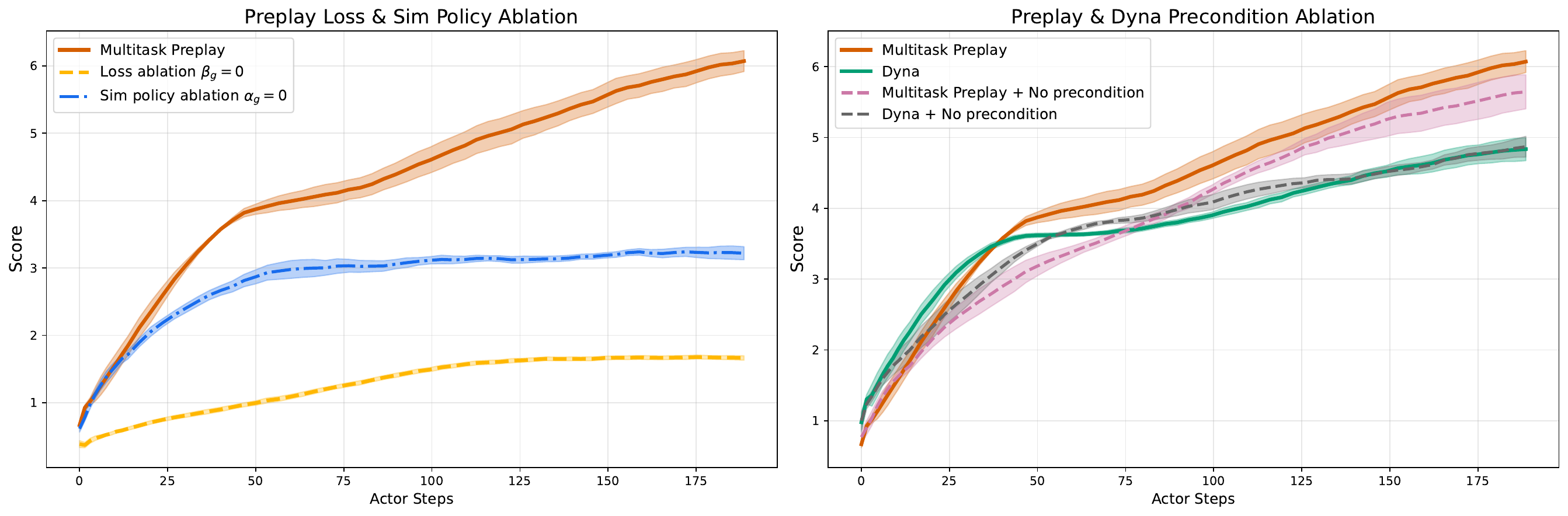}
    \caption{
        \textbf{(Left) Algorithm ablations}. It is not only key to simulate towards offtask goals $\offtask$, but to do so with a small bias towards the main goal $\goal$ ($\alpha_\goal > 0$), and to propagate this information to action-value predictions for $\goal$ (i.e. to have $\beta_\goal > 0$). Without either, the benefits of counterfactual task simulation are not fully realized.
        \textbf{(Right) Environment ablation}. Multitask preplay benefits from having $\phi_{\tt avail}$ knowledge of when preconditions are met, but this is not required to get the benefits off counterfactual task simulation. Indeed, without this precondition information (Multitask Preplay + No precondition), it still achieves a strong improvement over Dyna.
    }
    \label{ext:preplay_ablations}
\end{extfigure}

\begin{extfigure}[htbp]
    \centering
    \includegraphics[width=0.8\textwidth]{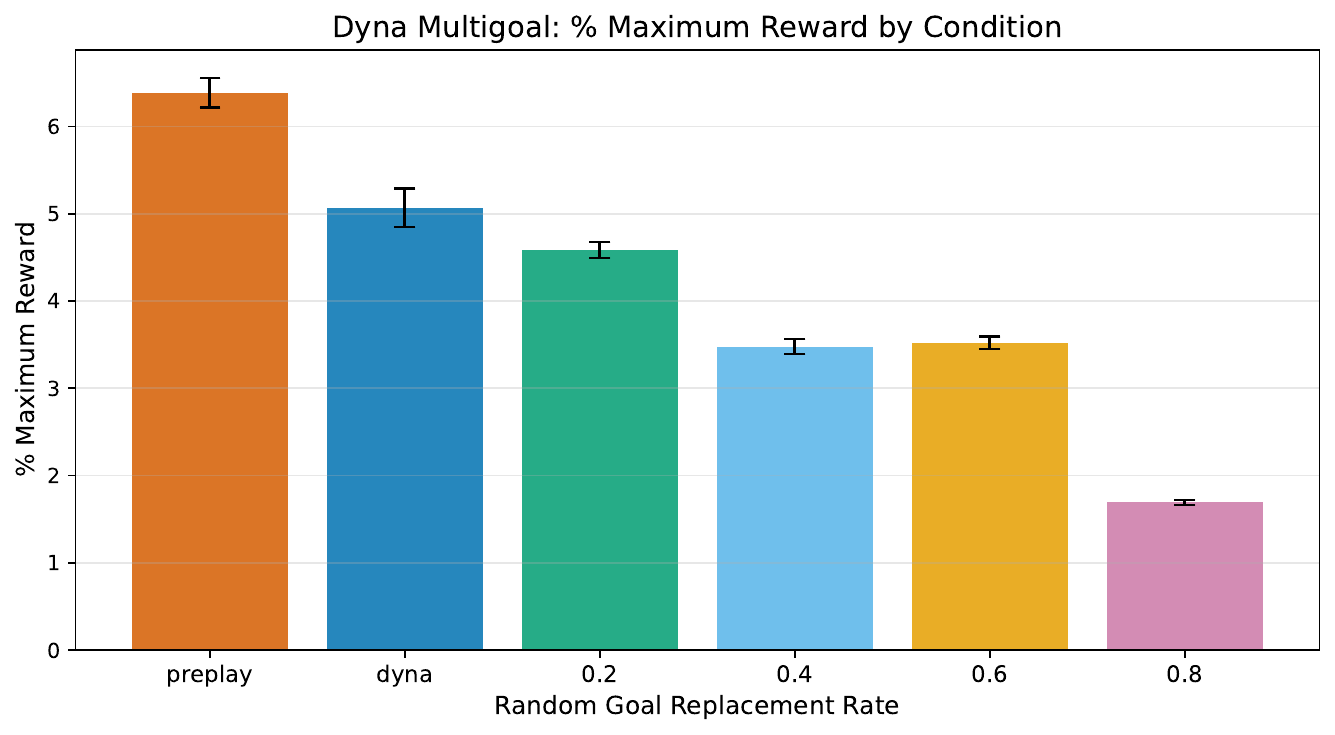}
    \caption{
        \textbf{Dyna with random goal replacement on Craftax.} Percentage of maximum reward achieved by Dyna variants that randomly replace the simulation goal with a uniformly sampled goal at varying rates, compared to standard Dyna and Multitask Preplay. Naively introducing off-task simulations does not match the performance of Multitask Preplay's structured all-goals learning.
    }
    \label{ext:dyna-multigoal}
\end{extfigure}

\end{document}